%% file: ms.tex
\documentclass{article}

\pdfoutput=1 % a flag for arXiv pdflatex processing

\usepackage{booktabs} % For formal tables
\usepackage{xcolor}

\usepackage{authblk} % add author affiliations
\usepackage{hyperref}

\usepackage[margin=3cm]{geometry}

\usepackage[toc,page]{appendix}
\usepackage{longtable}
\usepackage{pdflscape}

\usepackage{subfig}
\usepackage{microtype}
\usepackage[american]{babel}
\usepackage{amsmath, amsthm, amssymb, amsfonts}

\usepackage[ruled,linesnumbered]{algorithm2e}
\SetKwInOut{Require}{Require}
\SetKwInOut{Ensure}{Output}

\usepackage{todonotes}

\newcommand\aiprshort{AI$\leftrightarrow$PR}
\newcommand\aiprnn{\aiprshort{}-CNN}

\title{Learning Interpretable Shapelets for Time Series Classification through Adversarial Regularization}

\author[1]{Yichang Wang}
\author[2]{R\'emi Emonet}
\author[1]{Elisa Fromont}
\author[1]{Simon Malinowski}
\author[1]{Etienne Menager}
\author[1]{Lo\"ic Mosser}
\author[3]{Romain Tavenard}
\affil[1]{Univ Rennes, Inria, CNRS, IRISA, Rennes, France}
\affil[2]{Laboratoire Hubert Curien UMR 5516,  Univ Lyon, Saint-Etienne, France}
\affil[3]{Univ Rennes, LETG, IRISA, Rennes, France}

\date{}
\begin{document}

\maketitle

\begin{abstract}
Times series classification can be successfully tackled by jointly
learning a shapelet-based representation of the series in the dataset
and classifying the series according to this representation. However,
although the learned shapelets are discriminative, they are not always
similar to pieces of a real series in the dataset. This makes it
difficult to interpret the decision, i.e. difficult to analyze if
there are particular behaviors in a series that triggered the
decision. In this paper, we make use of a simple convolutional network
to tackle the time series classification task and we introduce an adversarial
regularization to constrain the model to
learn more interpretable shapelets. Our classification results on all the usual
time series benchmarks are comparable with the results obtained by similar state-of-the-art algorithms but our adversarially regularized method learns shapelets that are, by design, interpretable.
\end{abstract}

\input{sections/intro}
\input{sections/related}
\input{sections/method}
\input{sections/xp}

\section{Conclusion}
We have presented a new shapelet-based time series classification method that produces interpretable shapelets. The shapelets are deemed interpretable because they are similar to pieces of a real series and can thus be used to explain a particular model prediction. The method is based on a novel adversarial architecture where one convolutional neural network is used to classify the series and another one is used to constrain the first network to learn interpretable shapelets.
Our results show that the expected trade-off between accuracy and interpretability is satisfactory: our classification results are comparable with similar state-of-the-art methods while our shapelets are interpretable. 

We believe that the proposed adversarial regularization method could be used in many more applications where the regularization should be put on the parameters instead of the latent representation of the networks as done, for example, with Generative Adversarial Networks. 

In future work, we would like to first investigate the use of an additional regularization term, based on the group lasso~\cite{bascol2016Unsupervised}, 
to be able to determine automatically a minimal set of necessary interpretable shapelets.
We also want to use our regularization on other types of data (such as multivariate time series, spatial data, graphs) and in a deep(er) CNN.
Furthermore, we would like to adapt this architecture for unsupervised anomaly detection in time series with interpretable clues using neural network architectures such as convolutional auto-encoders or generative networks.

\bibliographystyle{abbrv}
\bibliography{biblioShapelets}

\clearpage{}
\input{sections/supp}

\end{document}

%% file: sections/intro.tex
\section{Introduction}
\label{sec:intro}

A time series (TS) $Z$ is a series of time-ordered values, $Z = \{{{z}}^{(1)},$ ${{z}}^{(2)}, \ldots, {{z}}^{(T)}\}$ where ${{z}}^{(t)} \in \mathbb{R}^{d}$, $T$ is the length of our time series and $d$ is the dimension of the feature vector describing each data point. If $d=1$, $Z$ is said univariate, otherwise it is said multivariate. In this paper, we are interested in the Time Series Classification (TSC) task. We are given a training set $\mathcal{T} = \left\{ ({Z}_1, y_1),\ldots, ({Z}_n, y_n) \right\}$, composed of $n$ time series ${Z}_i$ and their associated labels $y_i$ (target variable).  Our aim is to learn a function $h$ such that $h(Z_i) = y_i$, in order to predict the labels of new incoming time series. The time series classification problem has been studied in countless applications (see for example \cite{Shumway2005}) ranging from stock exchange evolution, daily energy consumption, medical sensors, videos, etc.

\begin{figure}[t]
	\centering
	\includegraphics[width=.6\linewidth]{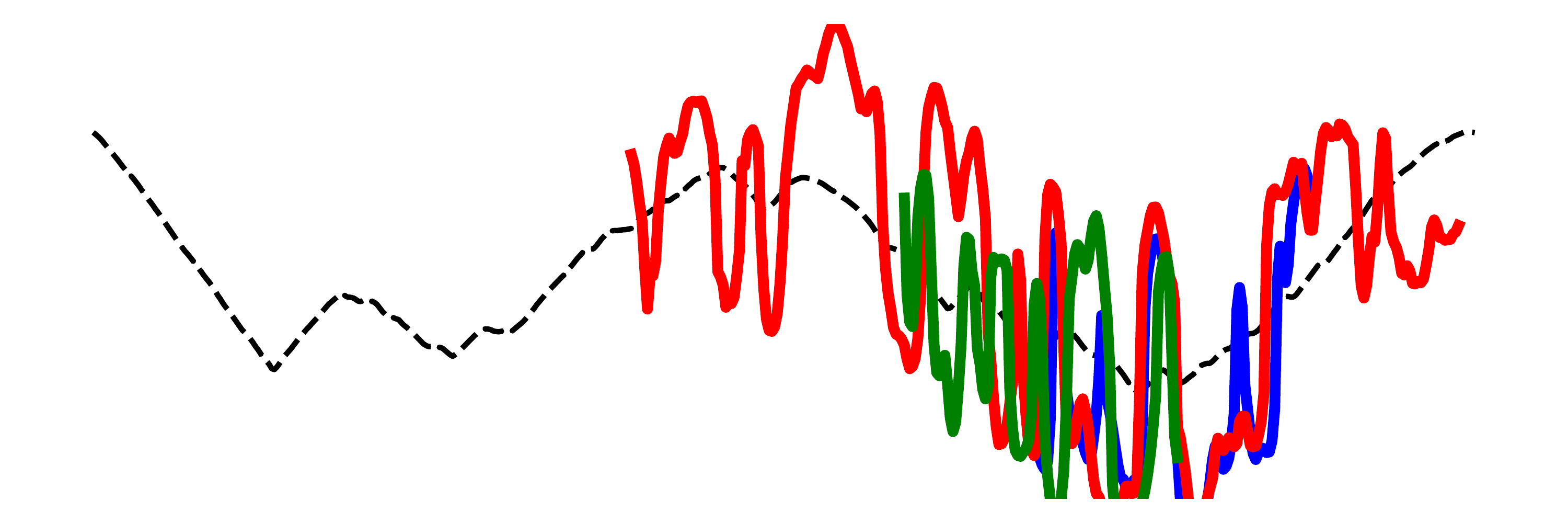}
	\includegraphics[width=.6\linewidth]{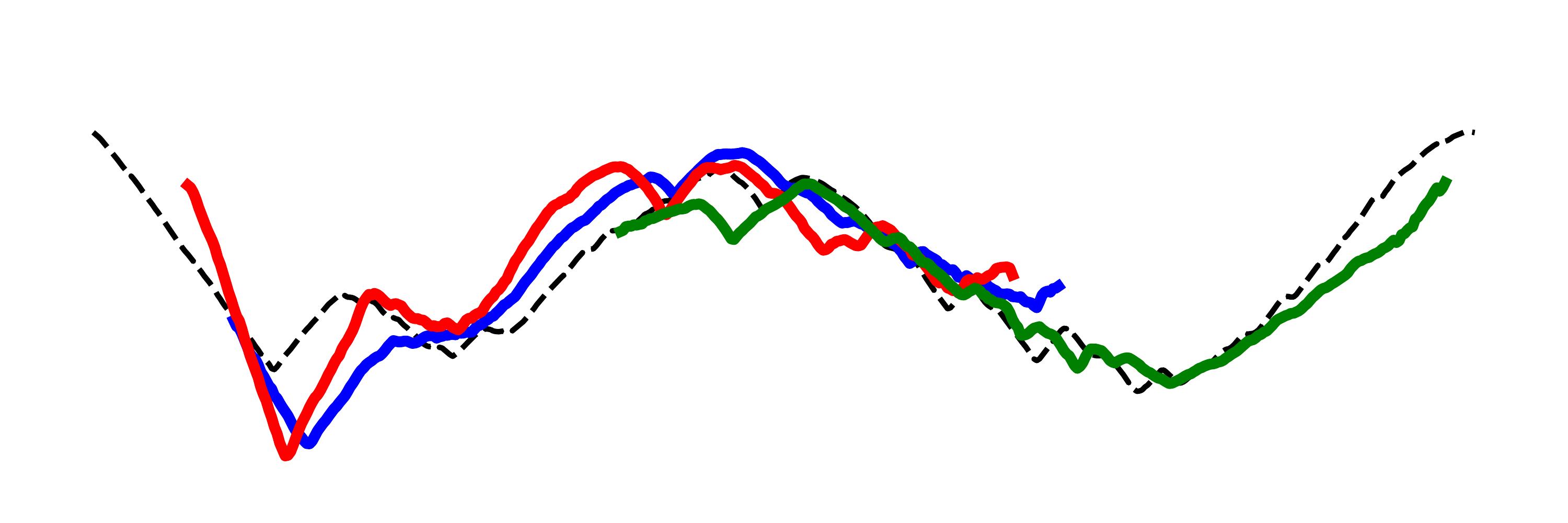}

	\caption{Example test time series and three most discriminative shapelets used for its classification for a baseline~\cite{Grabocka2014} (top) and for our proposed \aiprshort{}-CNN model (bottom) on the Herring classification problem. \label{fig:illus_intro}}
\end{figure}

Many methods have been developed to tackle this problem (see~\cite{Bagnall2017} for a review). One very successful category of methods consists in "finding" discriminative phase-independent subsequences, called \emph{shapelets}, that can be used to classify the series. In the first papers about shapelet-based time series classification \cite{ye2009time,Rak13}, the shapelets were directly extracted from the training set and the selected shapelets could be used \emph{a posteriori} to explain the classifier's decision. However, the shapelet enumeration and selection processes were either very costly or the selection was fast but did not yield good performance (as discussed in Section \ref{sec:related}). Jointly learning a shapelet-based representation of the series in the dataset and classifying the series according to this representation~\cite{lines2012shapelet,Grabocka2014} allowed to obtain discriminative shapelets in a much more efficient way. An example of such a learned shapelet, obtained with the method from~\cite{Grabocka2014}, is given in Figure~\ref{fig:illus_intro} (top). However, if the learned shapelets are definitively discriminative, they are often different from actual pieces of a real series in the dataset. As such, the classification decision is difficult to interpret, i.e. it is difficult to determine what particular behavior in a time series triggered the classification decision. Note that the same interpretability issue arises with ensemble classifiers such as~\cite{bagnall2015time} where one decision depends on the presence of multiple shapelets. One of the main challenge nowadays is to enrich Machine Learning (ML) systems, and in particular black box models such as neural networks, so that they have the ability to explain their outputs to human users. In many scenarios, it may be risky, unacceptable, or simply illegal, to let artificial intelligent systems make decisions without any human supervision~\cite{Guidotti2018}. Hence, it is necessary for ML systems to provide an explanation of their decisions to all the humans concerned.

In this paper, we make use of a simple convolutional network to classify time series and we show how one can use adversarial techniques to regularize the parameters of this network such that it learns shapelets that could be more useful to interpret the classifier's decision. Section~\ref{sec:related} presents the related work on time series classification, interpretability of models and adversarial training. We present our adversarial parameter regularization method in Section~\ref{sec:method}. In Section~\ref{sec:xp}, we show quantitative and qualitative results on the usual time series benchmarks~\cite{UCRArchive} that are both on par with state-of-the-art methods and very interesting to interpret the neural network predictions.

%% file: sections/related.tex
\section{Related Work}
\label{sec:related}

In this section we review the literature on Time Series Classification (TSC), on tools for understanding black box model predictions and on adversarial training.

\subsection{Time Series Classification}
In the TSC literature, two main families of approaches have been designed.
First, a dedicated metric can be used to compare the time series. In this case the decision is based on the resulting similarities. For example,~\cite{sakoe1978dynamic} uses Dynamic Time Warping (DTW) to find an optimal alignment between time series and provides an alignment cost that can be used to assess the similarity.
Another family of methods is based on the extraction of features in the time series. Among these works, shapelet-based classifiers have attracted a lot of attention from the research community.

Shapelets are discriminative subseries that can either be extracted
from a set of time series or learned so as to minimize an objective
function. They have been introduced in~\cite{ye2009time}, in which a
binary decision tree is built, whose nodes are shapelets and whose subtrees
contain subsets of time series that contain or not that shapelet. In
this work, shapelets are extracted from a training set of
time series and building the decision tree requires to test all
possible subseries from the training set, which makes the method
intractable for large-scale learning with an overall time complexity of $O(n^2 \cdot T^4)$ where $n$ is the number of training time series and $T$ is the average length of the time series in the training set.
This high time complexity has led to the use of heuristics in order to select
the shapelets more efficiently. In~\cite{Rak13} (Fast Shapelets), the authors rely on quantized time series and random projections in order
to fasten the shapelet search. Note however that these improvements
in time complexity are obtained at the cost of a lower classification
accuracy, as reported in~\cite{Bagnall2017}. The Shapelet Transform
(ST)~\cite{lines2012shapelet} consists in transforming time series
into a feature vector whose coordinates represent distances between
the time series and the shapelets selected beforehand. It hence needs to
select a shapelet set (as in~\cite{ye2009time}) before transforming
the time series. The resulting vectors are then given to a classifier
in order to build the decision function. The training time complexity for ST is also in $O(n^2 \cdot T^4)$~\cite{DBLP:journals/corr/abs-1809-04356}, % for a training set of $n$ time series of length $l$, 
which makes it unfit for large scale learning.

In order to face the high complexity that comes with search-based
methods, other strategies have been designed for shapelet
selection. On the one hand, some attention has been paid to random
sampling of shapelets from the training
set~\cite{Karlsson2016}. On the other hand, Grabocka
\emph{et al.}~\cite{Grabocka2014} showed that shapelets could be
learned using a gradient-descent-based optimization algorithm. The method, referred to as Learning Shapelets (LS) in the following, jointly learns the shapelets and the parameters of a logistic regression classifier. This makes the method very similar in spirit to a neural network with a single convolutional layer followed by a fully connected classification layer and where the convolution operation is replaced by a sliding-window local distance computation. A min-pooling aggregator should then be used for temporal aggregation.

Closely related to shapelet-based methods (as stated above), variants of Convolutional Neural Networks (CNN) have been introduced for the TSC task~\cite{wang2017time}. These are mostly mono-dimensional variants of CNN models developed in the Computer Vision field. Note however that most models are rather shallow, which is likely to be related to the moderate sizes of the benchmark datasets present in the UCR/UEA archive~\cite{UCRArchive}. A review of these models can be found in~\cite{DBLP:journals/corr/abs-1809-04356}.

Finally, ensemble-based methods, such as COTE~\cite{bagnall2015time}
or HIVE-COTE~\cite{lines2018time}, that rely on several of the
above-presented standalone classifiers are now considered
state-of-the-art for the TSC task. Note however
that these methods tend to be computationally expensive, with high
memory usage and difficult to interpret (as stated in Section~\ref{sec:intro}) due to the combination of many different core classifiers.

In this paper, we propose a method that is scalable (compared to methods such as Shapelets~\cite{ye2009time} or ST~\cite{lines2012shapelet}), yields interpretable results which can be used to explain the classifier's decision (compared to ensemble approaches or unconstrained approaches such as~\cite{Grabocka2014} or~\cite{lines2018time}), and exhibits good classification accuracy (compared to FS~\cite{Rak13}).

\subsection{Model Interpretability}
Among the vast number of existing classifiers, some are easily interpretable (e.g. decision trees, classification rules), while others are difficult to interpret (e.g. ensemble methods, neural networks that can be considered as black-boxes). Interpretation of black box classifiers usually consists in designing an interpretation layer between the classifier and the human level. Two criteria refine the category of methods to interpret classifiers: global versus local explanations, and black-box dependent versus agnostic. In this category, state-of-the-art methods are Local Interpretable Model-agnostic Explanations (LIME and Anchors) \cite{Ribeiro0G16, Ribeiro0G18} and SHapley Additive exPlanations (SHAP) \cite{LundbergL17}. SHAP values come with the black-box local estimation advantages of LIME, but also with theoretical guarantees. A higher absolute SHAP value of an attribute compared to another means that it has a higher predictive or discriminative power. 

In this paper, we are interested in making the decision of a neural network understandable.
We follow the concept of interpretable shapelet as in~\cite{Grabocka2014}: for a TSC model, a simple explanation should not directly come from the vector of attributes describing each point of each time series but rather from some discriminative shapelets internally learned to produce an intermediate representation to classify the series.
Solutions such as LIME, Anchors and SHAP which are not designed to inspect the internal representation of a model are thus not well suited for our problem.

Fang \emph{et al.}~\cite{fang2018efficient} have a similar goal as ours (to produce interpretable discriminative shapelets) and build on both the works from~\cite{lines2012shapelet} (in this case the candidate shapelets are extracted with a piecewise aggregate approximation) and from~\cite{Grabocka2014} to automatically refine the ``handcrafted'' shapelets. Contrarily to our method, there is no explicit constraint on the learning process that ensures the interpretability of the shapelets. Besides, their experimental validation makes it hard to fully grasp the benefits and limitations of the proposed method since the algorithm is evaluated on a small subset of UCR/UEA datasets~\cite{UCRArchive} and they provide visualizations for only a couple of the learned shapelets.

\subsection{Adversarial Training}
Adversarial training of neural networks has been popularized by Generative Adversarial Networks (GANs)~\cite{GoodfellowGAN} and their numerous variants.\footnote{See \url{https://github.com/hindupuravinash/the-gan-zoo} for a list.}
A GAN is a combination of two neural networks: a generator and a discriminator which compete against each other during the training process to reach an equilibrium where the discriminator cannot distinguish between the generator outputs and real training data. In a GAN, the adversarial network is used to push the generator towards producing data as similar to real data as possible.
Other (non generative) adversarial training settings have been studied, for example in the context of domain adaptation~\cite{TzengHSD17}. In this case, the adversarial network is used to regularize the latent representation learned by the classifier such that it becomes domain-independent.
The recent work from \cite{zhao18} also uses adversarial regularization to constrain the latent representation of an autoencoder to follow a given distribution.

In this paper, we propose an adversarial regularization approach which is unique as
1) we use a non-generative adversarial approach,
2) we do not work on a latent representation or on the output of a generator but rather on the CNN convolution filters (i.e., it is used as a parameter regularization),
and, 3) we leverage this regularization to encourage interpretability, by making the convolution filters similar to real subseries from the training data.

%% file: sections/method.tex
\section{Learning Interpretable Shapelets}
\label{sec:method}

\begin{figure*}[!ht]
	\includegraphics[width=\linewidth]{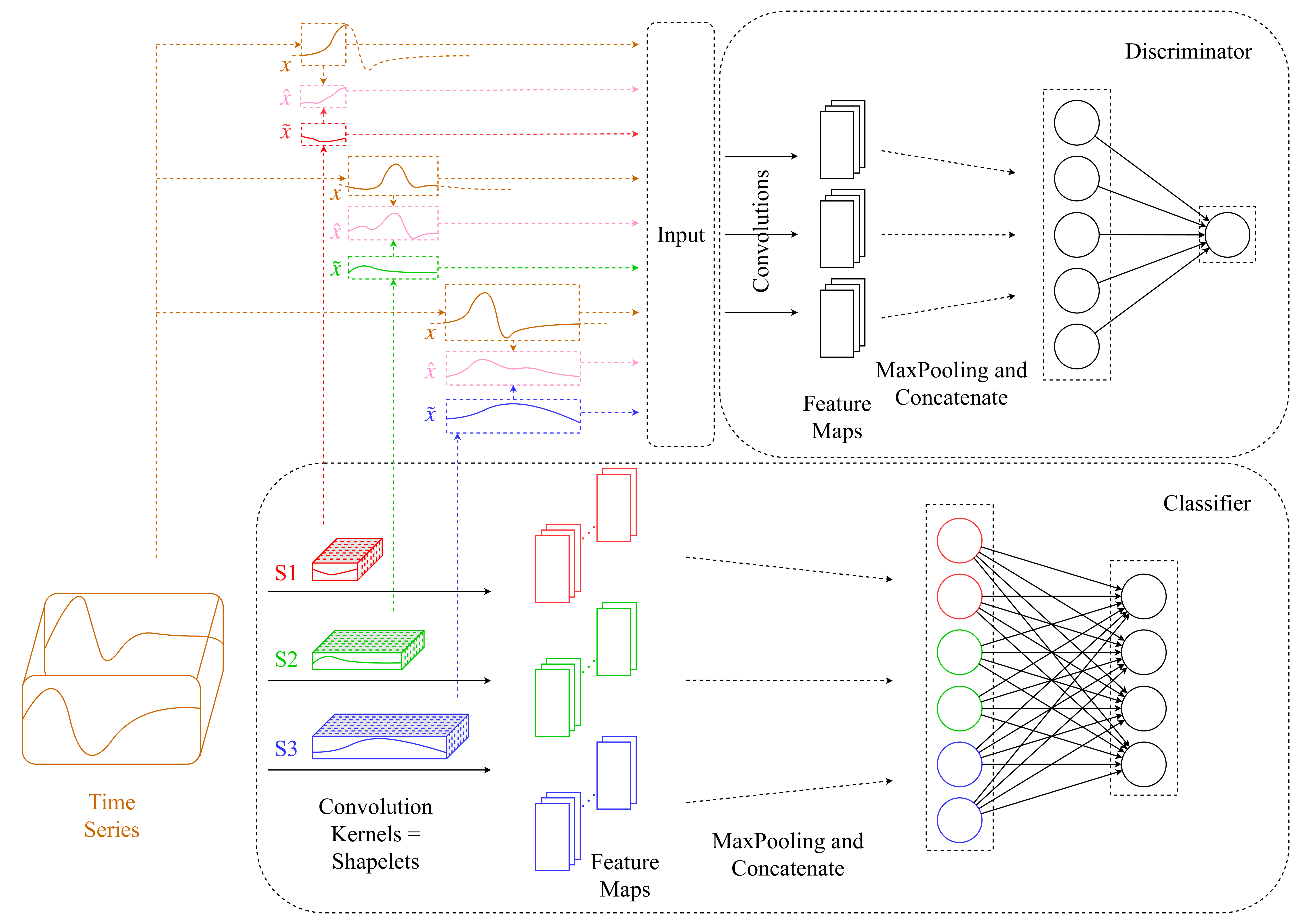}
	\caption{Architecture of our proposed Adversarially Input$\leftrightarrow$Parameter Regularized CNN (\aiprshort{}-CNN)}\label{fig:arn}
\end{figure*}

In this section, we present our approach to learn interpretable discriminative shapelets for time series classification.

Our base time series \emph{classifier} is a Convolutional Neural Network (CNN). %that is trained using standard mini-batch stochastic gradient descent methods dedicated to neural networks ~\cite{Goodfellow2016}. 
As explained in Section~\ref{sec:related}, this model is very similar in spirit to the Learning Shapelet (LS) model presented in~\cite{Grabocka2014}.

Both LS and CNN slide the shapelets on the series to compute local (dis)similarities.
The main difference between the classifier of LS and that of our method is the (dis)similarity between a shapelet and a series. 
LS uses a squared Euclidean distance between a portion of the time series $Z$ starting at index $i$ and a shapelet $S$ of length $L$:

\begin{equation}
	D(z_{i:i+L}, S)=\sum_{l=1}^{L} \left( z^{(i+l-1)} - S^{(l)}\right)^2 .	
\end{equation}
The smaller this distance, the closer the shapelet is to the considered subseries.
In a CNN, the feature map is obtained from a convolution, and hence encodes cross-correlation between a series and a shapelet:
\begin{equation}
	D(z_{i:i+L}, S) = \sum_{l=1}^{L} z^{(i+l-1)} \cdot S^{(l)} .
\end{equation}
Note that here, the higher $D(z_{i:i+L}, S)$, the more similar the shapelet is to the subseries.

As shown in Figure~\ref{fig:arn} (bottom), the convolutional layer of this classifier is made of three parallel convolutional blocks with shapelets of different lengths (red, green, blue) to be comparable with the structure proposed in LS.
We will loosely refer to the convolution filters of our classifier as \emph{Shapelets} in the following.

Inspired by previous works on adversarial training (see e.g. Section~\ref{sec:related}), in addition to our CNN classifier, we make use of an adversarial neural network (the discriminator at the top of Figure~\ref{fig:arn}) to regularize the convolution parameters of our classifier. This regularization acts as a soft constraint for the classifier to learn shapelets as similar to real pieces of the training time series as possible.

This novel regularization strategy is referred to, in the following, as Adversarial Input$\leftrightarrow$Parameter Regularization (\aiprshort{}) and the corresponding model is named \aiprshort{}-CNN. % in the following.

Contrarily to GANs, our adversarial architecture does not rely on a generator to produce fake samples from a latent space. The \aiprshort{} strategy iteratively modifies the shapelets (i.e. the convolution filters of the classifier) such that they become close to subseries from the training set. 
To execute this strategy, the discriminator is trained to distinguish between real subseries from the training set and the shapelets. 
During the regularization phase, the discriminator updates the shapelets so that they become more and more similar to real subseries. 

To obtain the best trade-off between the discriminative power of the shapelets (i.e. the final classification performance) and their interpretability, our training procedure alternates between training the discriminator and the classifier.

\begin{figure}
\centering
  \includegraphics[width=0.6\linewidth]{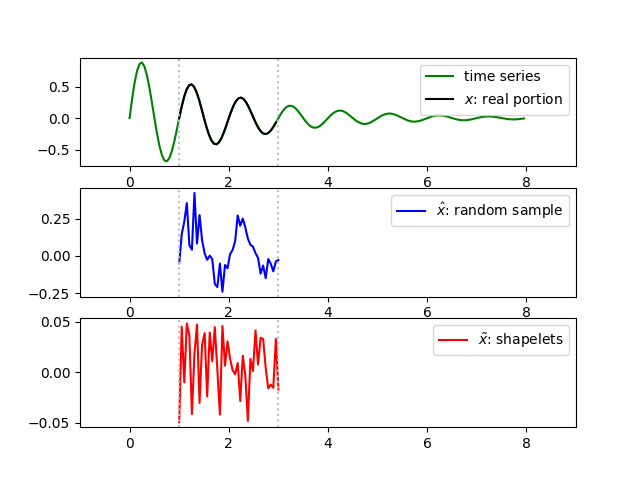}
	\caption{An example of samples provided as input to the discriminator. \label{fig:portions}}
\end{figure}

The type of data given as input to the discriminator is another major difference between a GAN and \aiprshort{}-CNN: in a GAN, the discriminator is fed with complete instances, while in \aiprshort{}-CNN, the discriminator takes subseries as input. These subseries can either be shapelets from the classifier model (denoted as $\tilde{x}$ in Figure~\ref{fig:arn}), portions of training time series (denoted as $x$) or interpolations between shapelets and training time series portions ($\hat{x}$, see the following section for more details on those), as illustrated in Figure~\ref{fig:portions}.
This process allows the discriminator to alter the shapelets for better interpretability.

\subsection{Loss Function}
As for GANs, our optimization process alternates between losses attached to the subparts of our \aiprshort{}-CNN model. Here, each training epoch consists of three main steps that are (i) optimizing the classifier parameters for correct classification, (ii) optimizing the discriminator parameters to better distinguish between real subseries and shapelets and (iii) optimizing shapelets to fool the discriminator.
Each of these steps is attached to a loss function that we describe in the following.

Firstly, a multi-class cross entropy loss is used for the classifier. It is denoted by $L_c(\theta_c)$ where $\theta_c$ is the set of all classifier parameters.

Secondly, our discriminator is trained using a loss function derived from the Wasserstein GANs with Gradient Penalty (WGAN-GP)~\cite{Gulrajani2017improved}:
\begin{align}
	L_d(\theta_d) = & \mathop{\mathbb{E}}_{\tilde{x}\sim\mathbb{P}_{S}} \left[ D(\tilde{x}) \right] 
					- \mathop{\mathbb{E}}_{x\sim\mathbb{P}_{x}} \left[D(x)\right] \nonumber \\
					& + \lambda \mathop{\mathbb{E}}_{\hat{x}\sim\mathbb{P}_{\hat{x}}}\left[(||\nabla_{\hat{x}}{D(\hat{x})}||_2-1)^2\right]
	\label{ld}
\end{align}
where $\mathbb{P}_{S}$ is the empirical distribution over the shapelets, $\mathbb{P}_{x}$ is the empirical distribution over the training subseries, and 
\begin{equation}
		\hat{x} = \epsilon x + (1 - \epsilon) \tilde{x} ,
\end{equation}
where $\epsilon$ is drawn uniformly at random from the interval $[0, 1]$ (cf. Figure~\ref{fig:portions}) .

Thirdly, shapelets are updated to fool the discriminator by optimizing on the loss $L_r(\theta_s)$ where $\theta_s \subset \theta_c$ is the set of shapelet coefficients:
\begin{equation}
	L_r(\theta_s) = - \mathop{\mathbb{E}}_{{\tilde{x}}\sim\mathbb{P}_{S}}[D(\tilde{x})] .
	\label{lr}
\end{equation}

\subsection{Learning Algorithm}

{\SetAlgoNoLine
	\begin{algorithm}[!t]
	\caption{Learning Interpretable Shapelet}\label{algo}
		\LinesNumbered
		\DontPrintSemicolon
        \Require{number of shapelets $n_S$}
        \Require{random initialization for the classifier/discriminator/shapelets $\theta_c, \theta_d, \theta_s\subset\theta_c$}
		\Require{gradient penalty coefficient $\lambda$}
        \Require{number of epochs $n_\text{epochs}$, mini-batch size $m$}
        \Require{number of classifier/discriminator/regularization mini-batches per epoch $n_c, n_d, n_r$}
		\Require{optimizer (Adam) hyperparameters $\alpha, \beta_1, \beta_2$}
		\For{$i=1, \dots, n_\text{epochs}$}{
			\For{$t=1, \dots, n_c$}{
				\For{$j = 1, \dots, m$}{
					Sample a pair $(Z_j, y_j)$ from the training set\;
					$\hat{y}_j \leftarrow h_{\theta_c}(Z_j)$\;
					$L_c^{(j)} \leftarrow \text{CrossEntropy}(y_j, \hat{y}_j)$\;
				}
				$\theta_c \leftarrow \text{Adam}(\nabla_{\theta_c}\frac{1}{m}\sum_{j=1}^{m}L_c^{(j)}, \theta_c, \alpha, \beta_1, \beta_2)$\;
			}
			\For{$t=1, ..., n_d$}{
				\For{$j = 1, ..., m$}{
					Sample a shapelet $\tilde{x}_j$ from the set $\theta_s$, a subseries $x_j$ from the training set and a random number $\epsilon\sim U[0,1]$\;
					$\hat{x}_j \leftarrow \epsilon x_j + (1-\epsilon) \tilde{x}_j$\;
					$L_d^{(j)} \leftarrow D(\tilde{x}_j)-D(x_j)+\lambda(||\nabla_{\hat{x}}{D(\hat{x}_j)}||_2-1)^2$\;
				}
				$\theta_d \leftarrow \text{Adam}(\nabla_{\theta_d}\frac{1}{m}\sum_{j=1}^{m}L_d^{(j)}, \theta_d, \alpha, \beta_1, \beta_2)$\;
			}
			\For{$t=1, \dots, n_r$}{
				\For{$j=1, \dots, n_S$}{
					$\tilde{x}_j \leftarrow \theta_s[j]$\;
					$L_r^{(j)} \leftarrow -D(\tilde{x}_j)$\;
				}
				$\theta_s \leftarrow \text{Adam}(\nabla_{\theta_s}\frac{1}{n_S}\sum_{j=1}^{n_S}L_r^{(j)}, \theta_s, \alpha, \beta_1, \beta_2$)
			}
		}
	\end{algorithm}
}

Algorithm \ref{algo} presents the whole training procedure to update the parameters of our \aiprshort{}-CNN model.
At each epoch of this algorithm, the three steps presented above are executed sequentially.
Note that in the second step (lines 10--17), sampling classifier shapelets, as well as sampling subseries from the training set, is performed uniformly at random.

%% file: sections/xp.tex
\section{Experiments}
\label{sec:xp}

In this section, we will detail the training procedure for the \aiprnn{} and present both quantitative and qualitative experimental results.

\subsection{Experimental Setting}

\begin{figure*}[ht!]
	\subfloat[Wasserstein loss $L_d$]{
                \label{fig:shapelet_evolve:wass}
		\includegraphics[width=.33\linewidth]{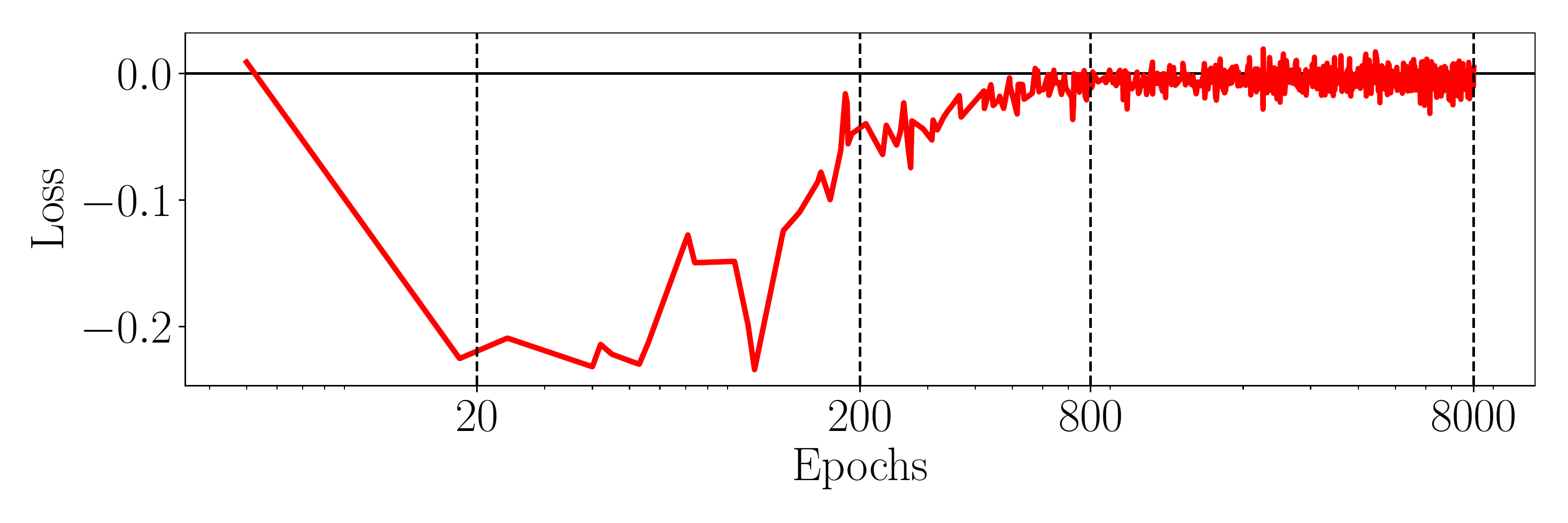}
	}
	\subfloat[Shapelet at epoch $20$]{
		\includegraphics[width=.33\linewidth]{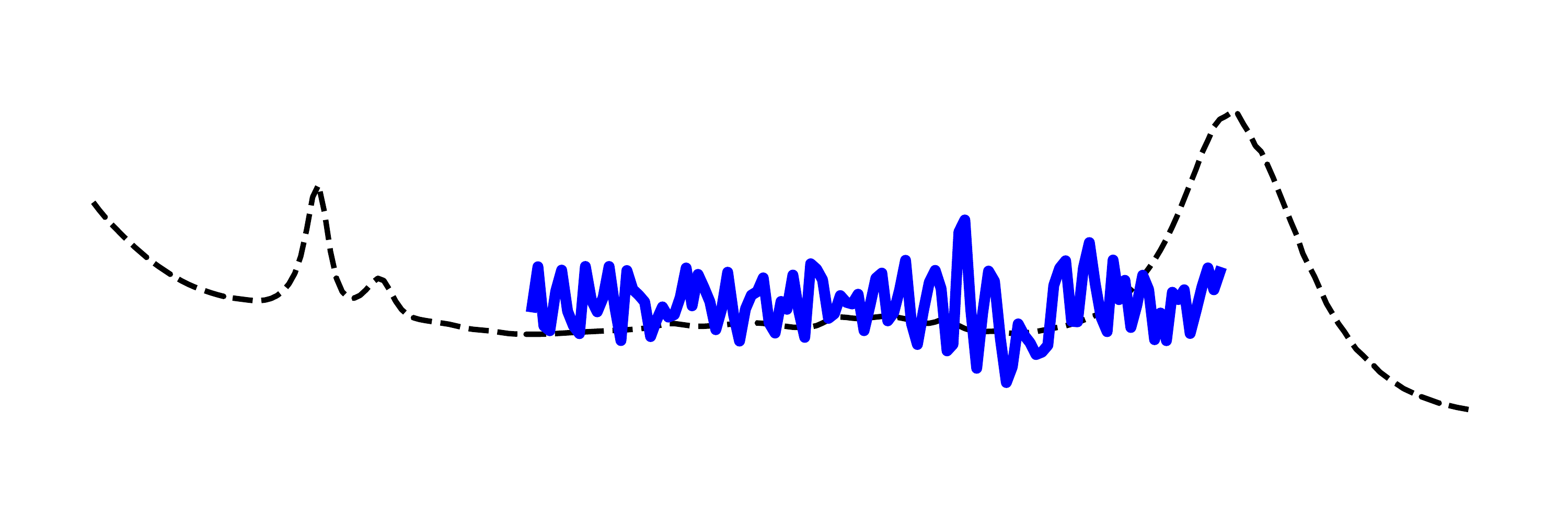}
	}
	\subfloat[Shapelet at epoch $200$]{
		\includegraphics[width=.33\linewidth]{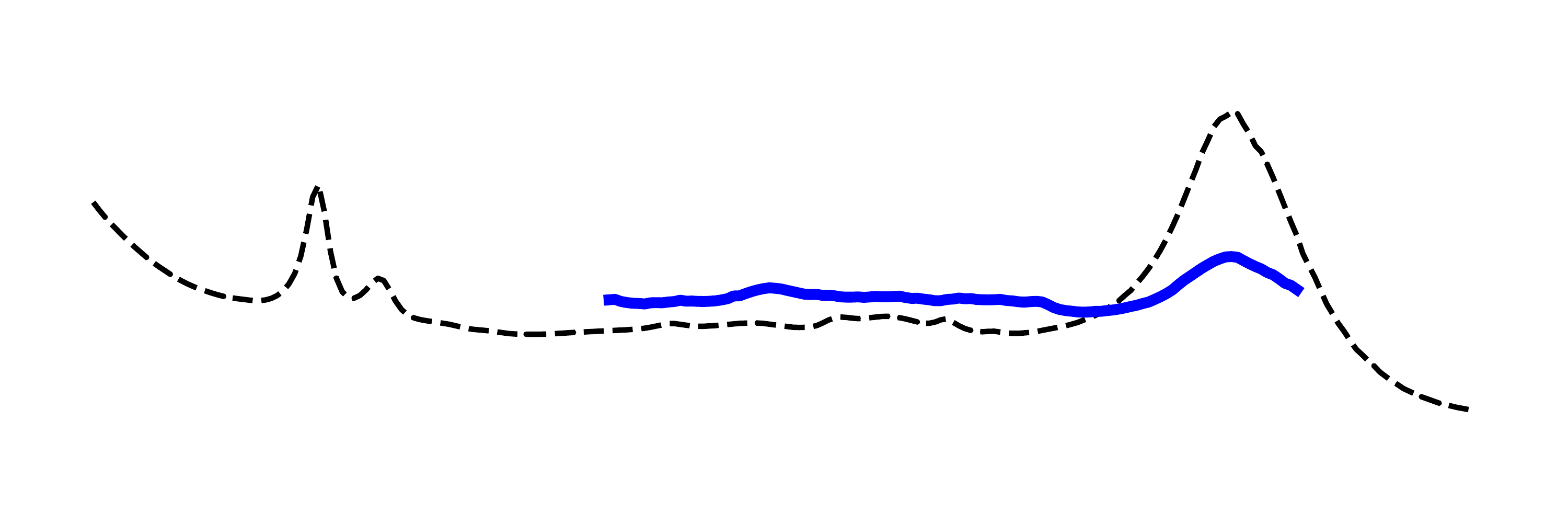}
	}\\
	\subfloat[Cross-entropy loss $L_c$]{
                \label{fig:shapelet_evolve:xent}
		\includegraphics[width=.33\linewidth]{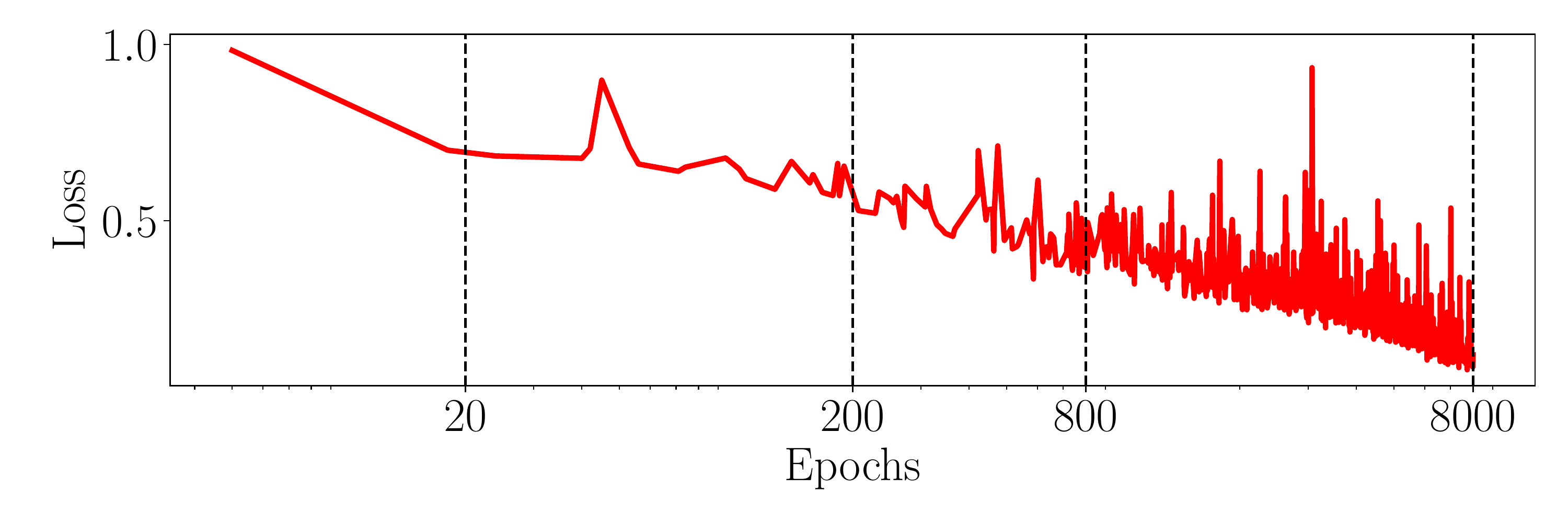}
	}
	\subfloat[Shapelet at epoch $800$]{
		\includegraphics[width=.33\linewidth]{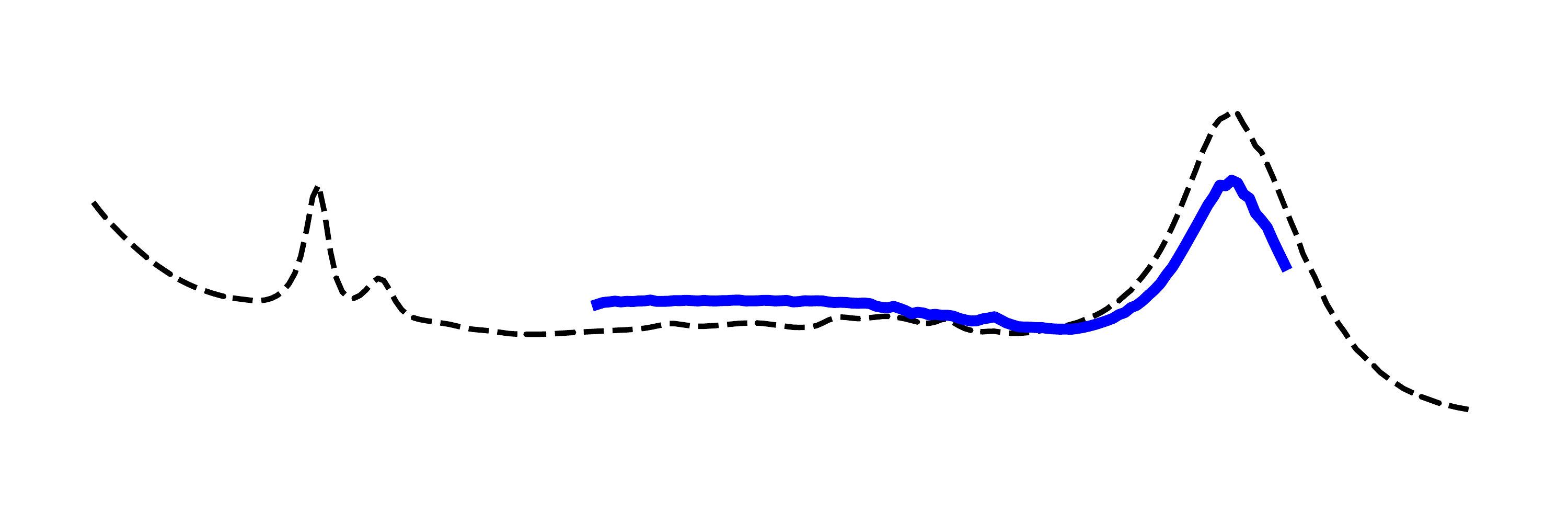}
	}
	\subfloat[Shapelet at epoch $8000$]{
		\includegraphics[width=.33\linewidth]{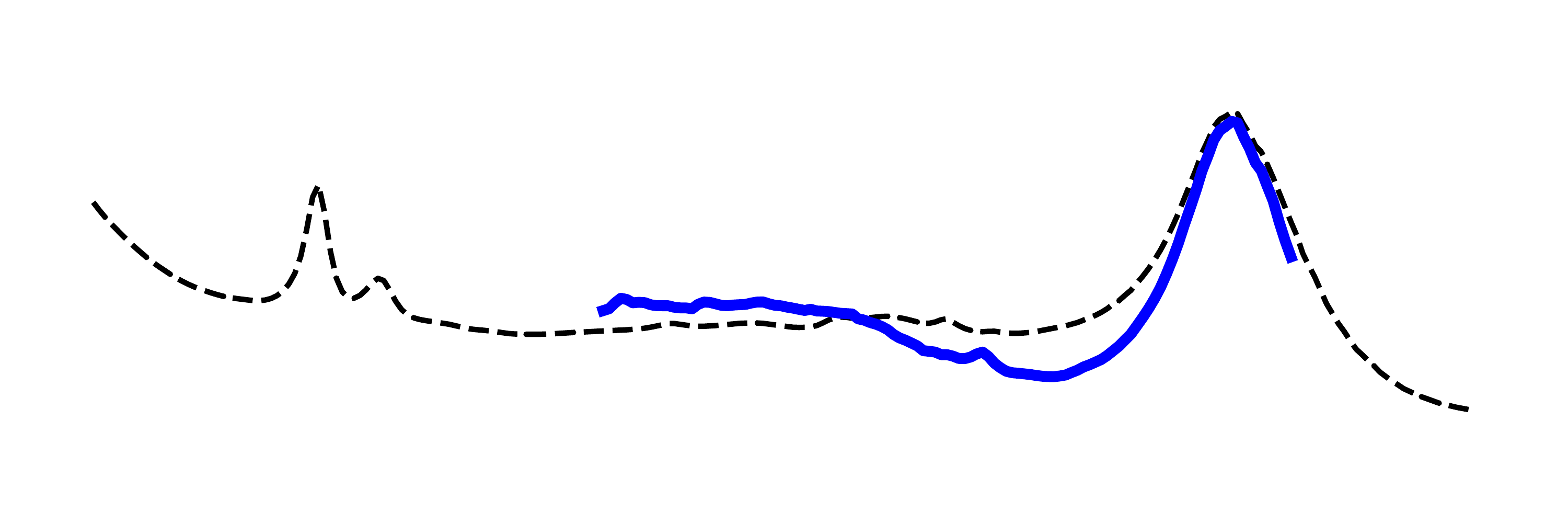}
	}
	\caption{Illustration of the evolution of a shapelet during training (for the Wine dataset). \label{fig:shapelet_evolve}}
\end{figure*}

As explained in Section \ref{sec:related}, our most relevant competitor is Learning Shapelets (LS) from \cite{Grabocka2014} as it also describes a shapelet-based model where the shapelets are learned and where a single model is used for classification. % (so it has a similar complexity as ours). 
In the following sections, all the results presented for LS are retrieved from the UCR/UEA repository~\cite{UCRArchive} and the shapelets presented for LS are obtained using the \emph{tslearn} implementation~\cite{tslearn}.

\subsubsection{Datasets}
To compare our proposed method with~\cite{ye2009time, Rak13, Grabocka2014}, we use the 85 univariate time series datasets from the UCR/UEA repository for which all the baselines are available~\cite{UCRArchive}.\footnote{See \url{http://www.timeseriesclassification.com/singleTrainTest.csv} for all used datasets and baseline results.}
Note that our CNN-based method is not, by design, limited to univariate time series. However, for a fair comparison, we limited ourself to these datasets for this study. The datasets are significantly different from one to another, including seven types of data with various number of instances, lengths, and classes.
The splits between training and test sets are provided in the repository. %Since the datasets are relatively small, we did not use any validation set but we set some of our hyper-parameters by cross-validation (others were chosen to be comparable with~\cite{Grabocka2014} and are fixed for all the used datasets). Before the data manipulation, the datasets are loaded and normalized with tslearn~\cite{tslearn}. 

\subsubsection{Architecture details and parameter setting}
We have implemented the \aiprnn{} model using TensorFlow~\cite{tensorflow} following the general architecture illustrated in Figure~\ref{fig:arn}.
The classifier is composed of one 1D convolution layer with ReLU activation, followed by a maxpooling layer along the temporal dimension and a fully connected layer with a softmax activation.
The shapelets use a Glorot uniform initializer~\cite{Glorot2010} while the other weights are initialized uniformly (using a fixed range).
For each dataset, three different shapelet lengths are considered, inspired by the heuristic from~\cite{Grabocka2014} but without resorting to hyper-parameter search: we consider 3 groups of $20 \times n_\text{classes}$ shapelets of length $0.2 T$, $0.4 T$ and $0.6 T$, where $n_\text{classes}$ is the number of classes in the dataset and $T$ is the length of the time series at stake.

%For the length of the shapelets, we used the heuristic of Grabocka~\cite{Grabocka2014}, but not exactly the same. Grabocka did a hyperparameter search for different dataset, that the length of the shapelets varies in \{$0.2, 0.4, 0.6$\} times length of the time series, and the number of shapelets was given by$$\log(TS_{length}-S_{length}+1)\times(ClassNumber-1).$$ But during our test, we found out that if we choose the number of each shapelet as $ 20 \times ClassNumber $, the accuracy will be better.

The convolution filters of the classifier, i.e. the shapelets, are given as input to the discriminator which has the same structure as the classifier, but with shorter convolution filters (100 filters of size $0.06T$, $0.12T$ and $0.18T$)  and a single-neuron $tanh$ activation instead of the softmax in the last layer. % (since it has a single output).
For optimization, we use Adam optimizer with a standard parametrization ($\alpha=10^{-3}$, $\beta_1=0.9$ and $\beta_2 = 0.999$) and each epoch consists in $n_c=15$ (resp. $n_d=20$ and $n_r=17$) mini-batches of optimization for the classifier loss (resp. discriminator and regularizer losses).

Experimental results are reported in terms of test accuracy and aggregated over five random initializations.
All experiments are run for 8000 training epochs.
The authors are devoted to the reproduciblility of the results.%\footnote{All the code related to this work will be made publicly available if this paper is accepted.}

\begin{figure*}[ht!]
	\centering
	\begin{minipage}[b]{.45\linewidth}
		\centering
		\includegraphics[width=.33\linewidth,height=2.1cm]{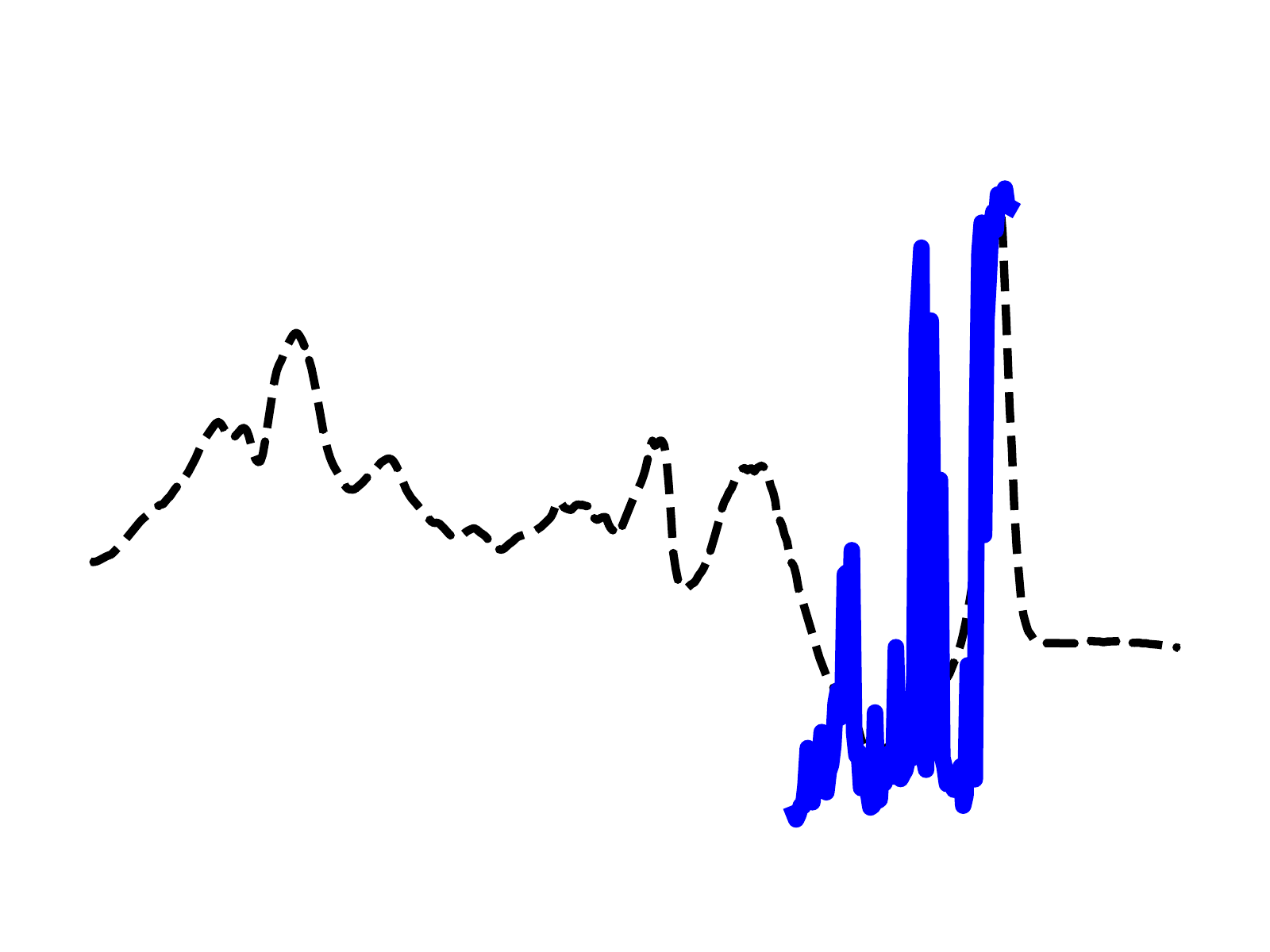}%
		\includegraphics[width=.33\linewidth,height=2.1cm]{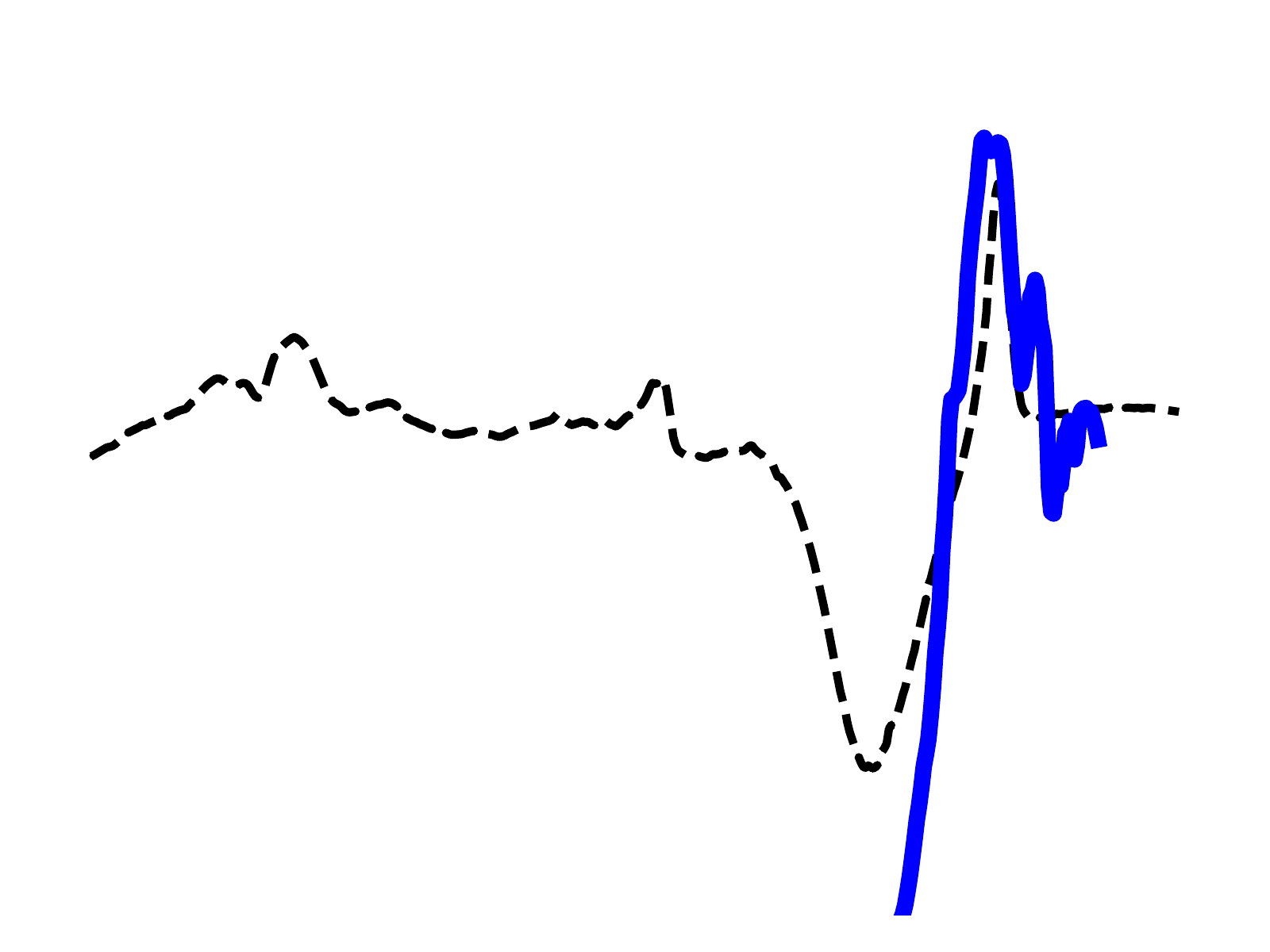}%
		\includegraphics[width=.33\linewidth,height=2.1cm]{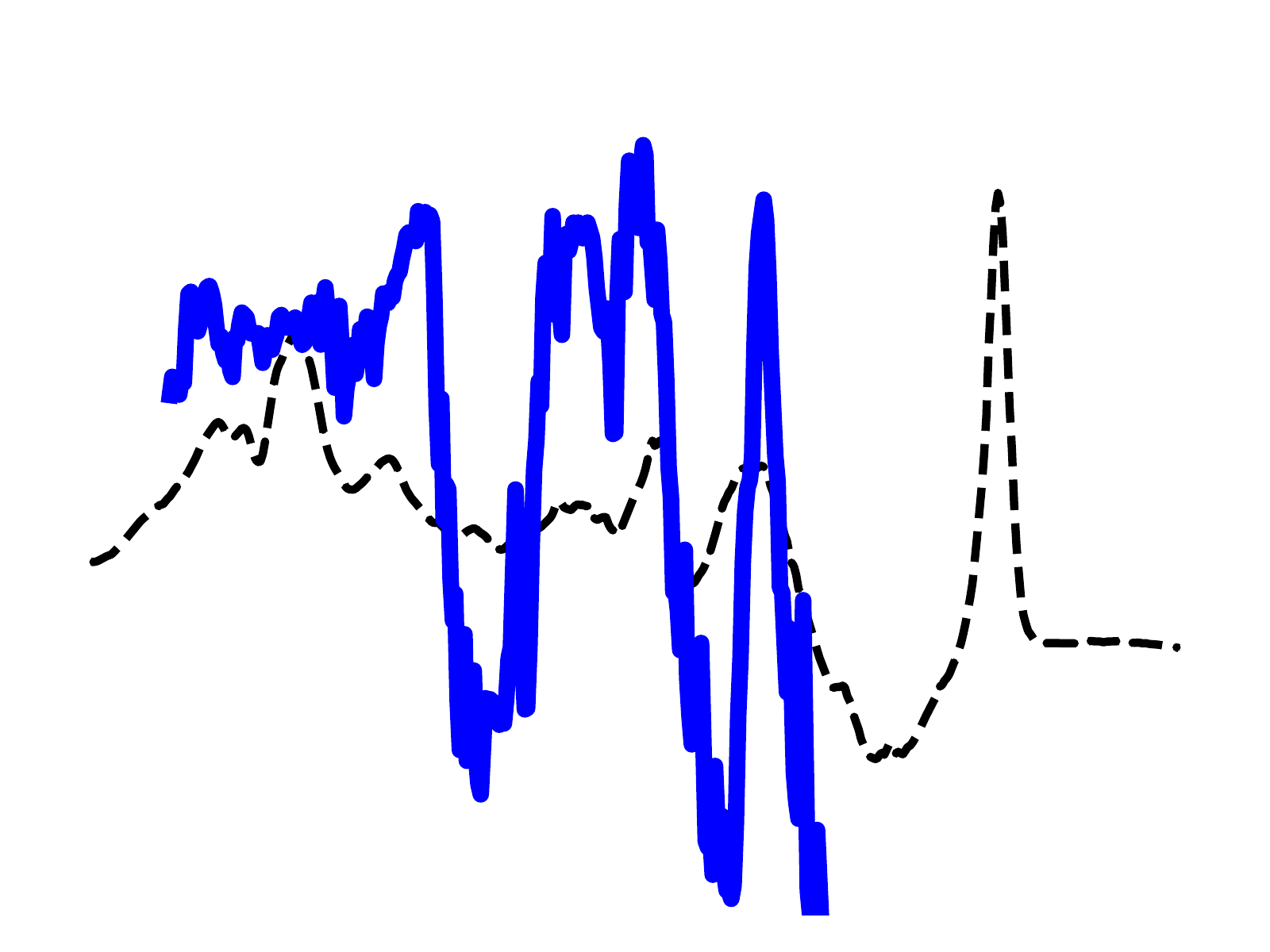}
		\includegraphics[width=.33\linewidth,height=2.1cm]{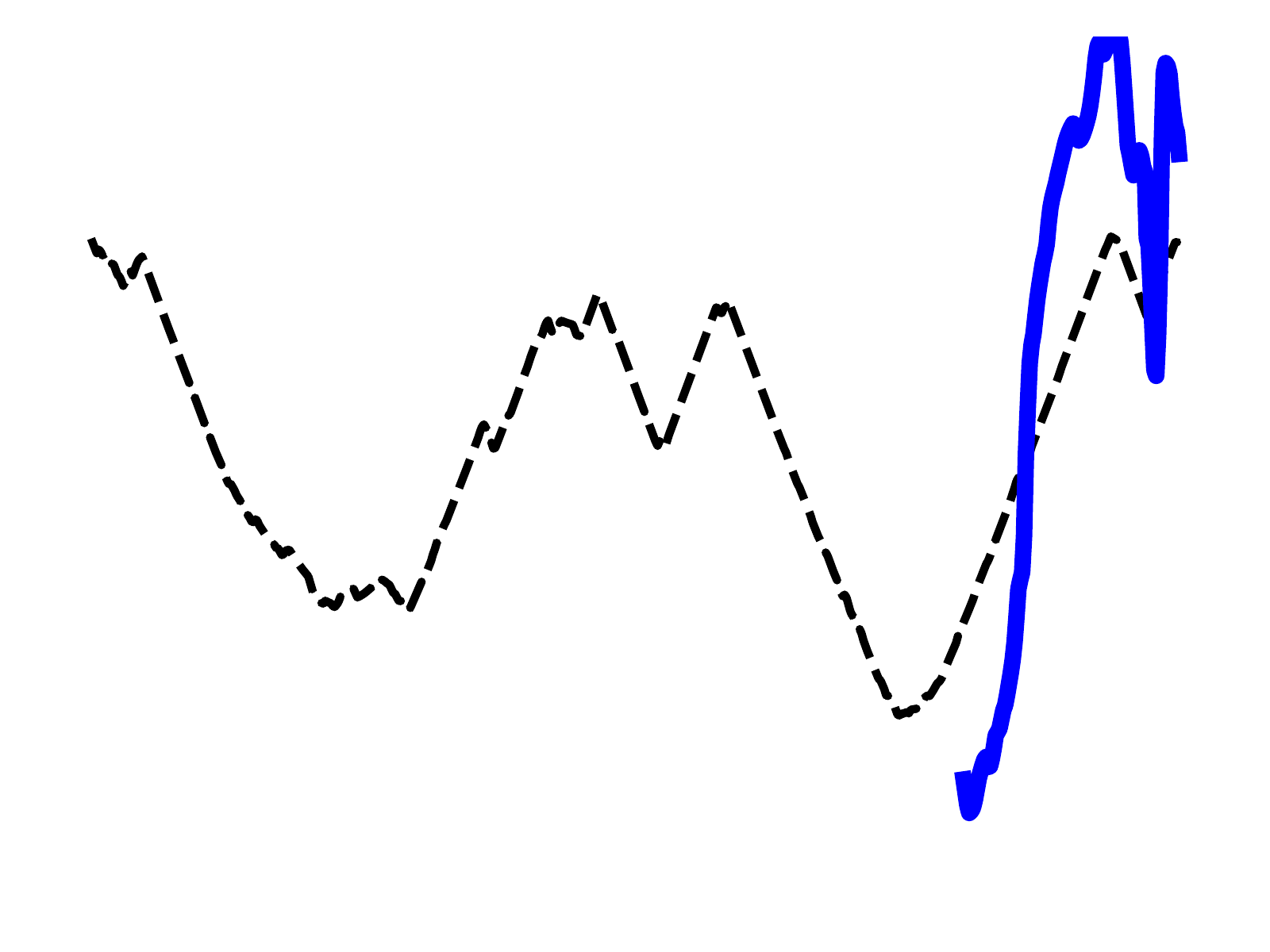}%
		\includegraphics[width=.33\linewidth,height=2.1cm]{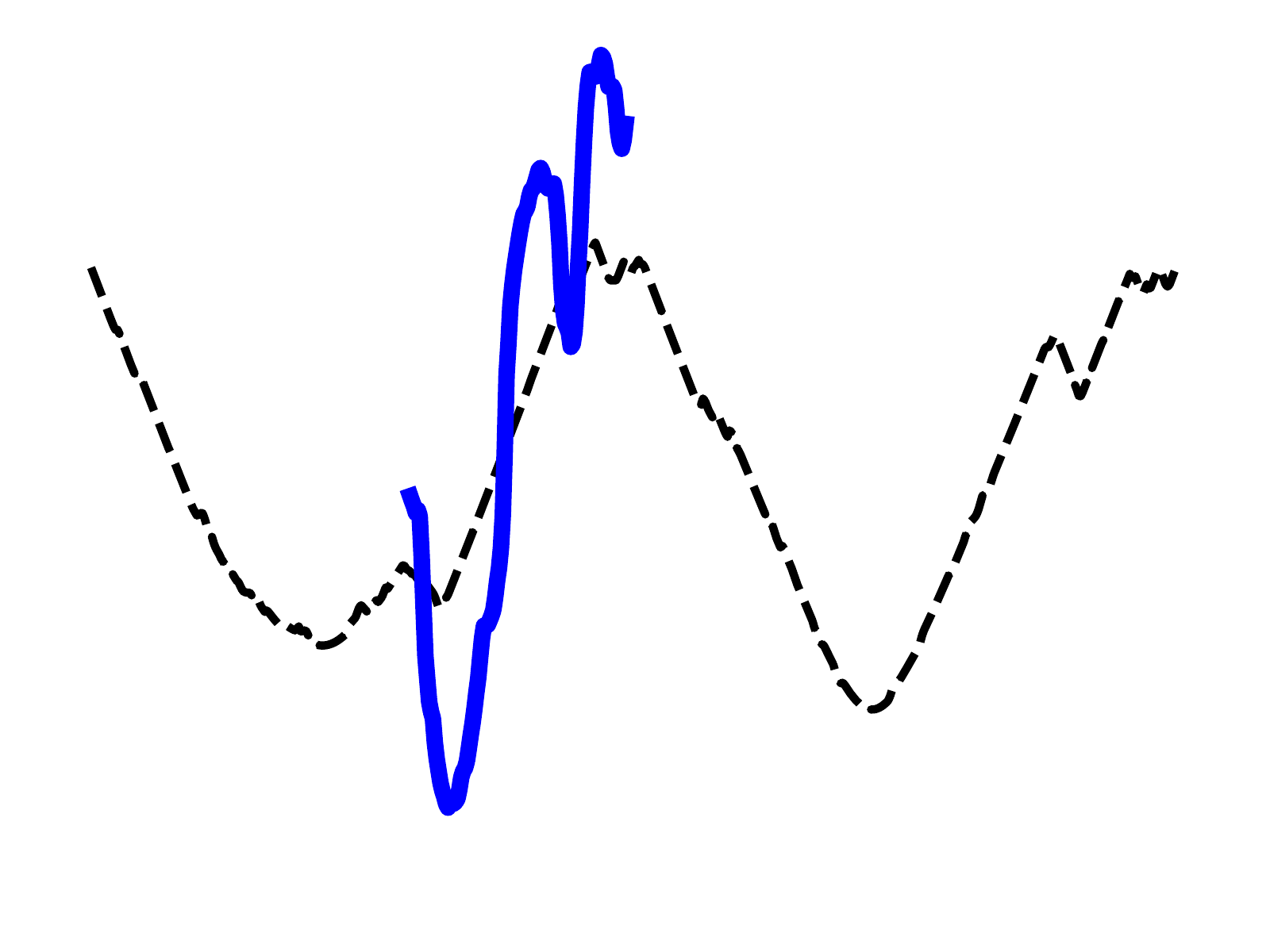}%
		\includegraphics[width=.33\linewidth,height=2.1cm]{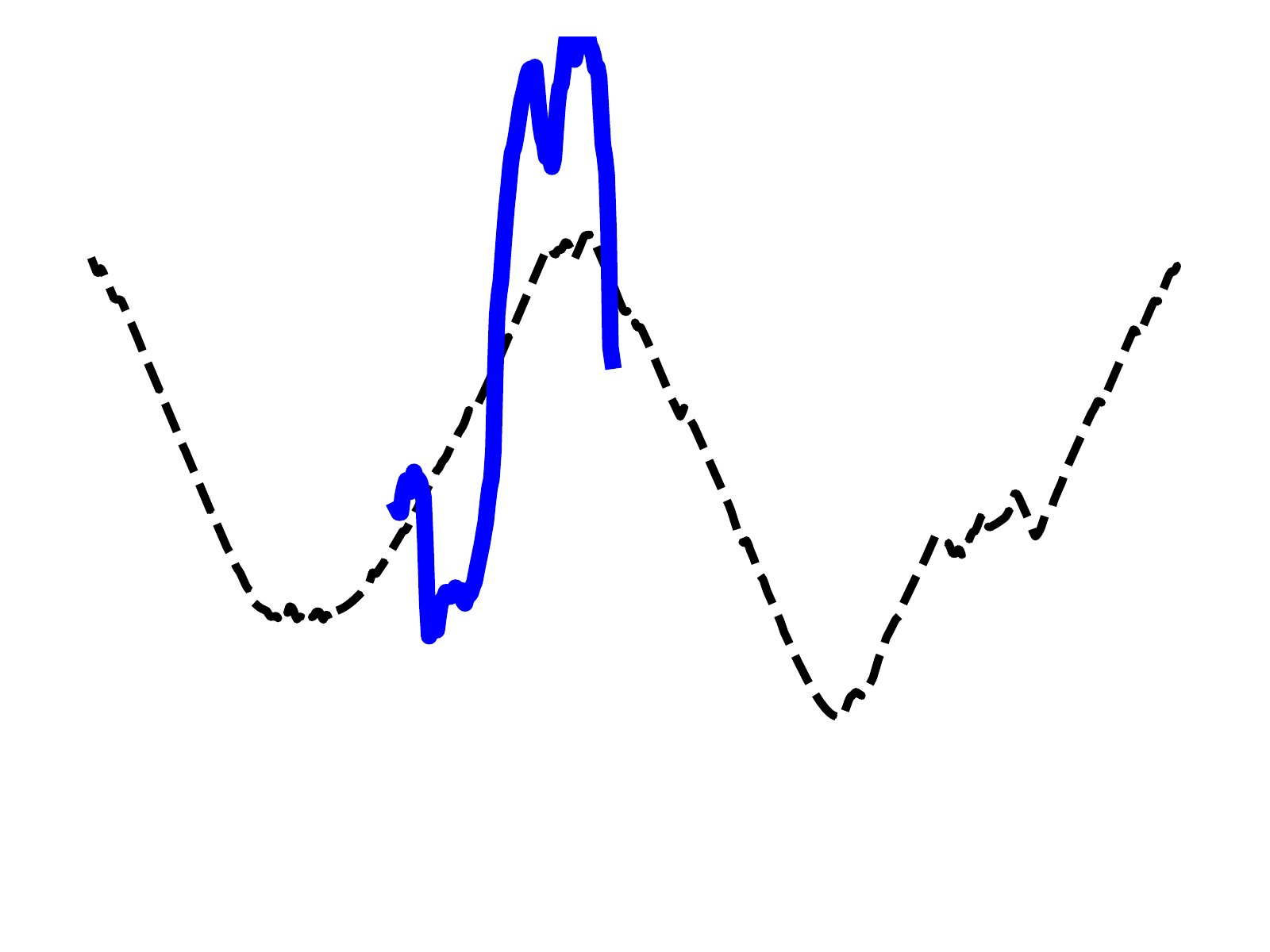}
		\includegraphics[width=.33\linewidth,height=2.1cm]{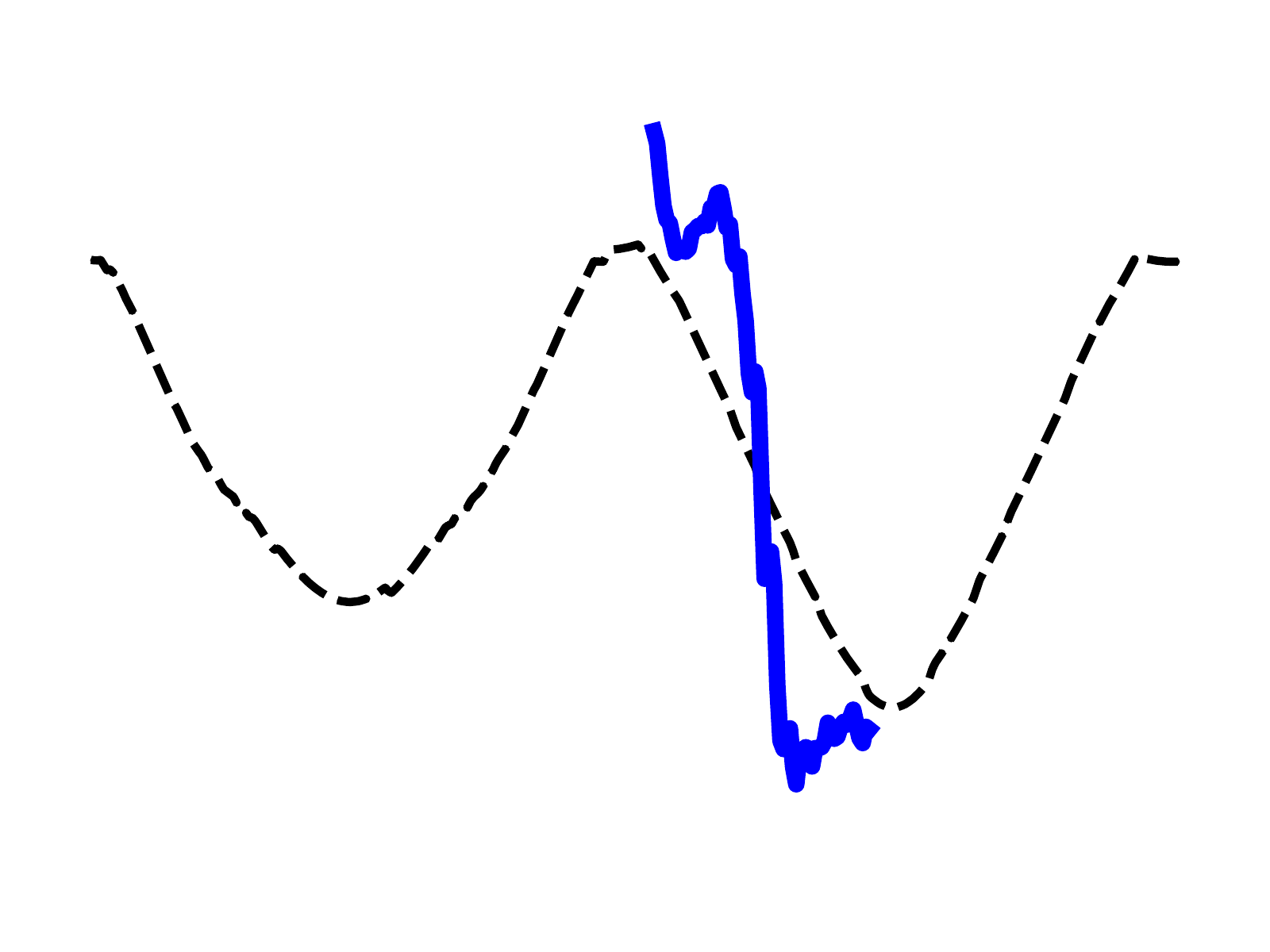}%
		\includegraphics[width=.33\linewidth,height=2.1cm]{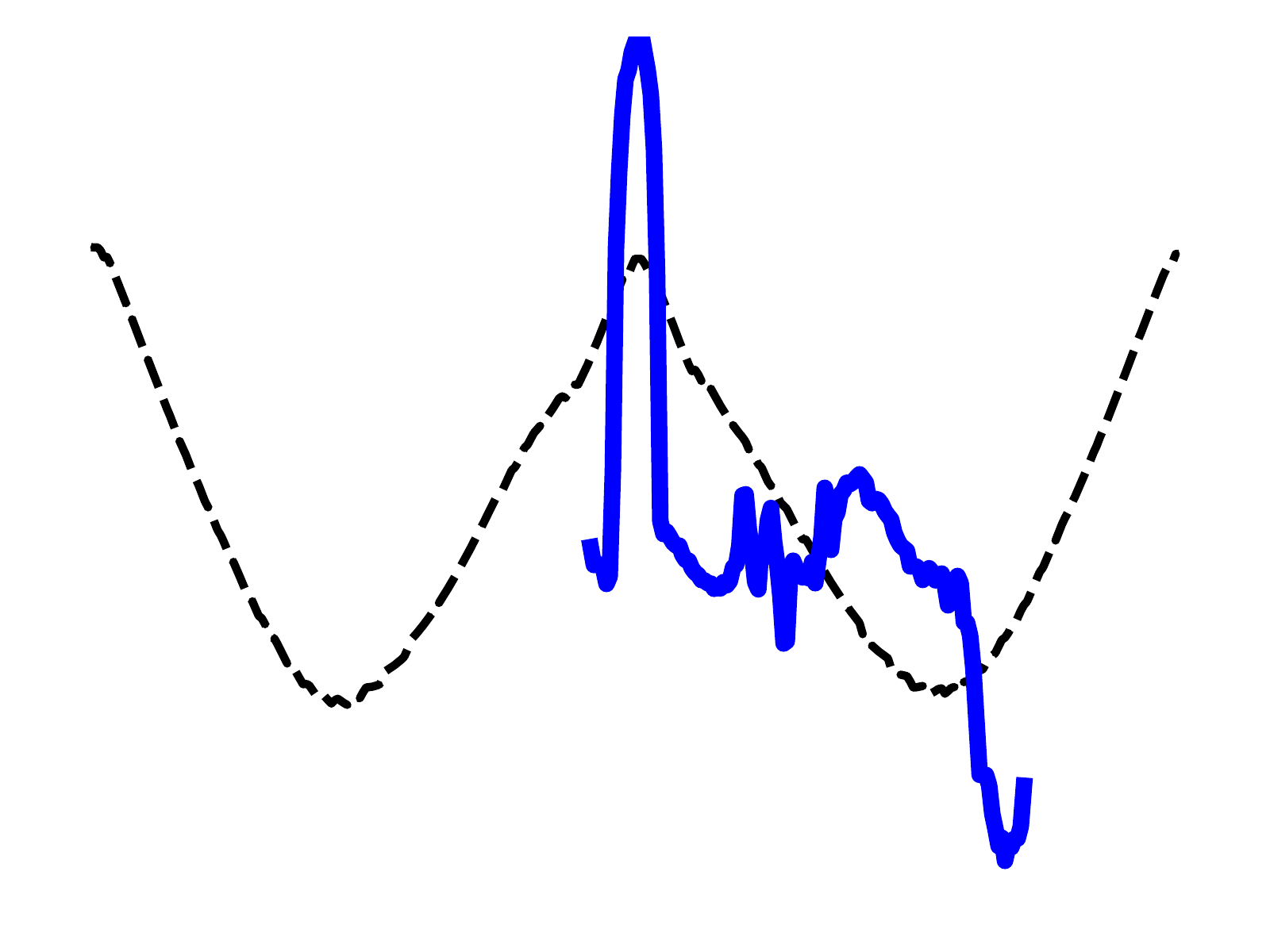}%
		\includegraphics[width=.33\linewidth,height=2.1cm]{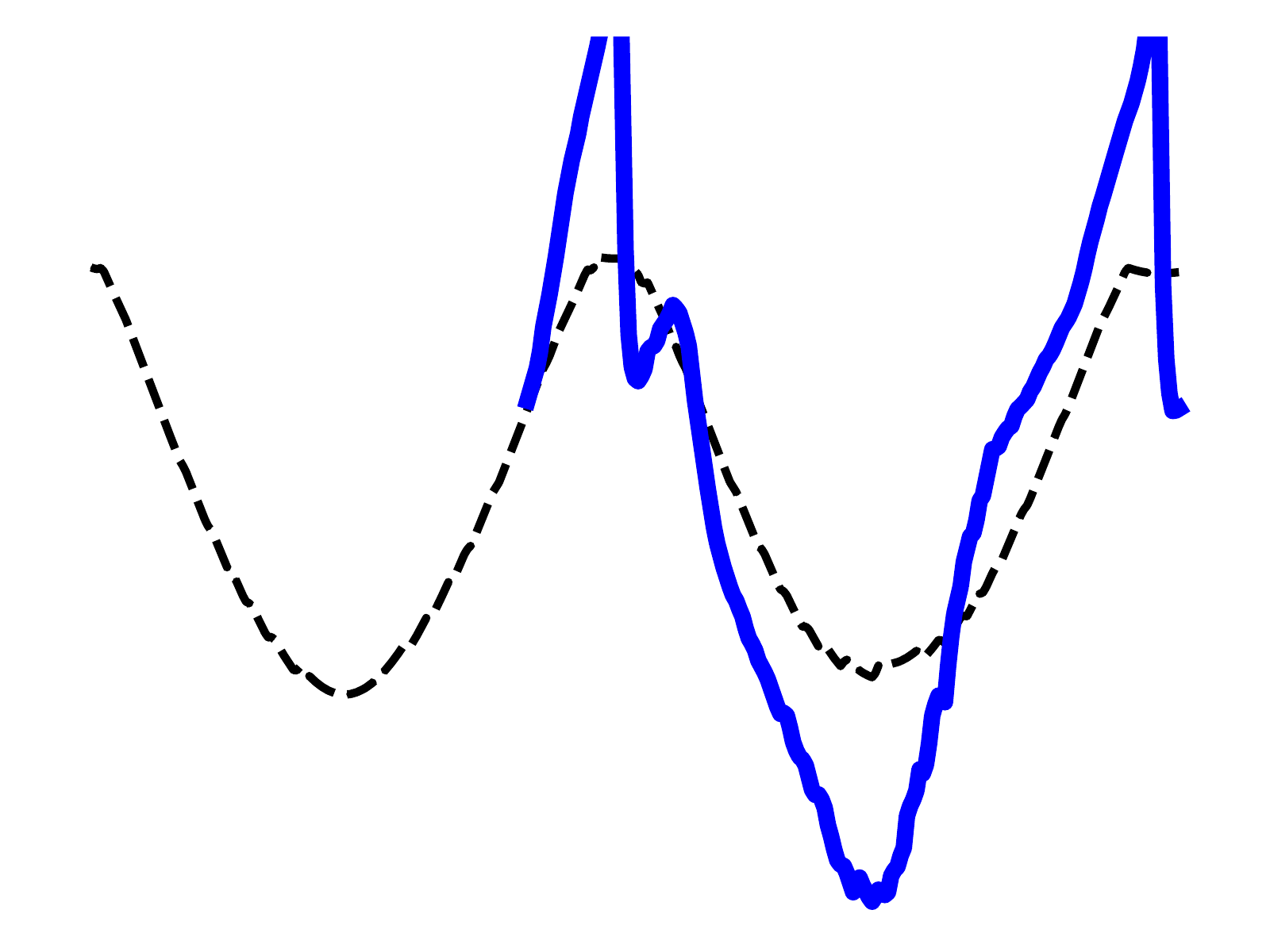}
		\includegraphics[width=.33\linewidth]{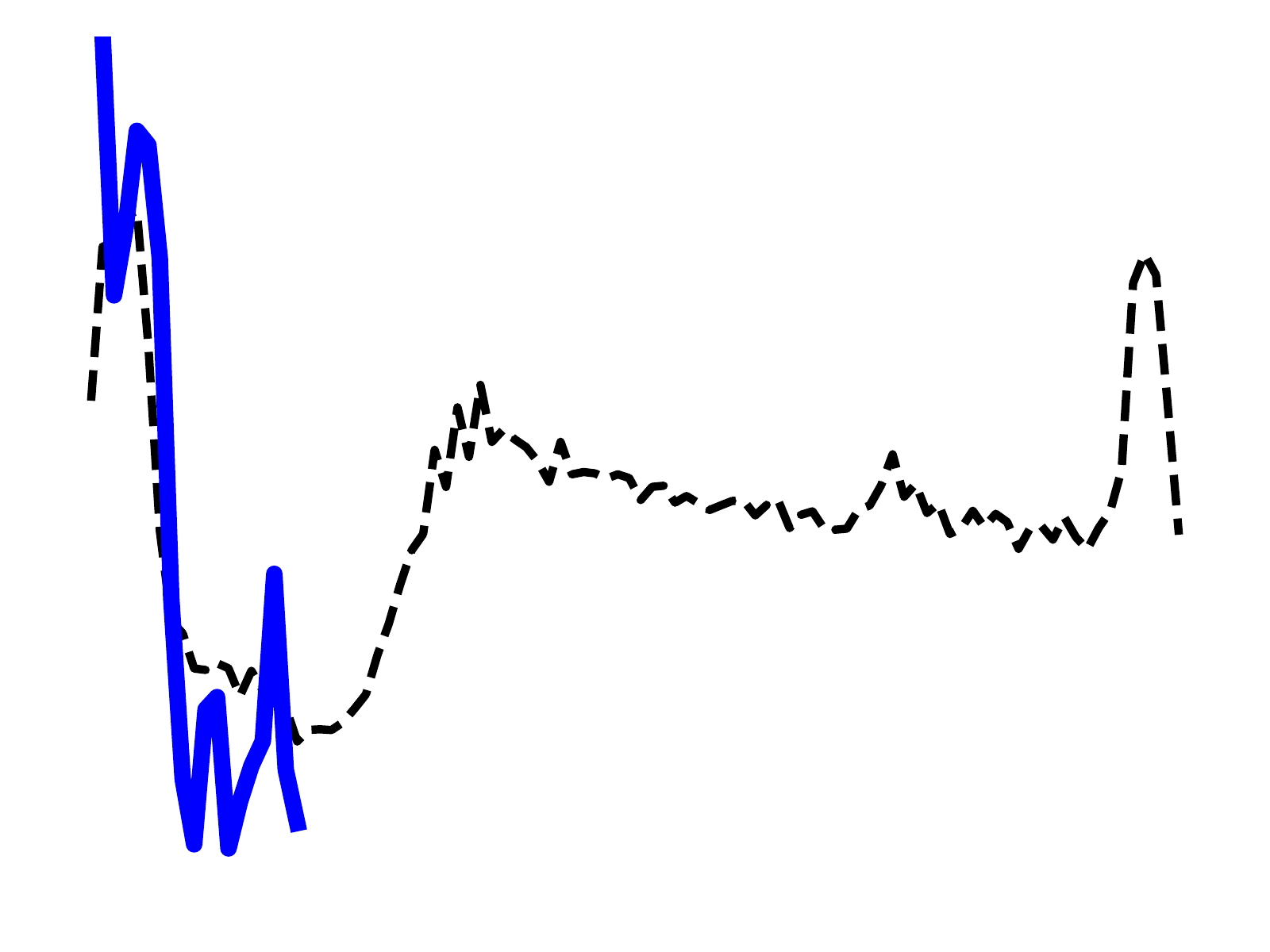}%
		\includegraphics[width=.33\linewidth]{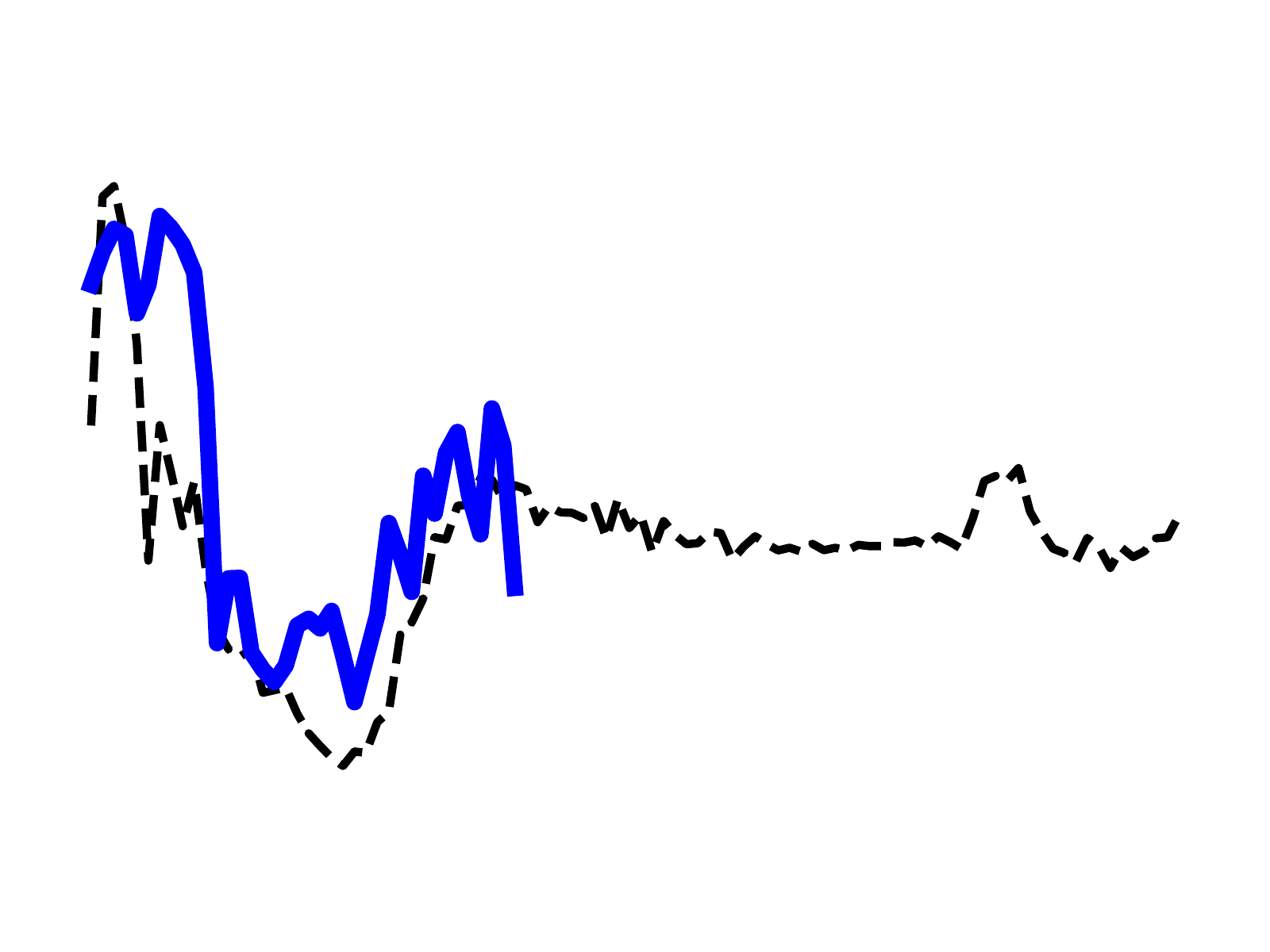}%
		\includegraphics[width=.33\linewidth]{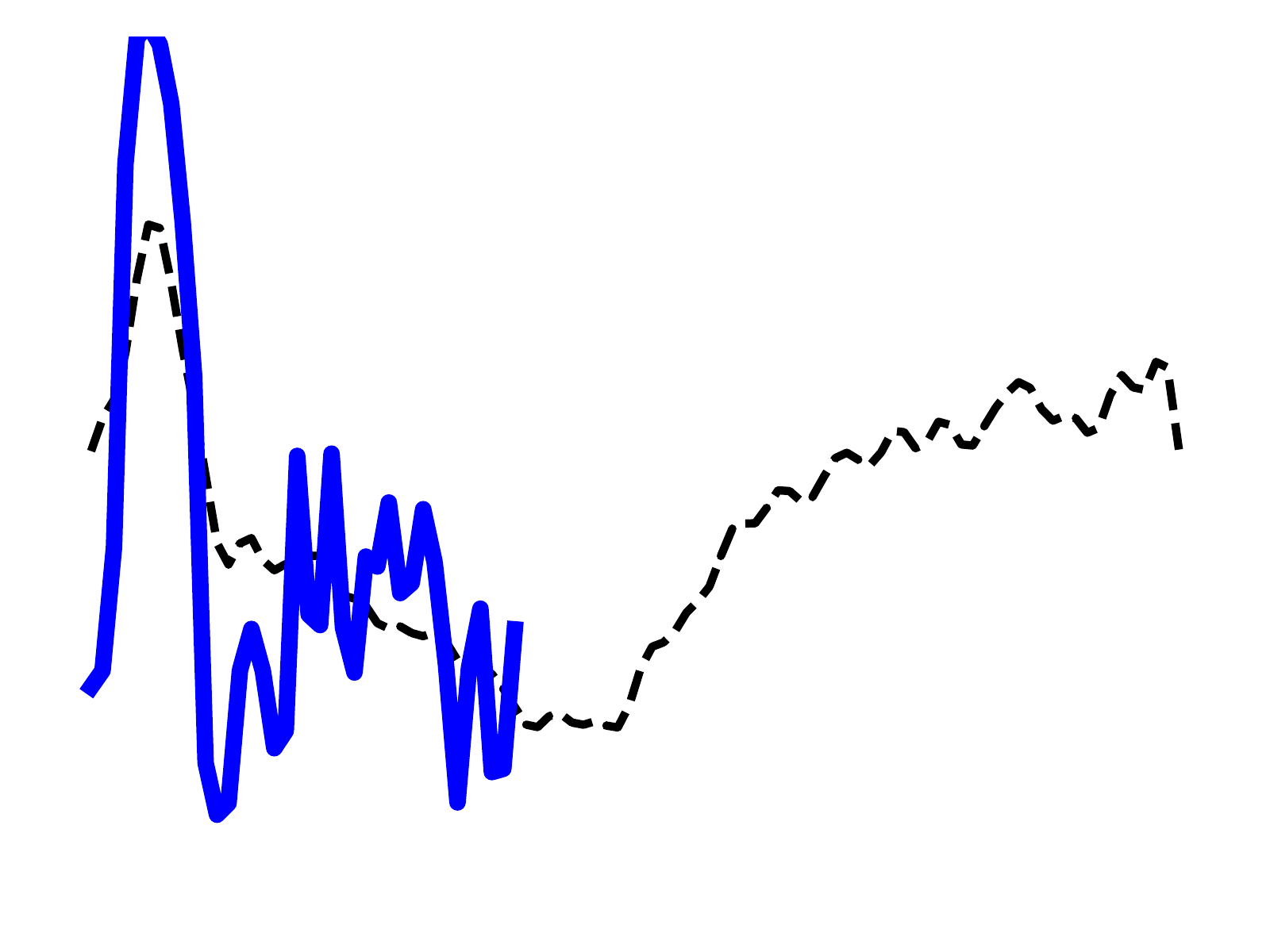}
		\includegraphics[width=.33\linewidth]{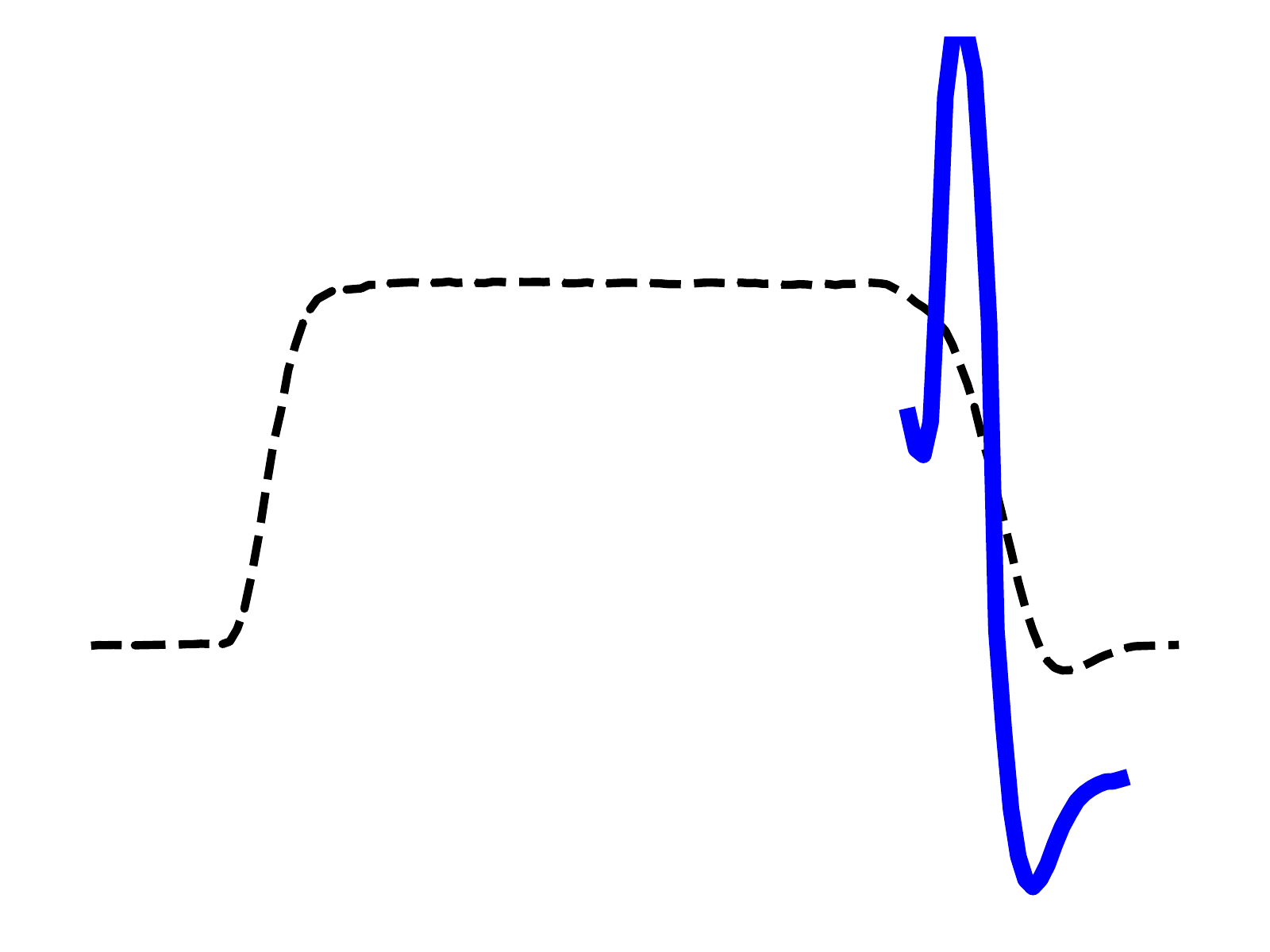}%
		\includegraphics[width=.33\linewidth]{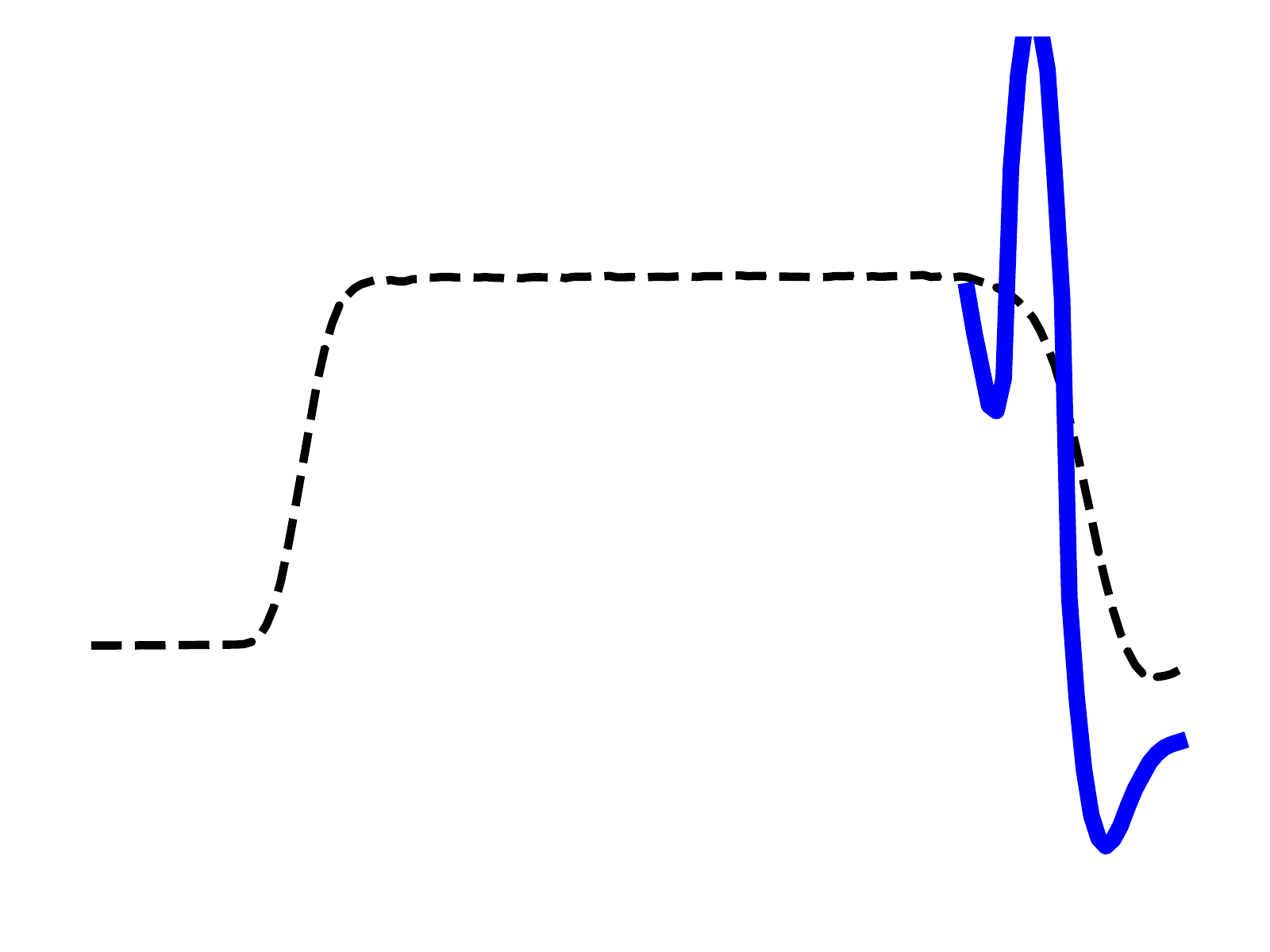}%
		\includegraphics[width=.33\linewidth]{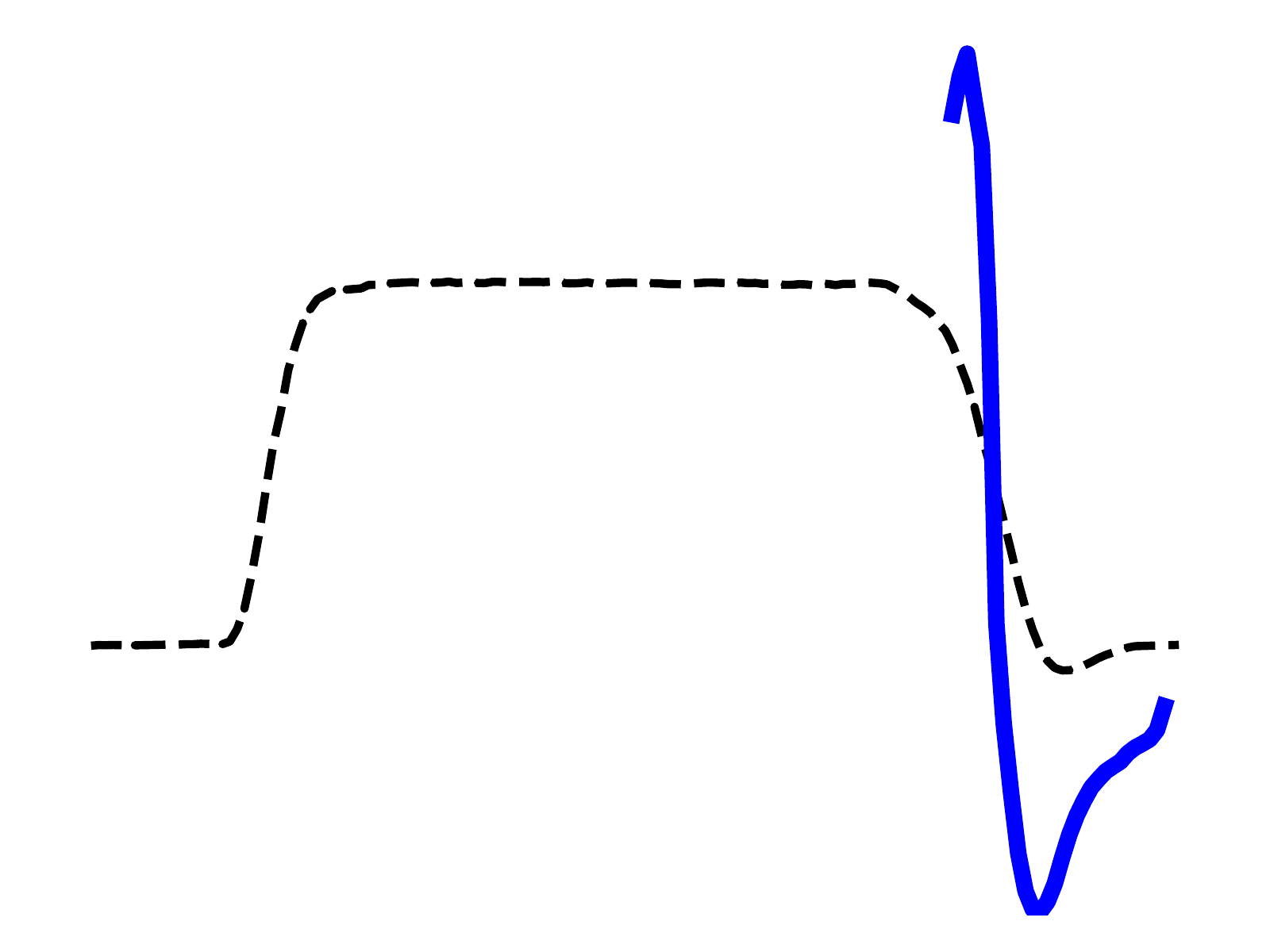}
		\includegraphics[width=.33\linewidth]{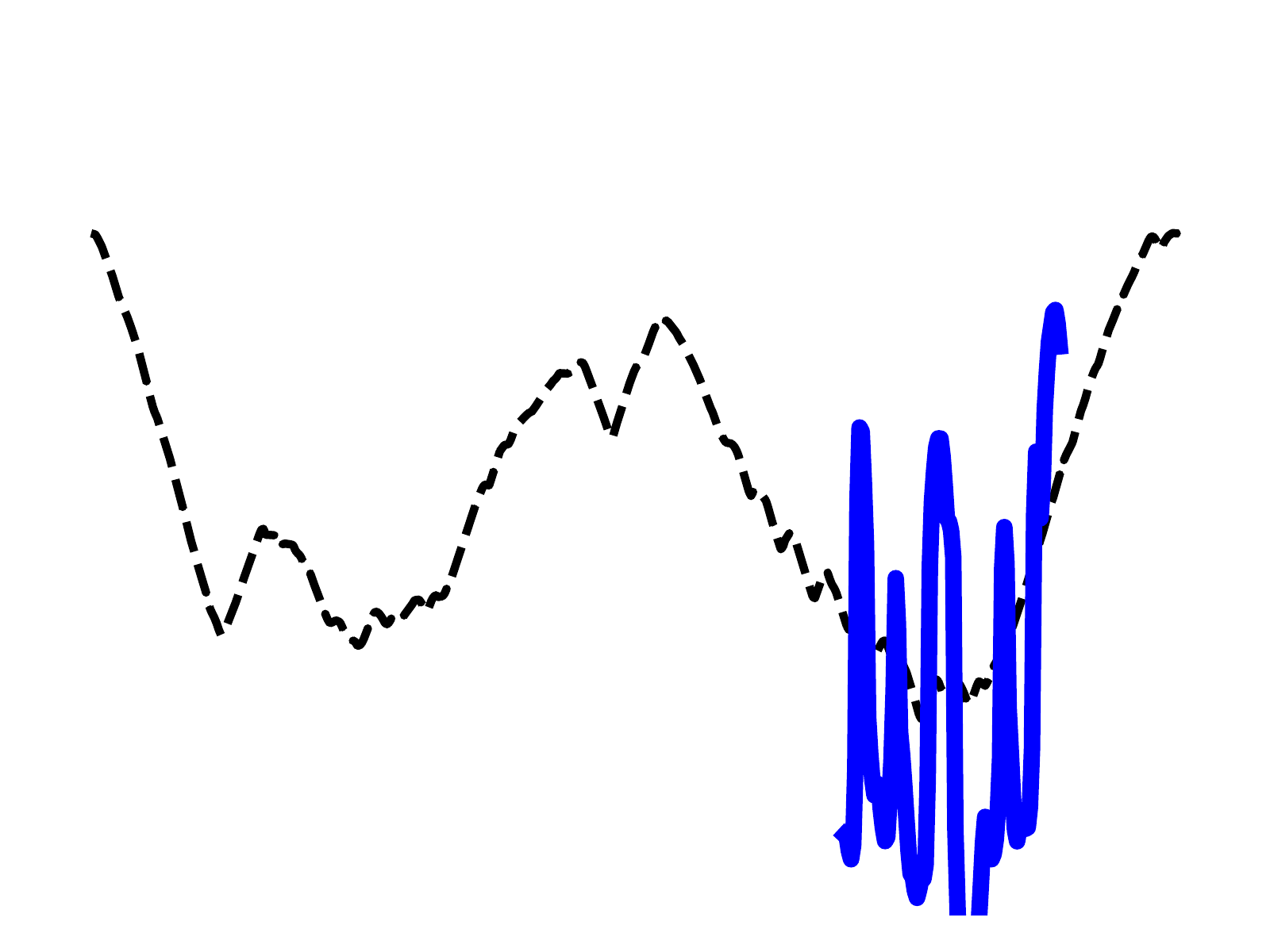}%
		\includegraphics[width=.33\linewidth]{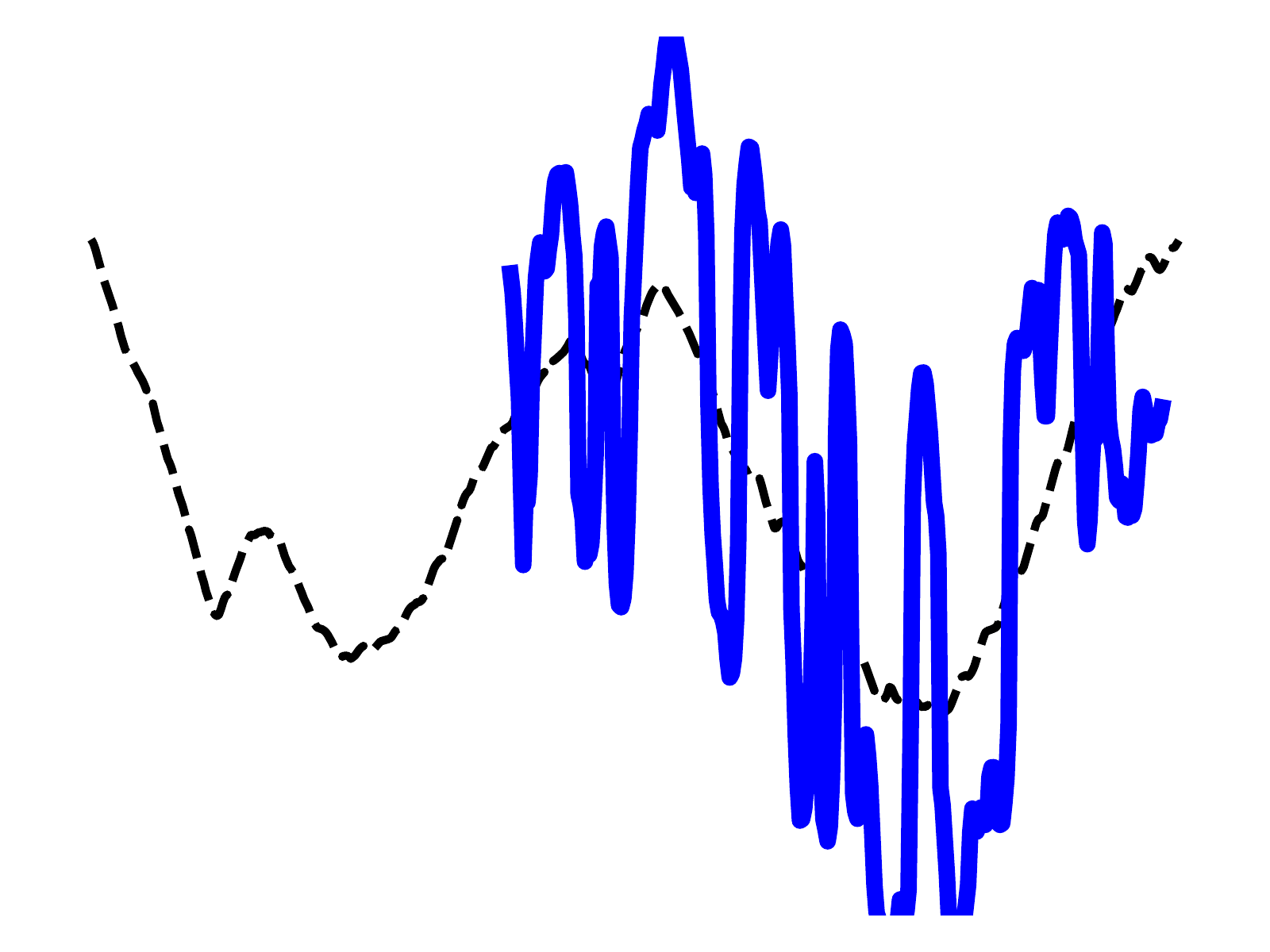}%
		\includegraphics[width=.33\linewidth]{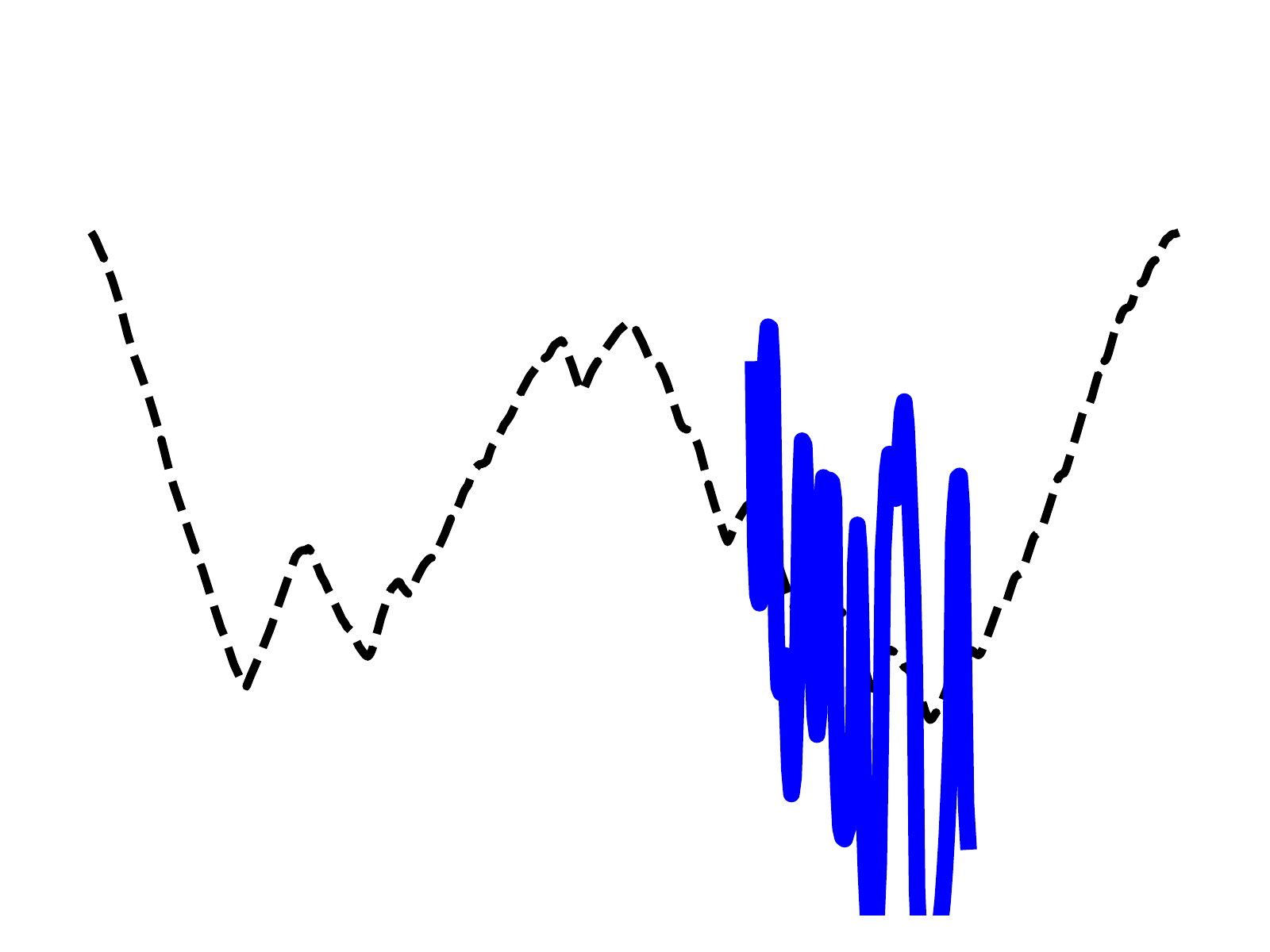}
		\includegraphics[width=.33\linewidth,height=2.1cm]{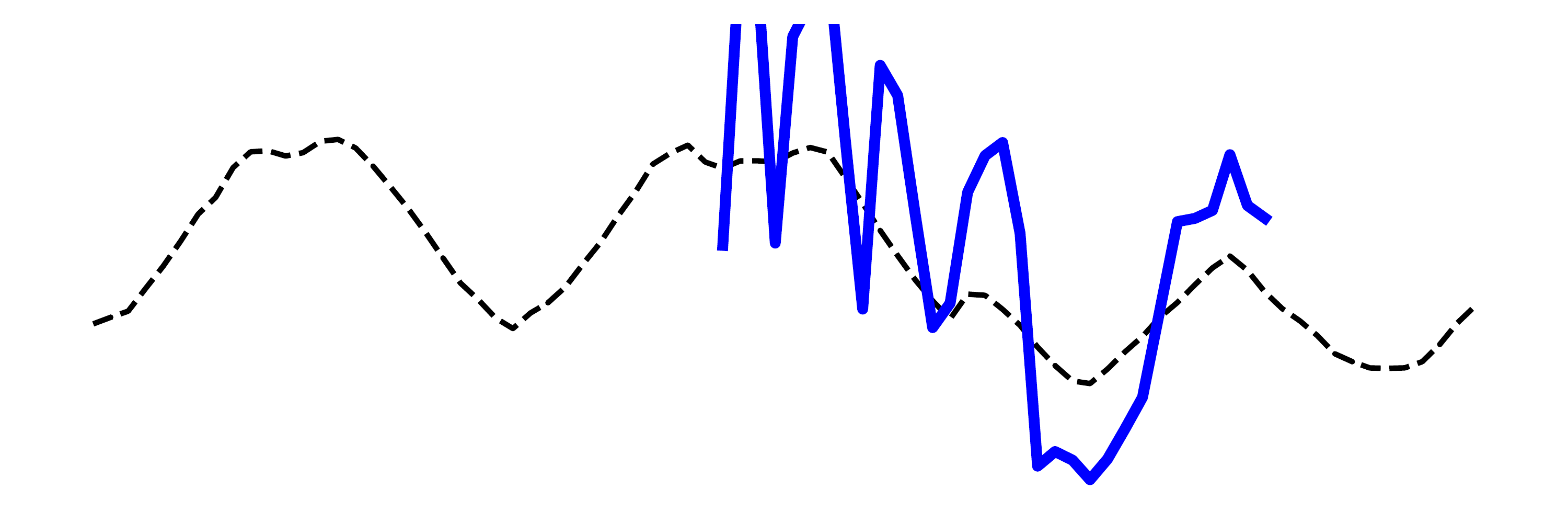}%
		\includegraphics[width=.33\linewidth,height=2.1cm]{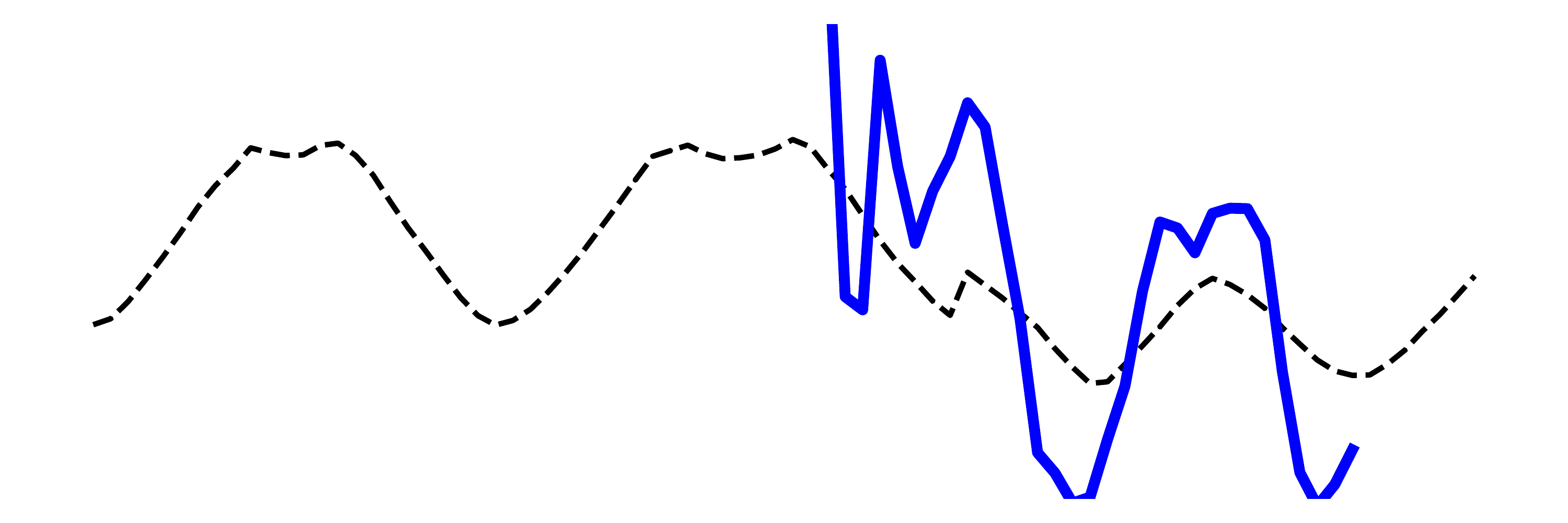}%
		\includegraphics[width=.33\linewidth,height=2.1cm]{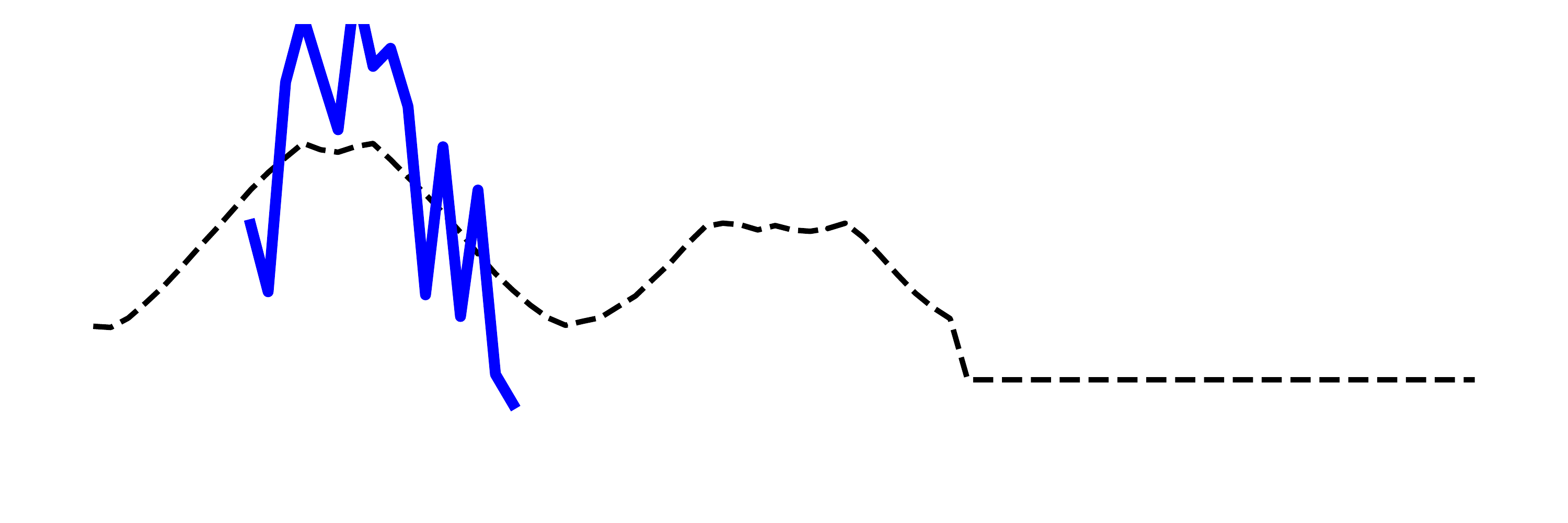}
		\includegraphics[width=.33\linewidth]{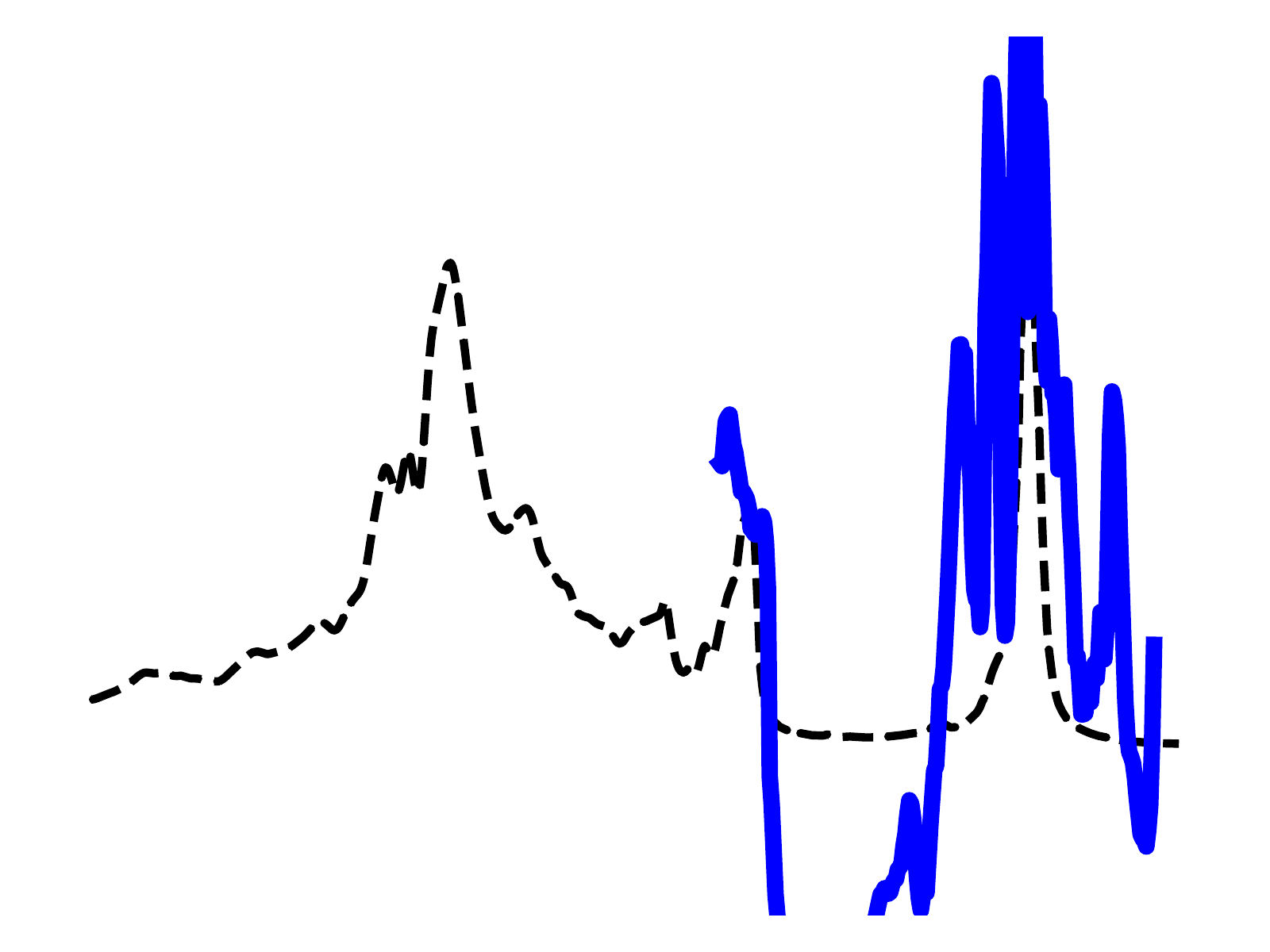}%
		\includegraphics[width=.33\linewidth]{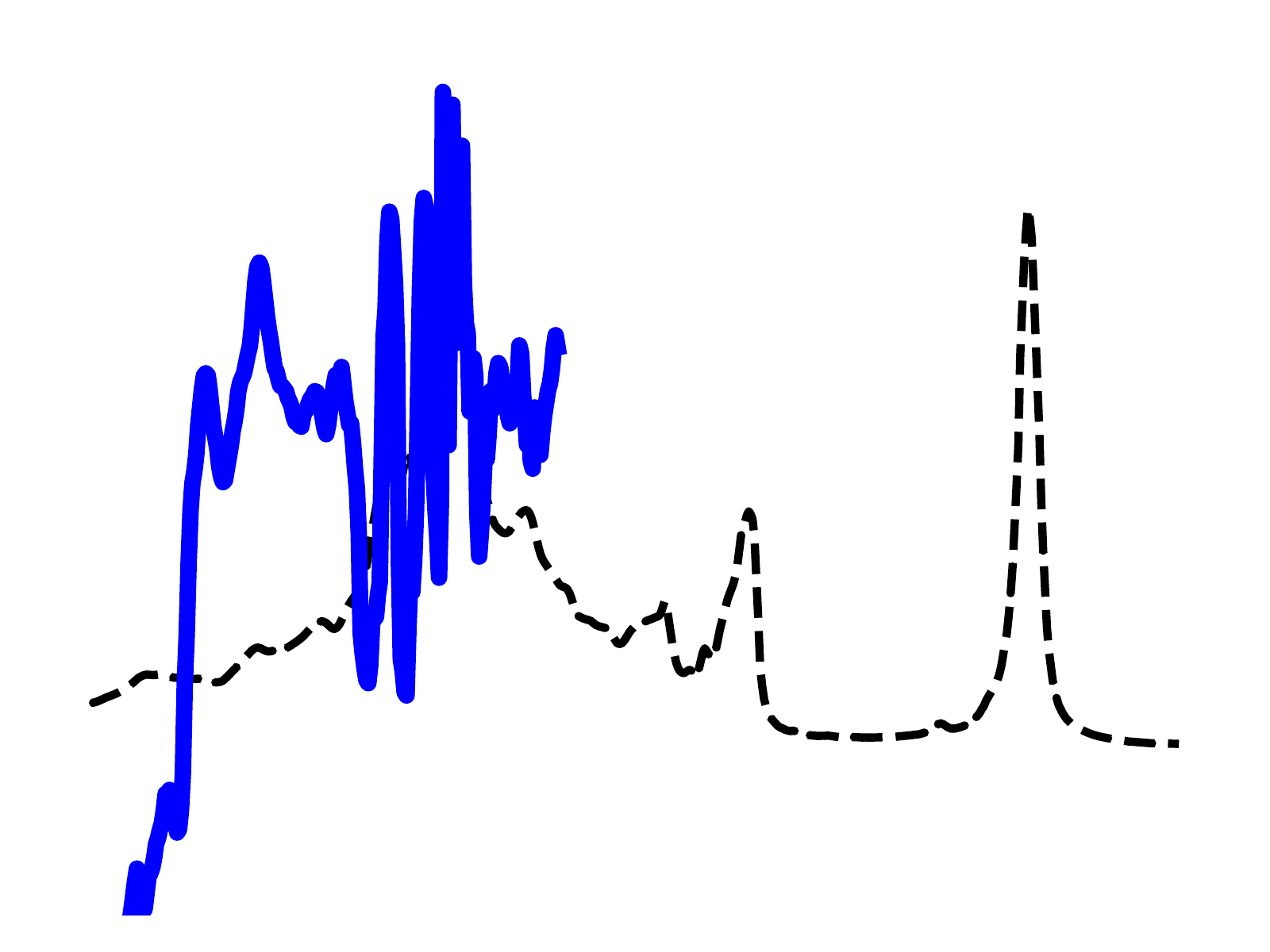}%
		\includegraphics[width=.33\linewidth]{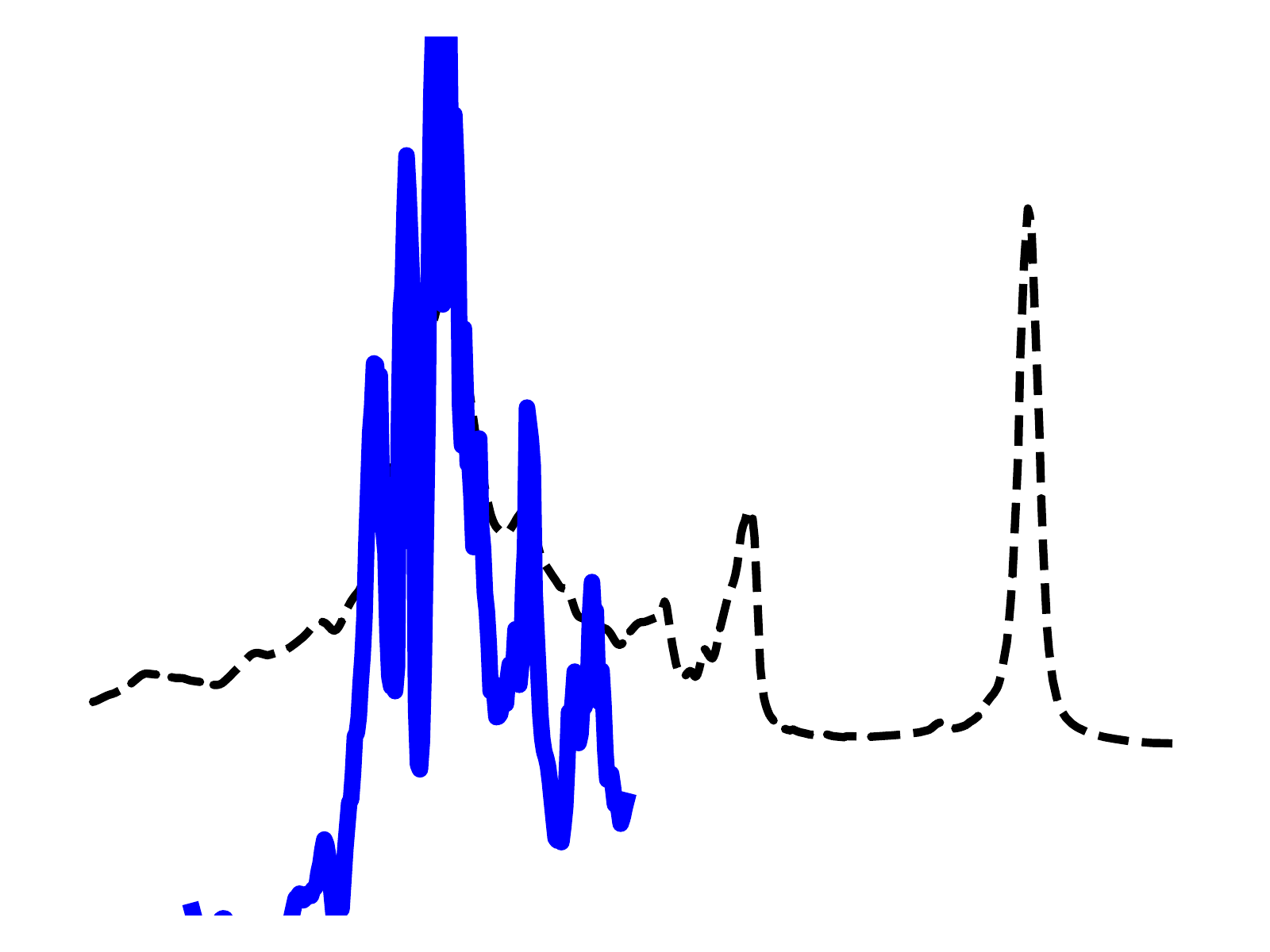}
		\includegraphics[width=.33\linewidth,height=2.1cm]{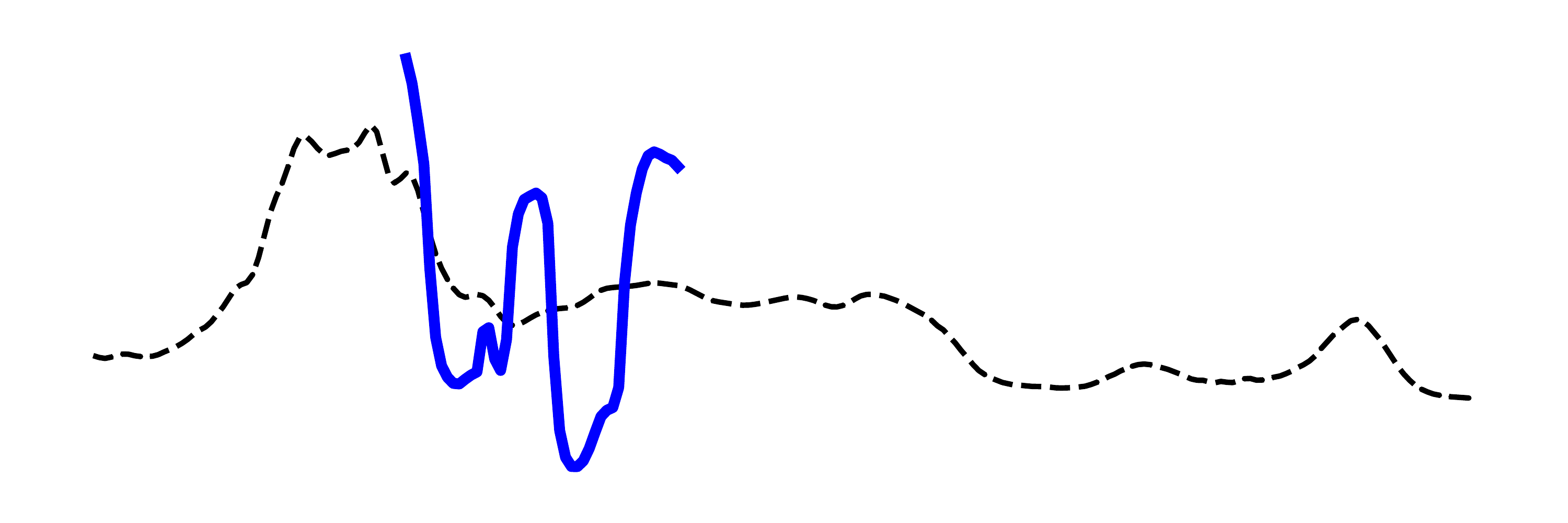}%
		\includegraphics[width=.33\linewidth,height=2.1cm]{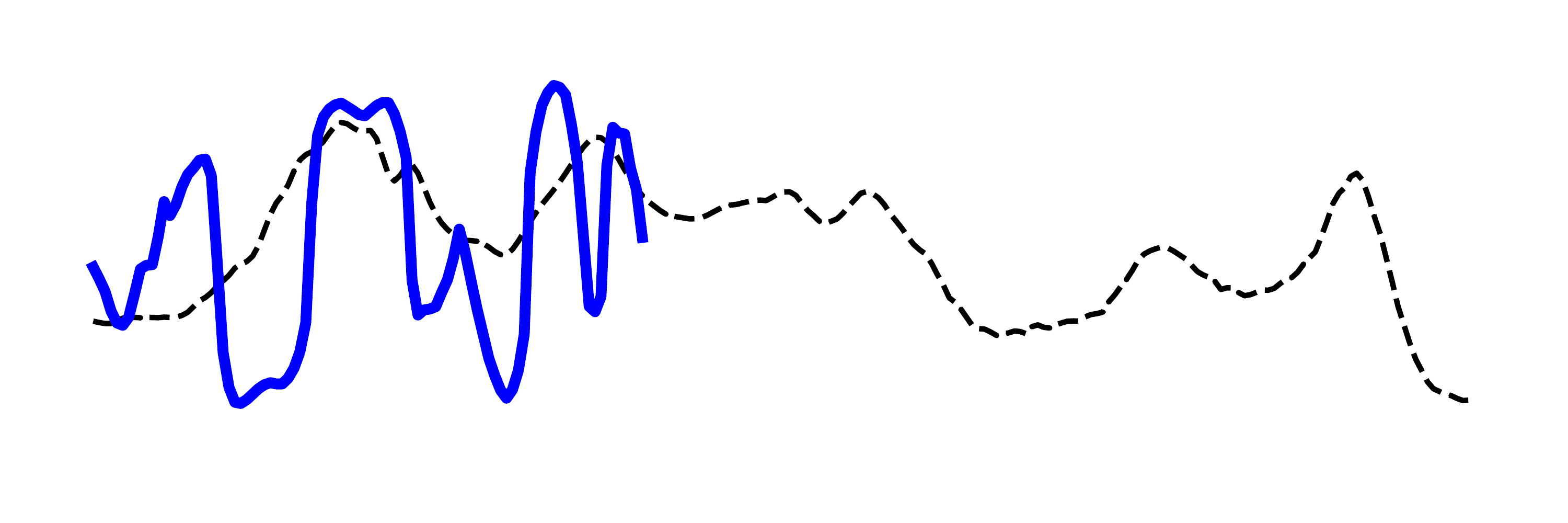}%
		\includegraphics[width=.33\linewidth,height=2.1cm]{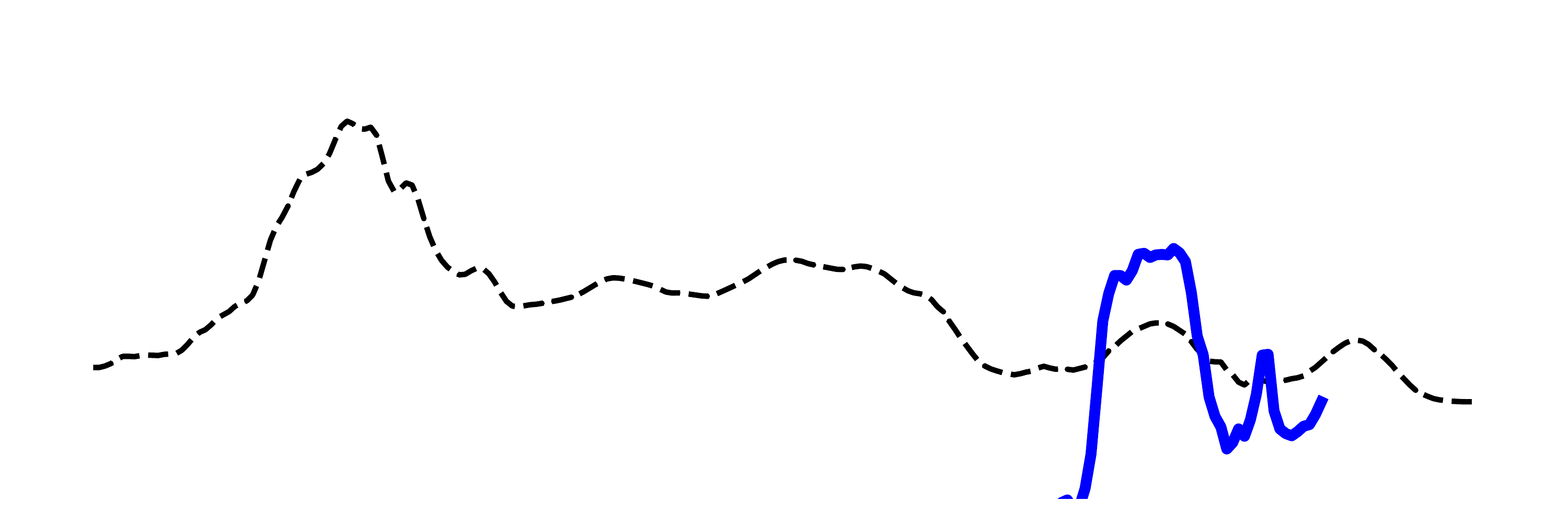}

		\caption*{(a) Learning Shapelets~\cite{Grabocka2014}}
	\end{minipage}
	\hfill
	\begin{minipage}[b]{.45\linewidth}
		\includegraphics[width=.33\linewidth,height=2.1cm]{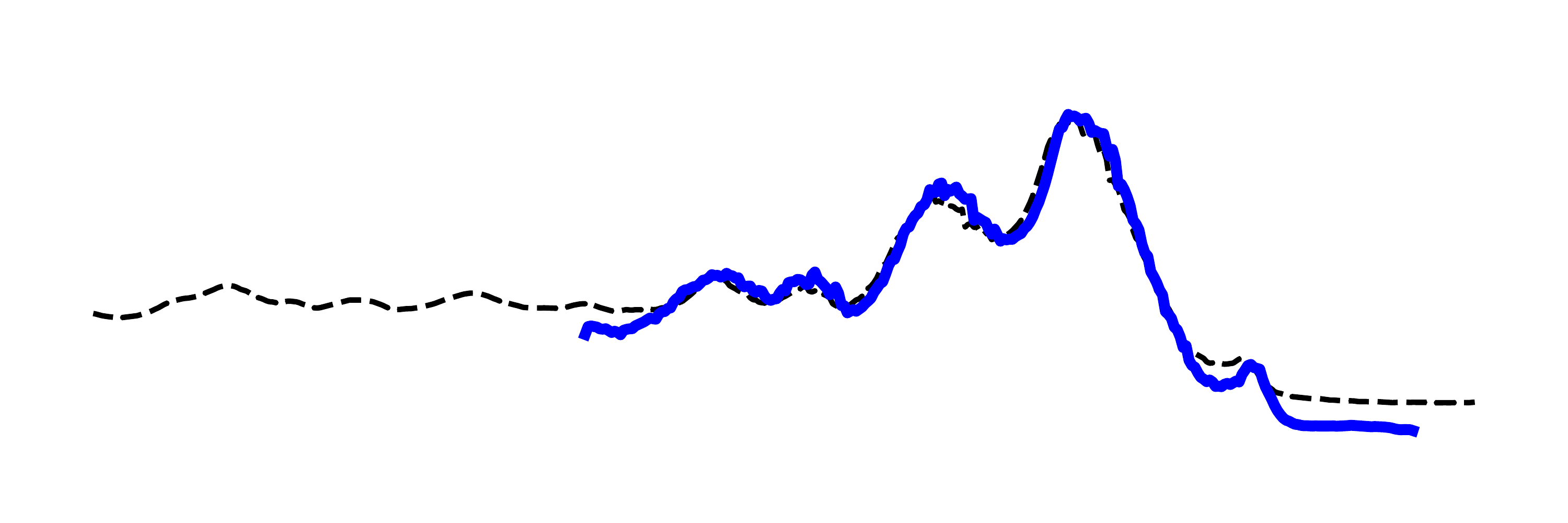}%
		\includegraphics[width=.33\linewidth,height=2.1cm]{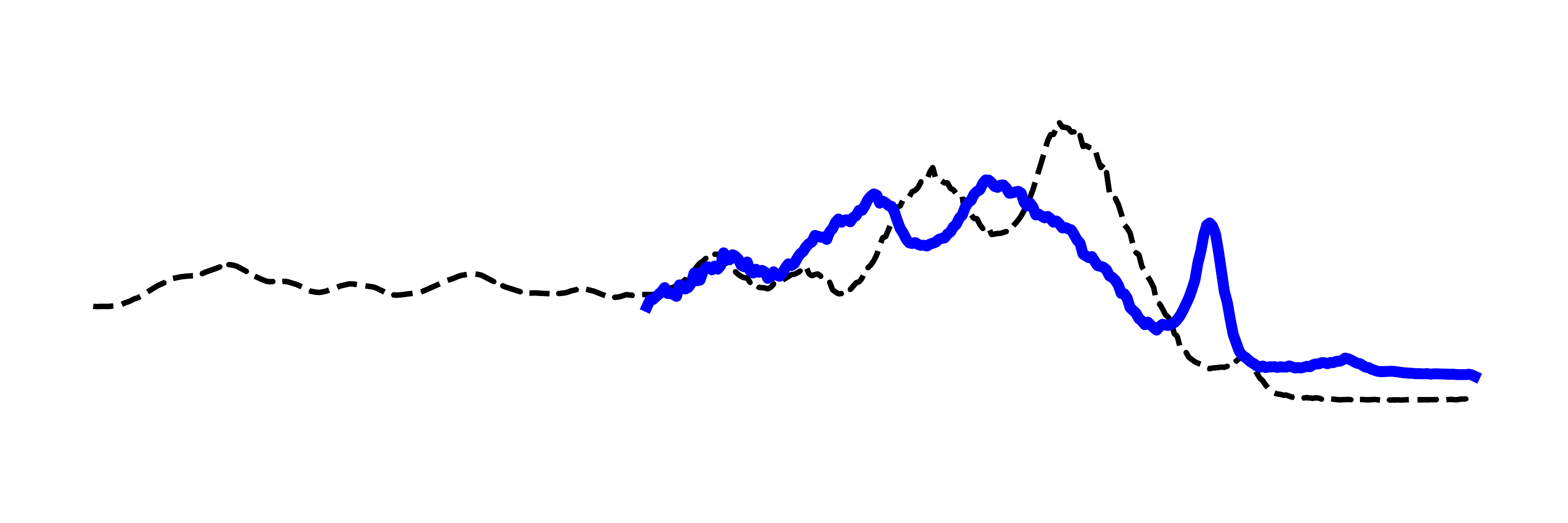}%
		\includegraphics[width=.33\linewidth,height=2.1cm]{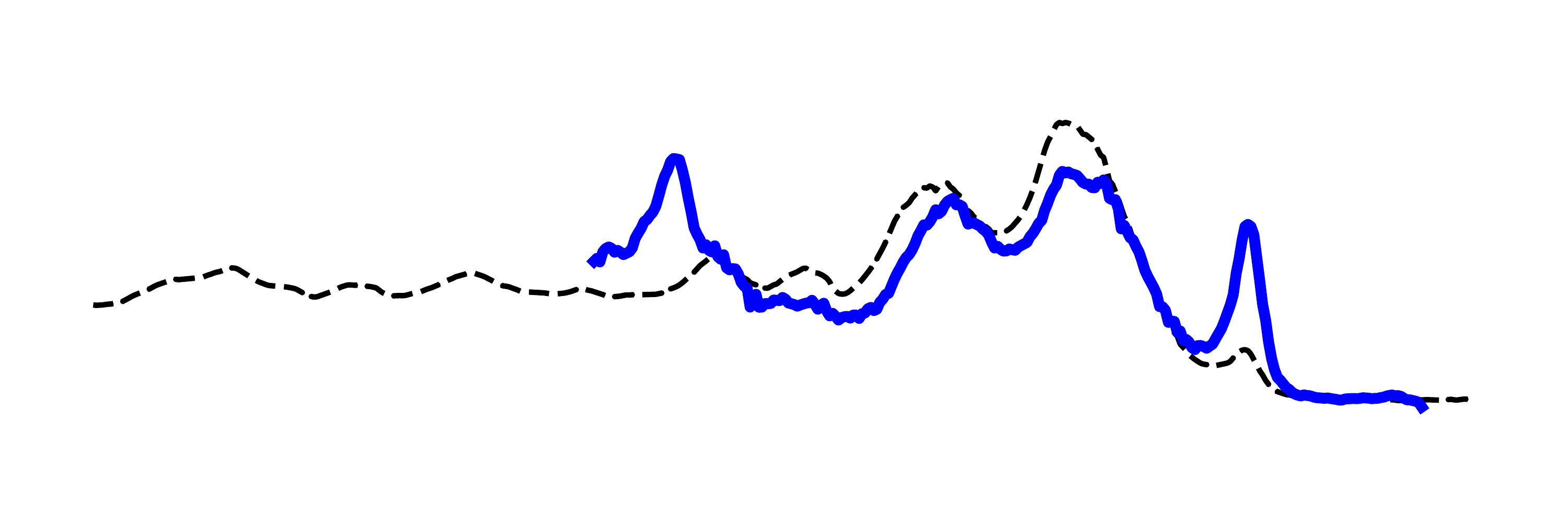}
		\includegraphics[width=.33\linewidth,height=2.1cm]{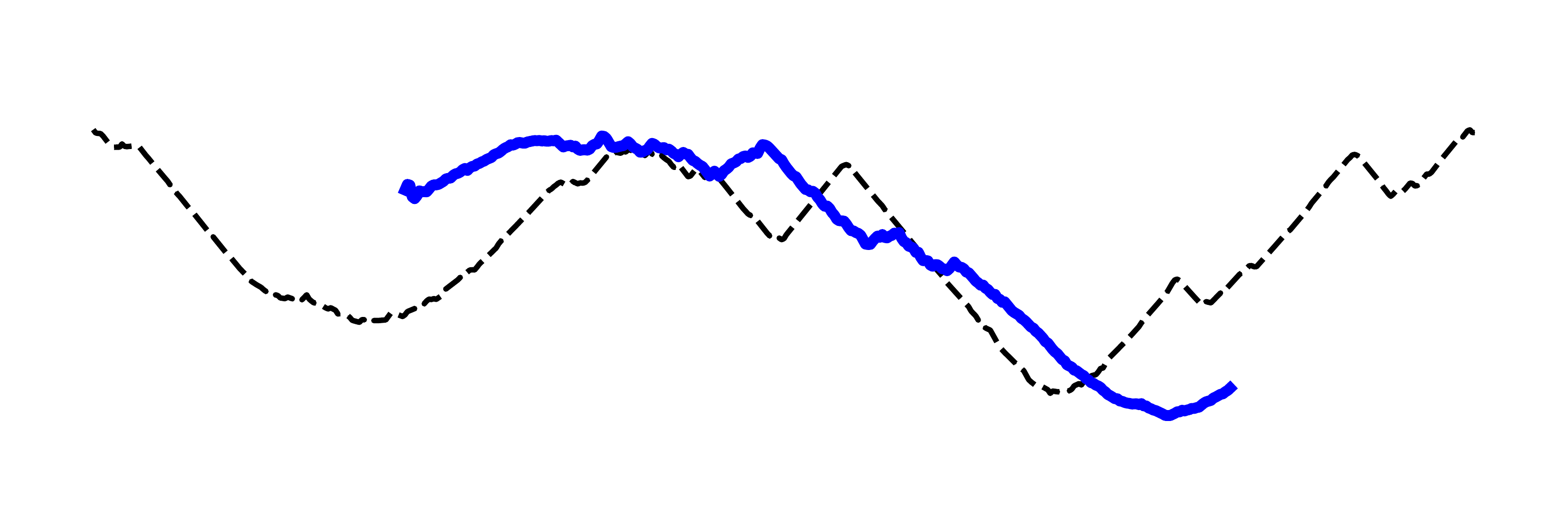}%
		\includegraphics[width=.33\linewidth,height=2.1cm]{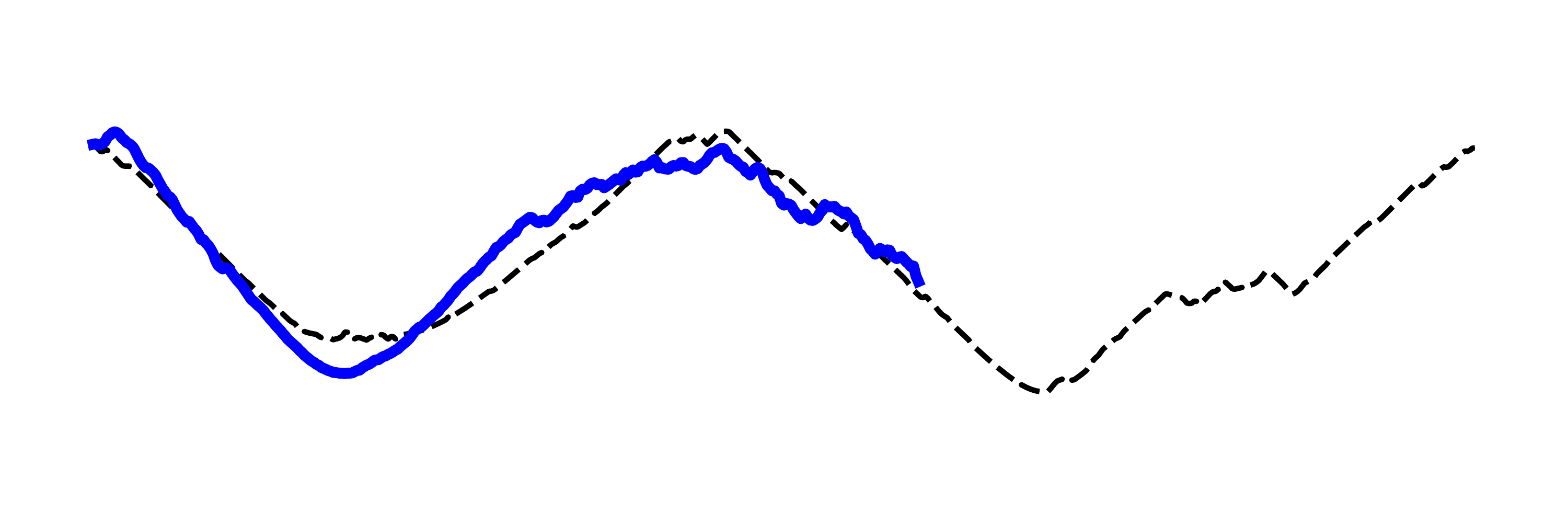}%
		\includegraphics[width=.33\linewidth,height=2.1cm]{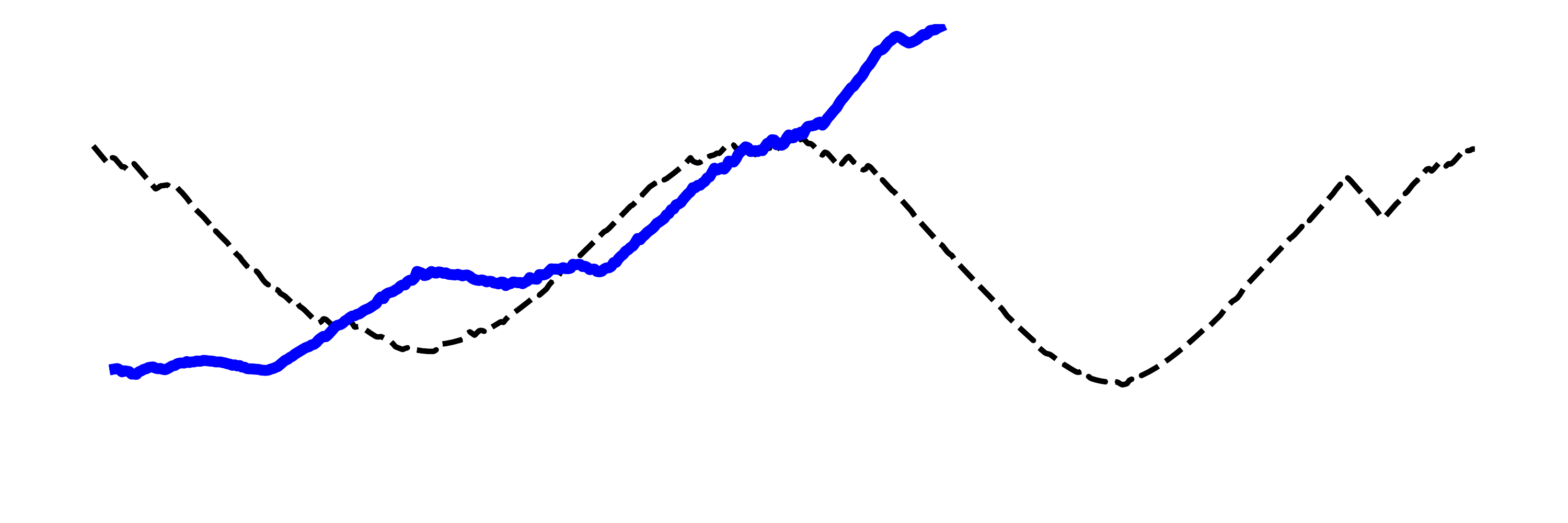}
		\includegraphics[width=.33\linewidth,height=2.1cm]{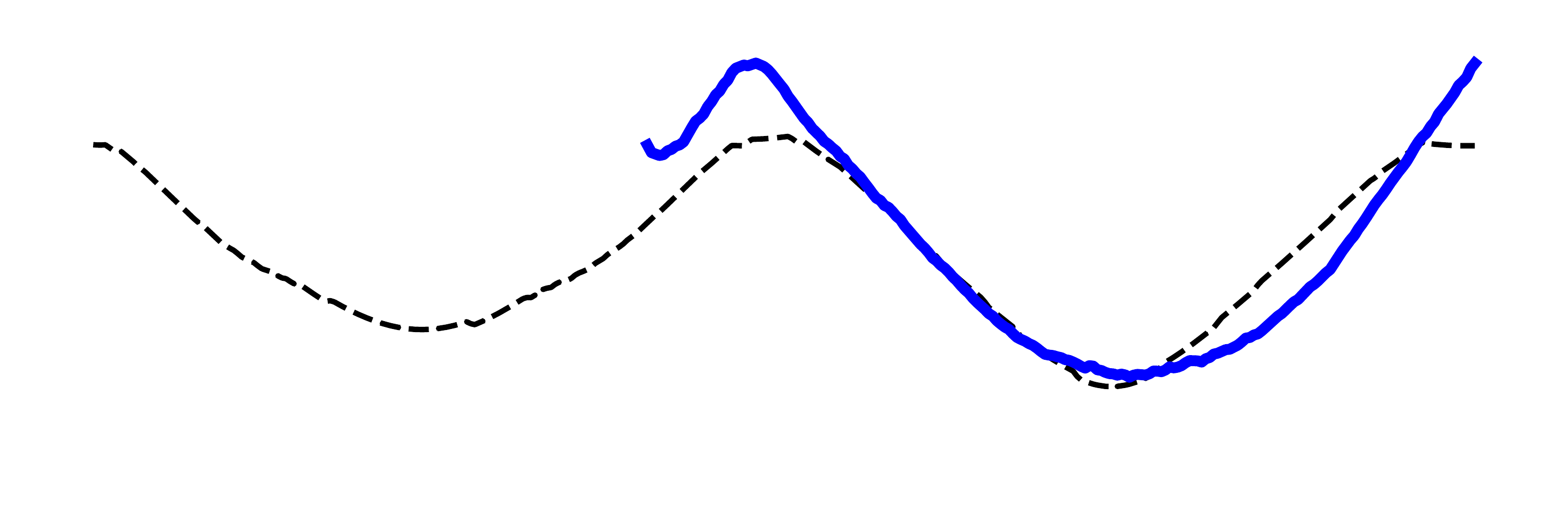}%
		\includegraphics[width=.33\linewidth,height=2.1cm]{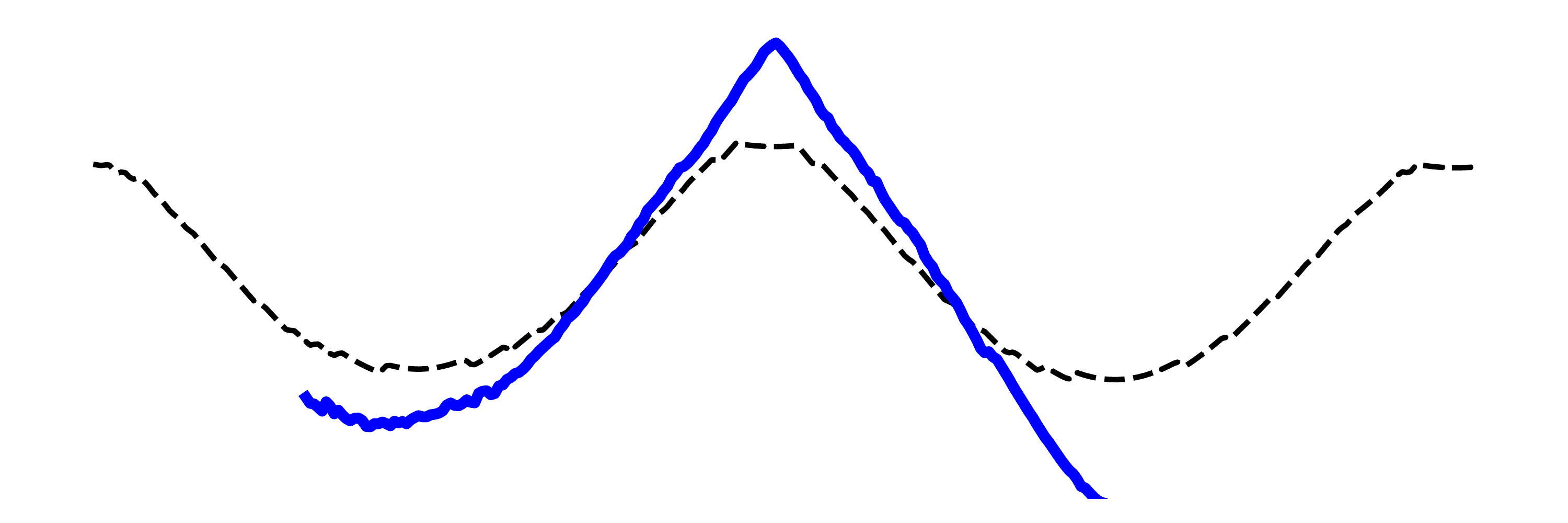}%
		\includegraphics[width=.33\linewidth,height=2.1cm]{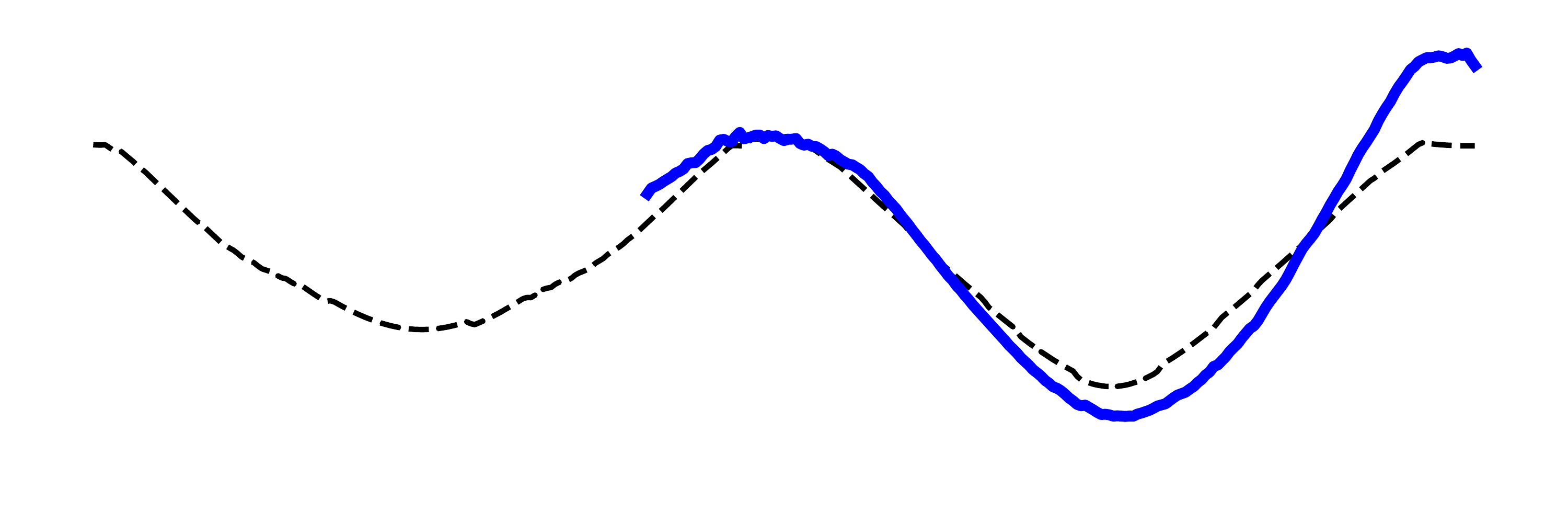}
		\includegraphics[width=.33\linewidth]{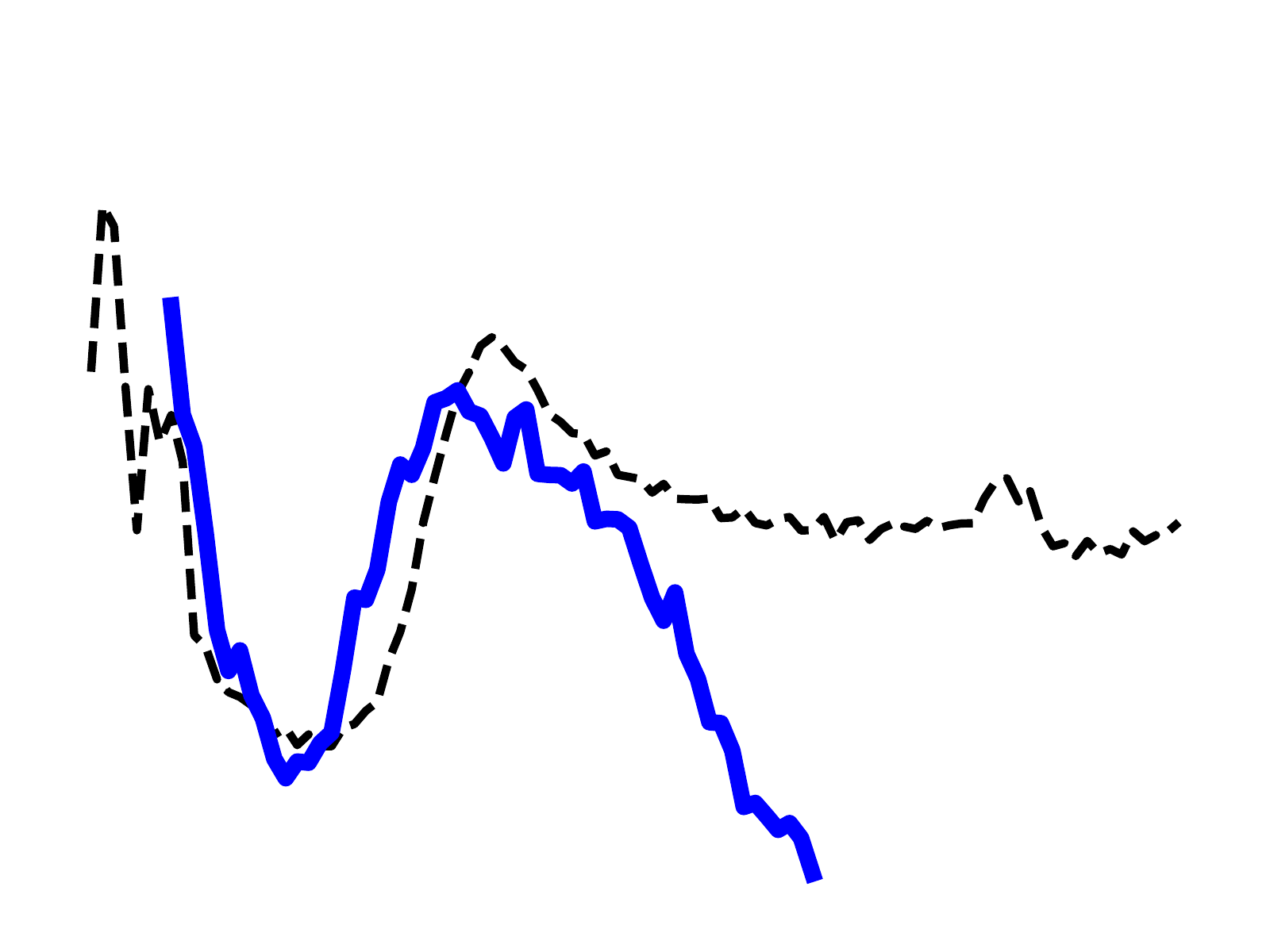}%
		\includegraphics[width=.33\linewidth]{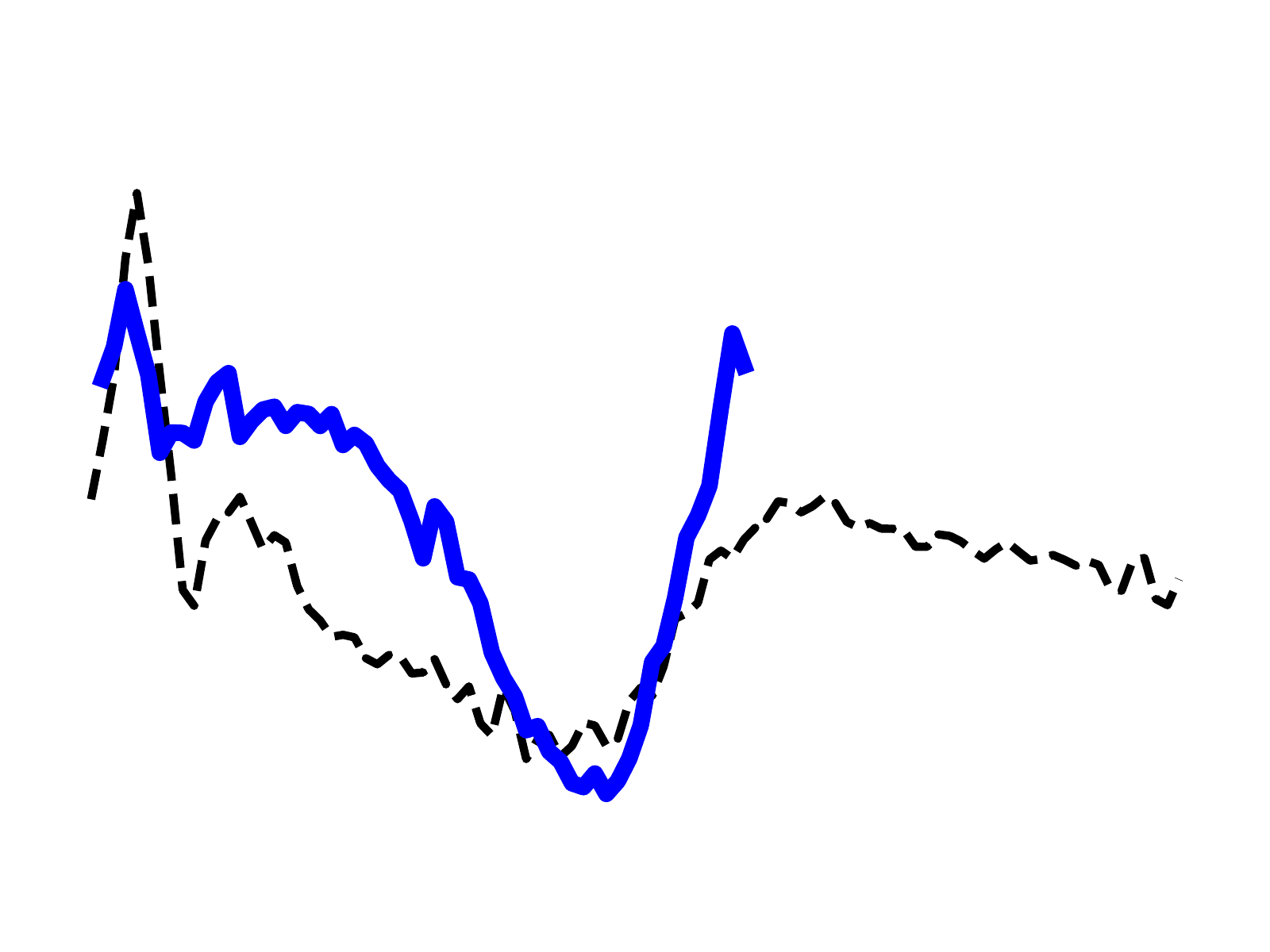}%
		\includegraphics[width=.33\linewidth]{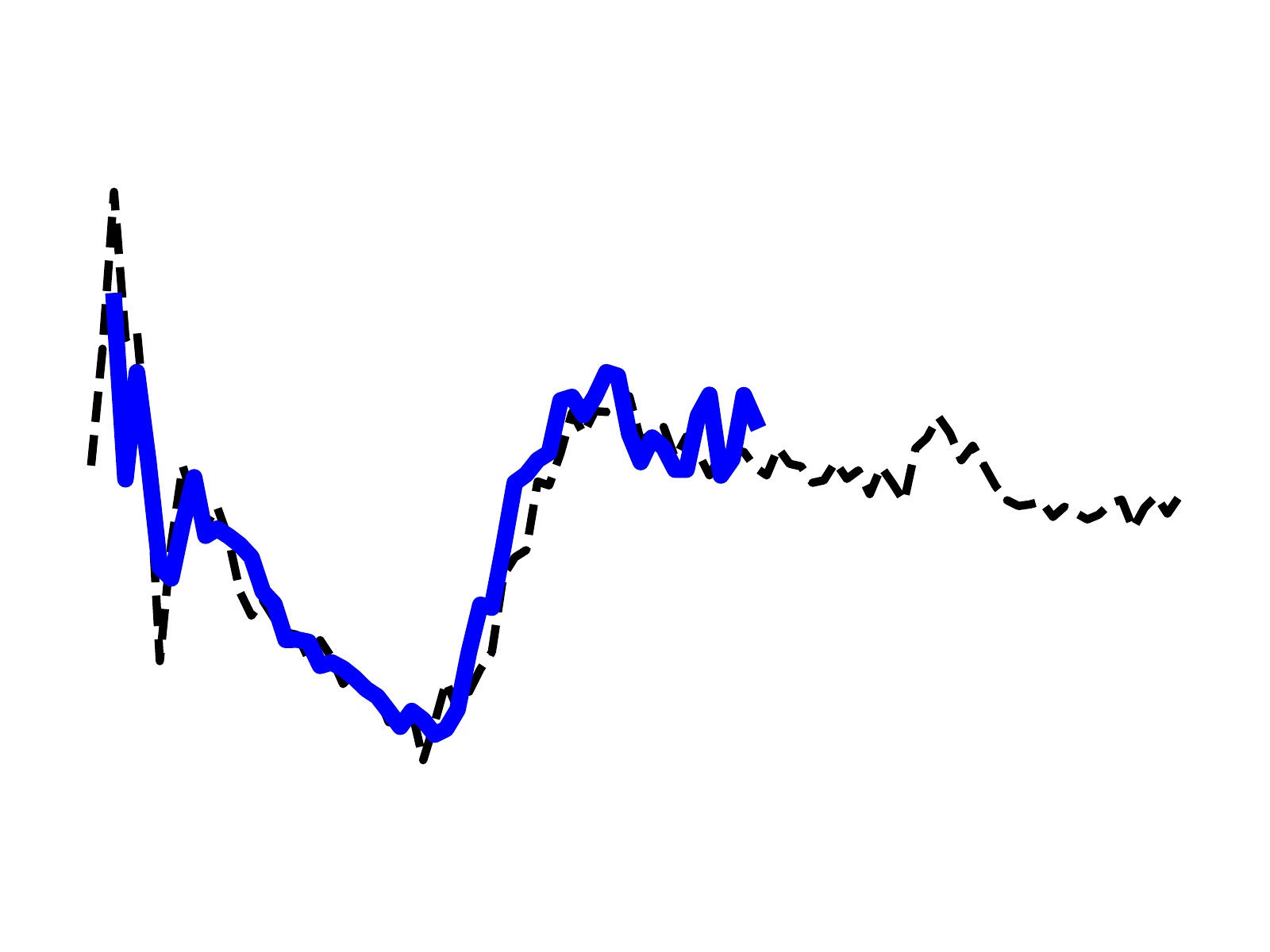}
		\includegraphics[width=.33\linewidth]{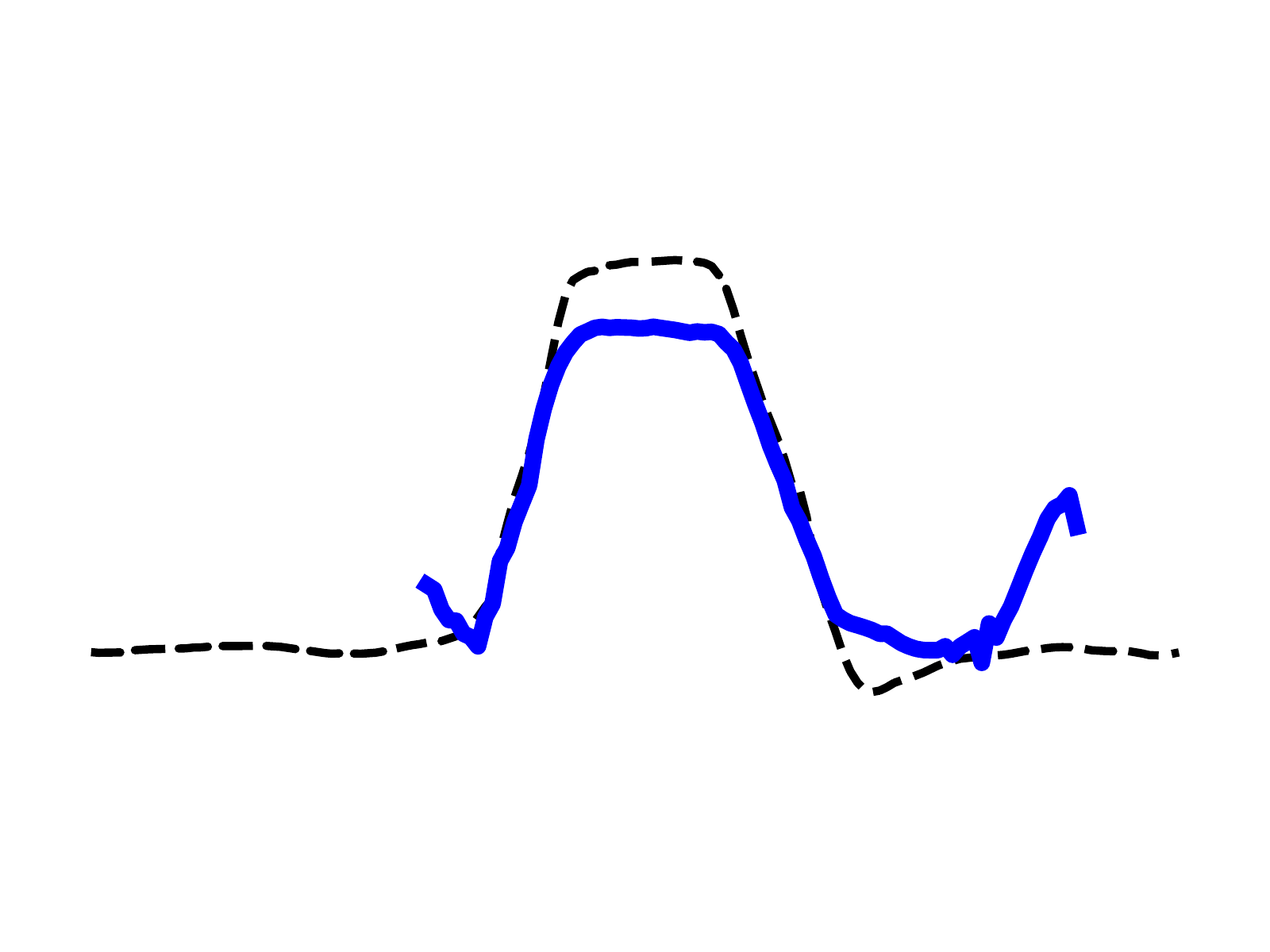}%
		\includegraphics[width=.33\linewidth]{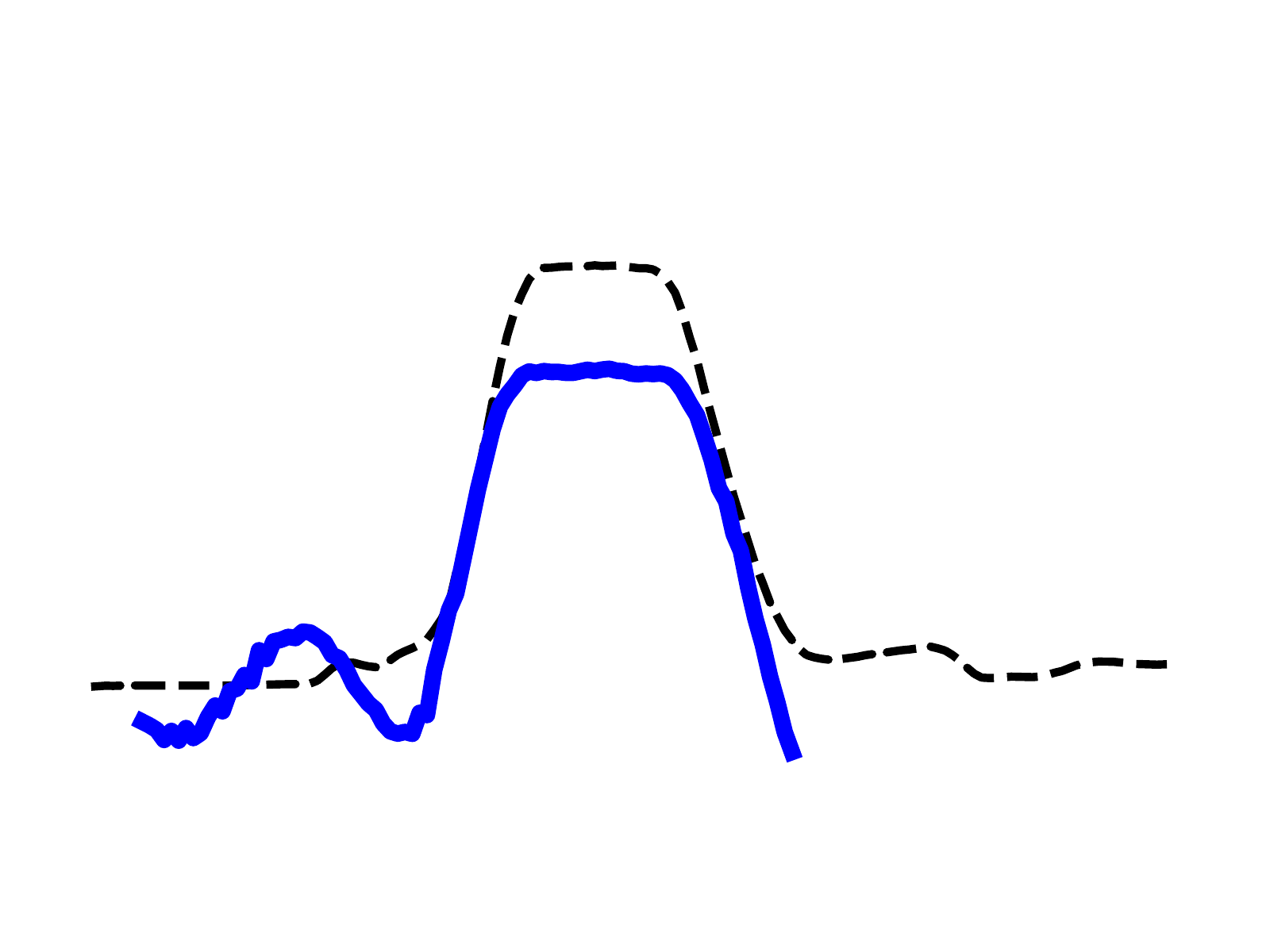}%
		\includegraphics[width=.33\linewidth]{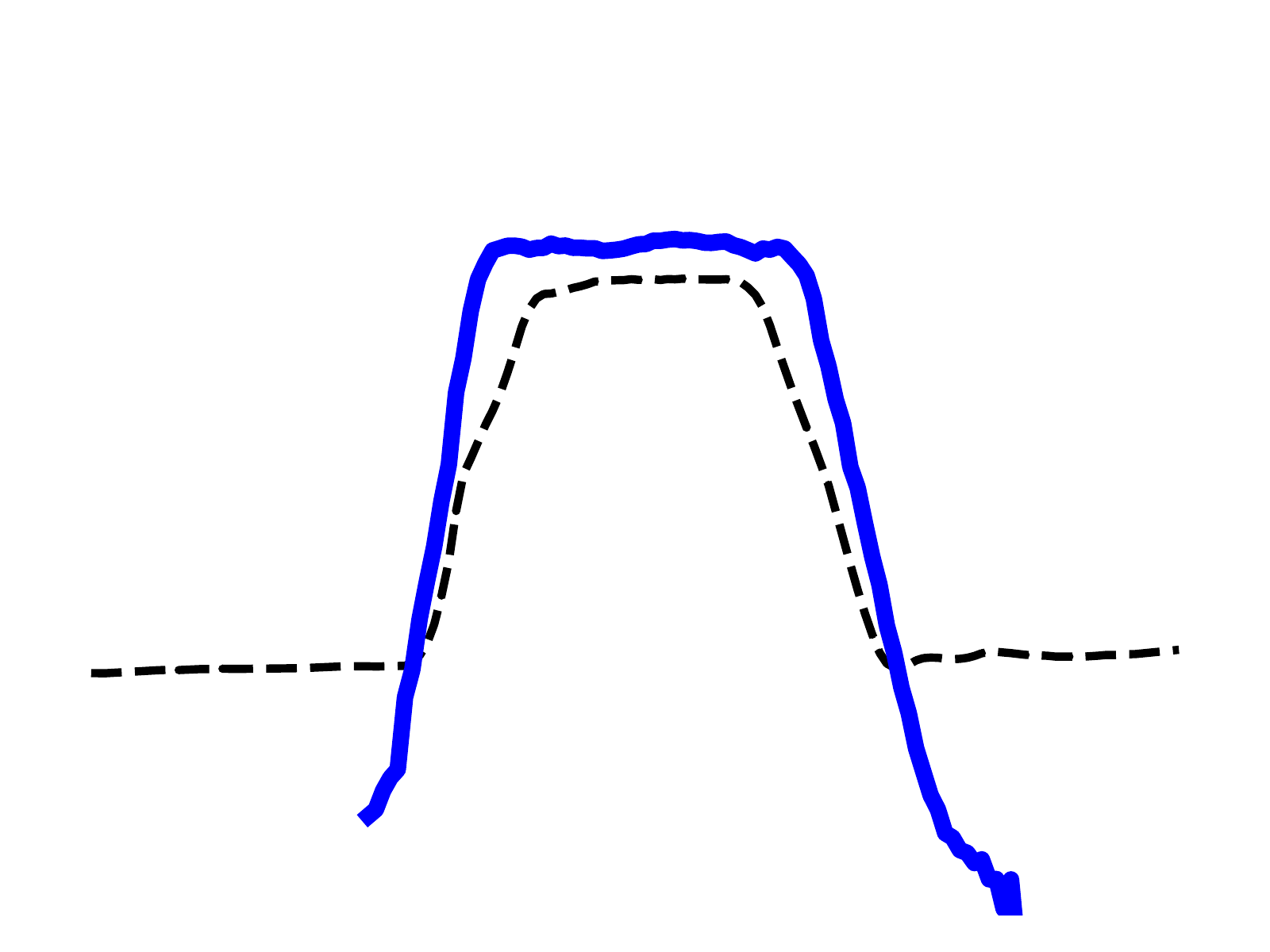}
		\includegraphics[width=.33\linewidth]{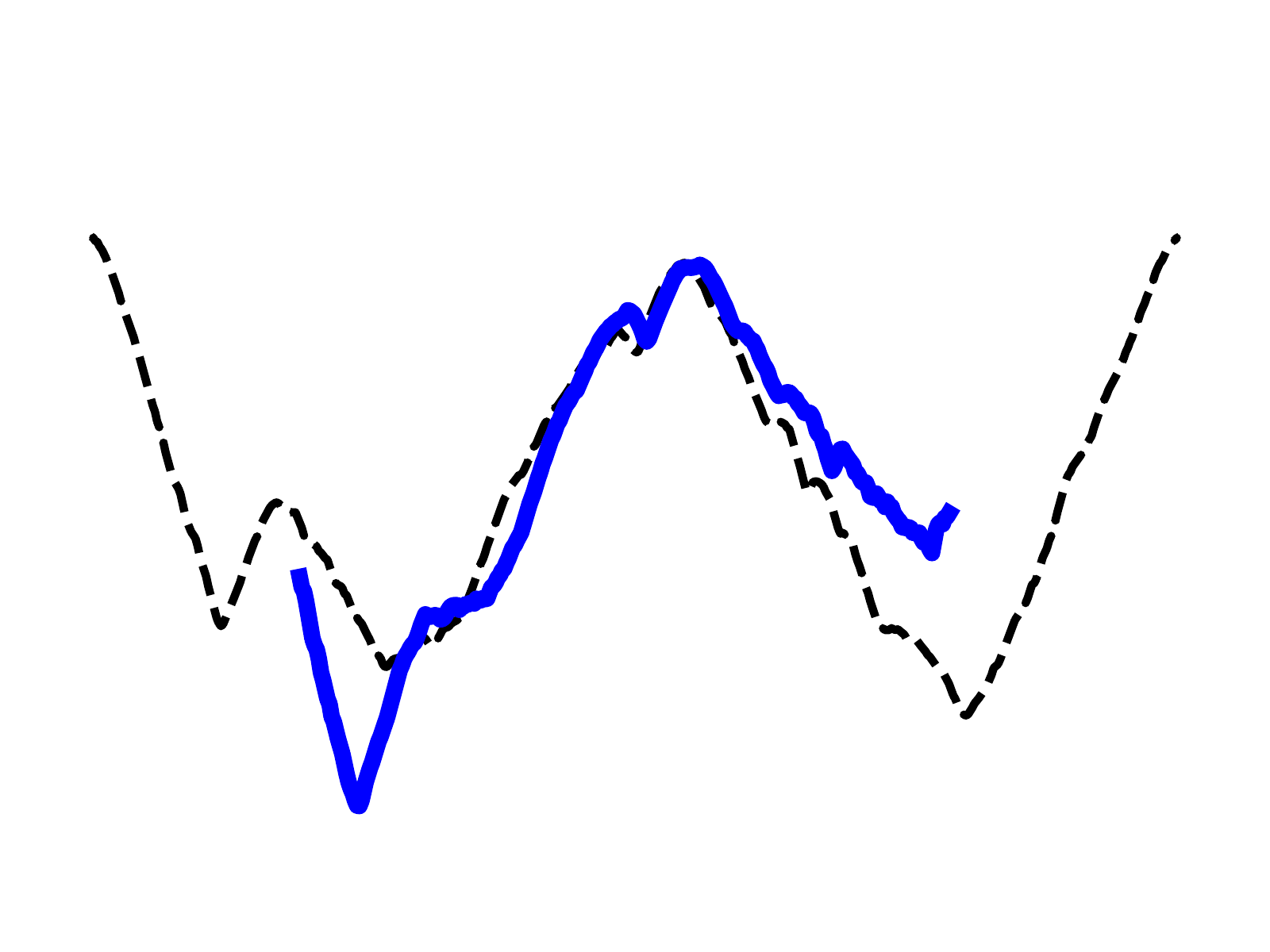}%
		\includegraphics[width=.33\linewidth]{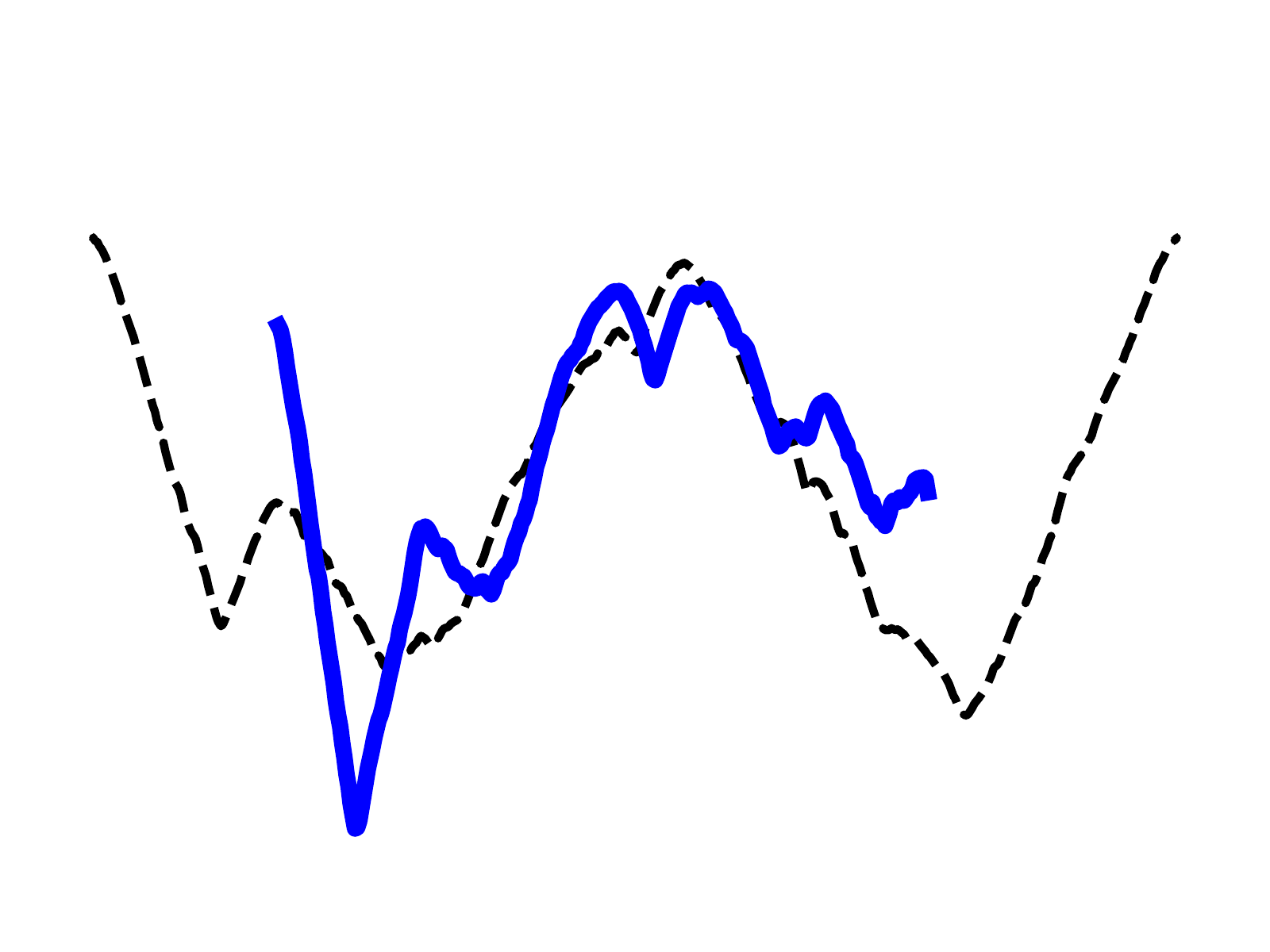}%
		\includegraphics[width=.33\linewidth]{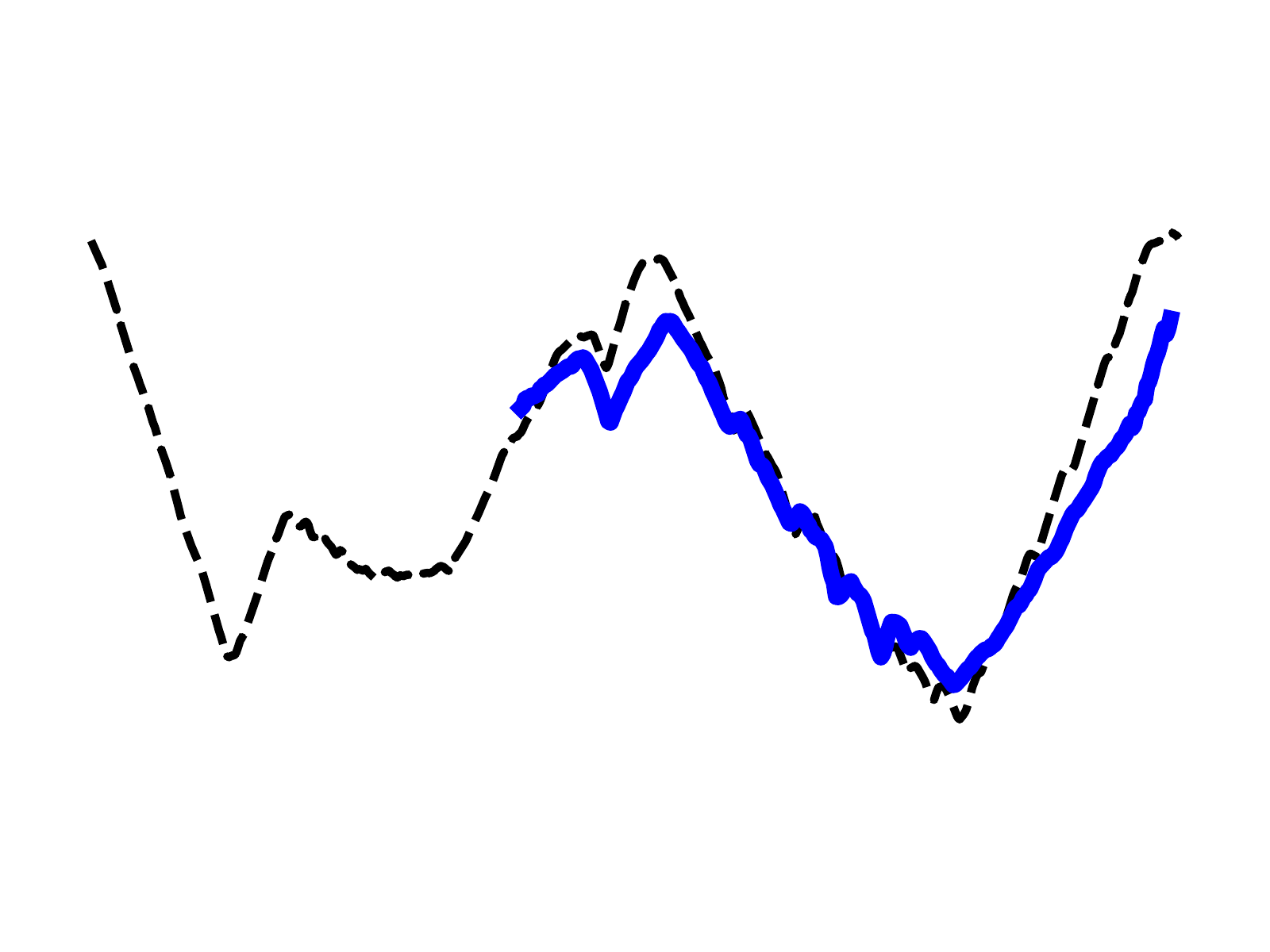}
		\includegraphics[width=.33\linewidth,height=2.1cm]{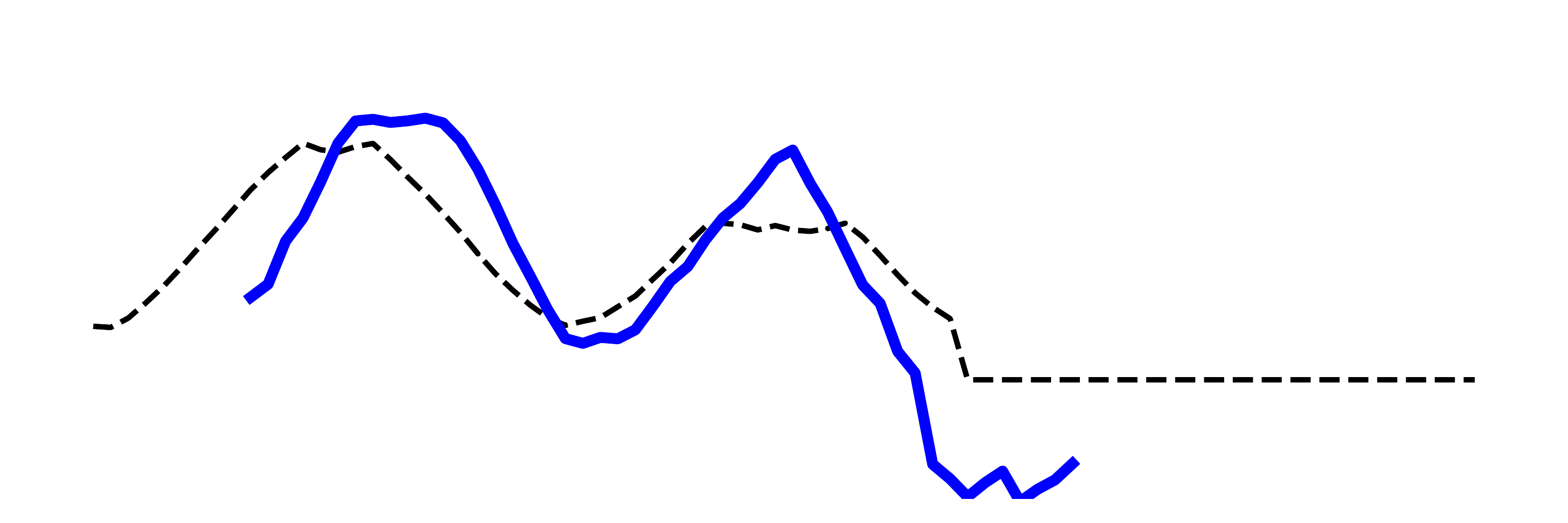}%
		\includegraphics[width=.33\linewidth,height=2.1cm]{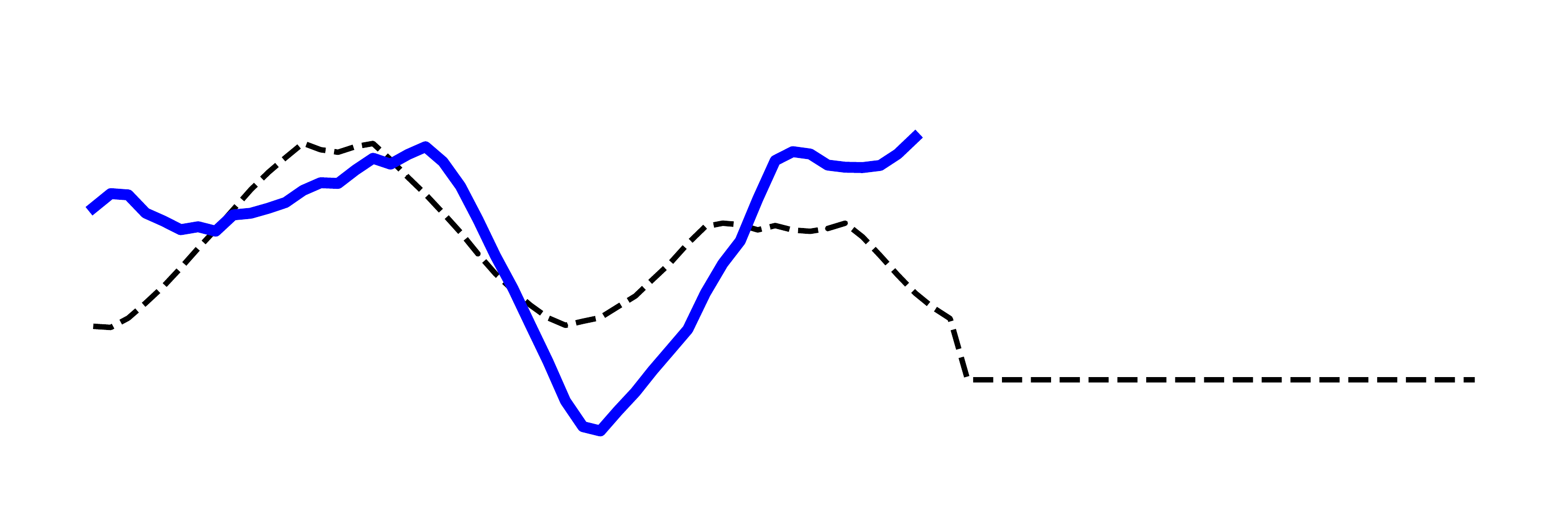}%
		\includegraphics[width=.33\linewidth,height=2.1cm]{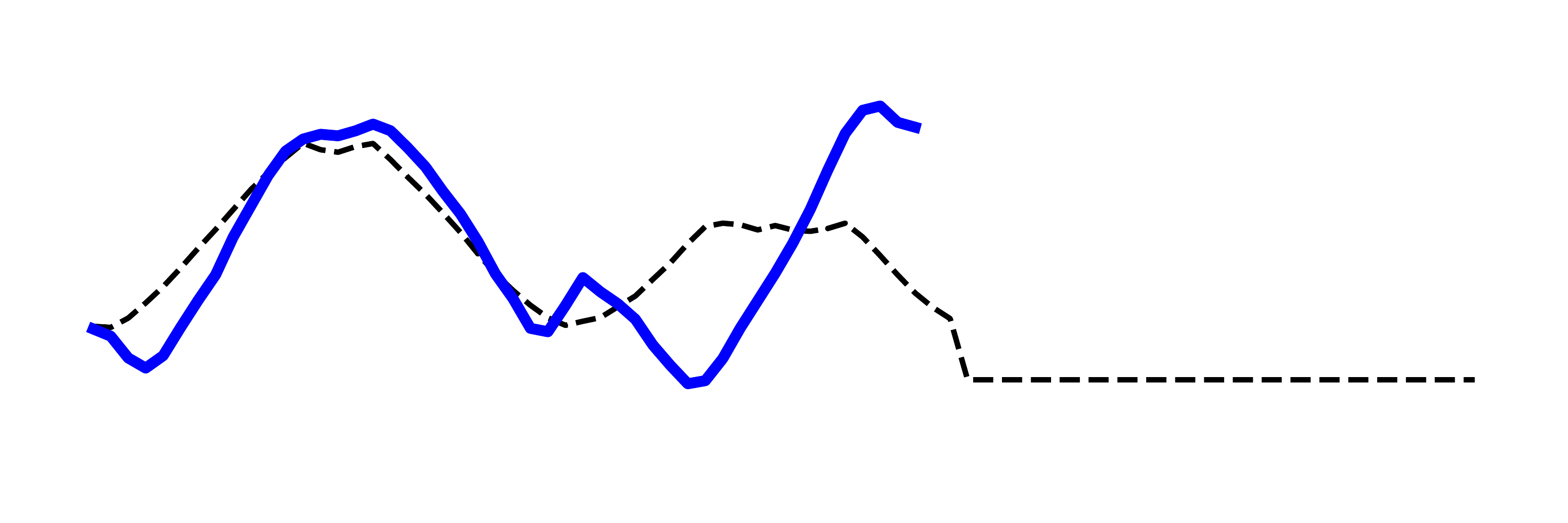}
		\includegraphics[width=.33\linewidth]{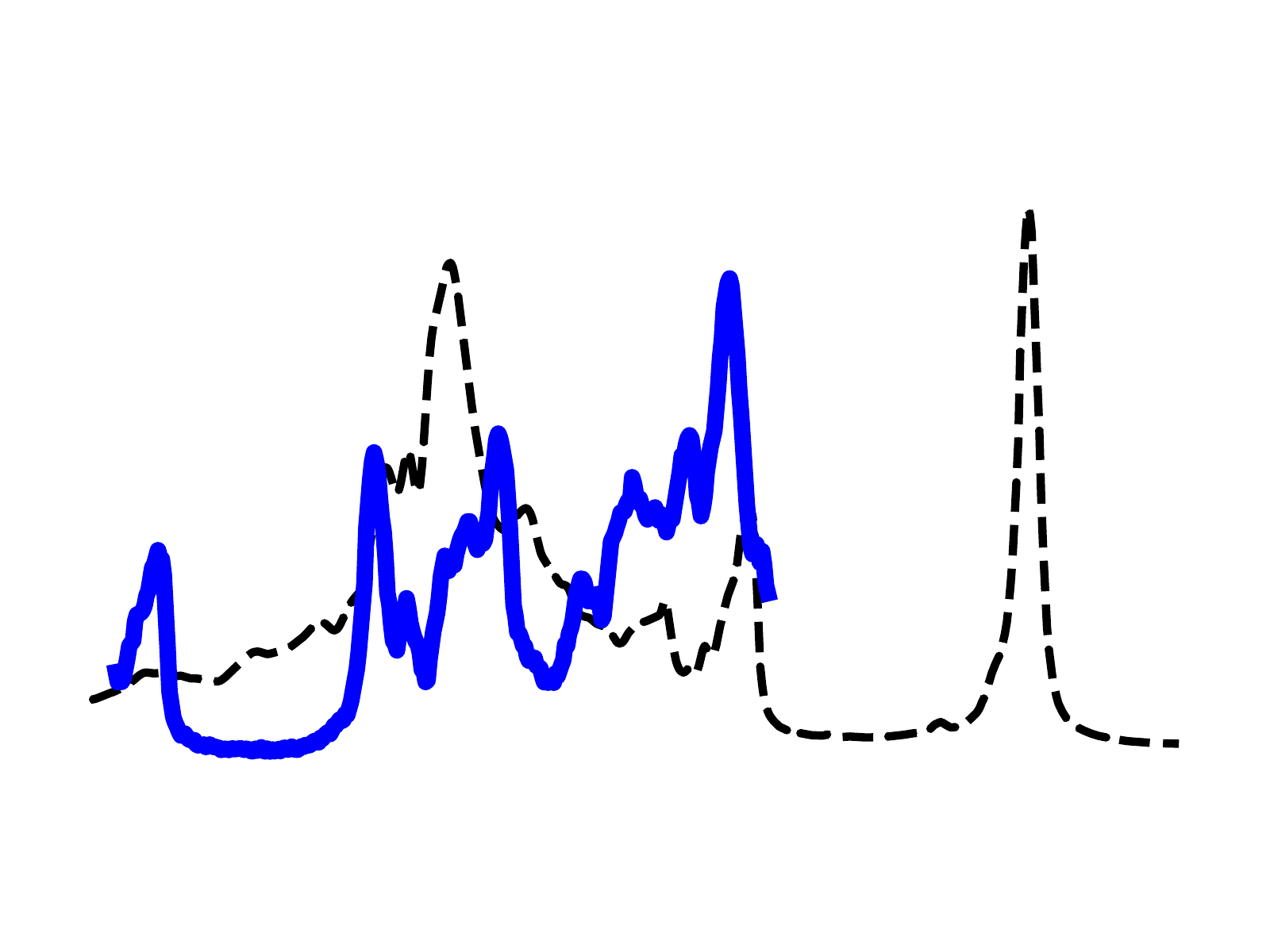}%
		\includegraphics[width=.33\linewidth]{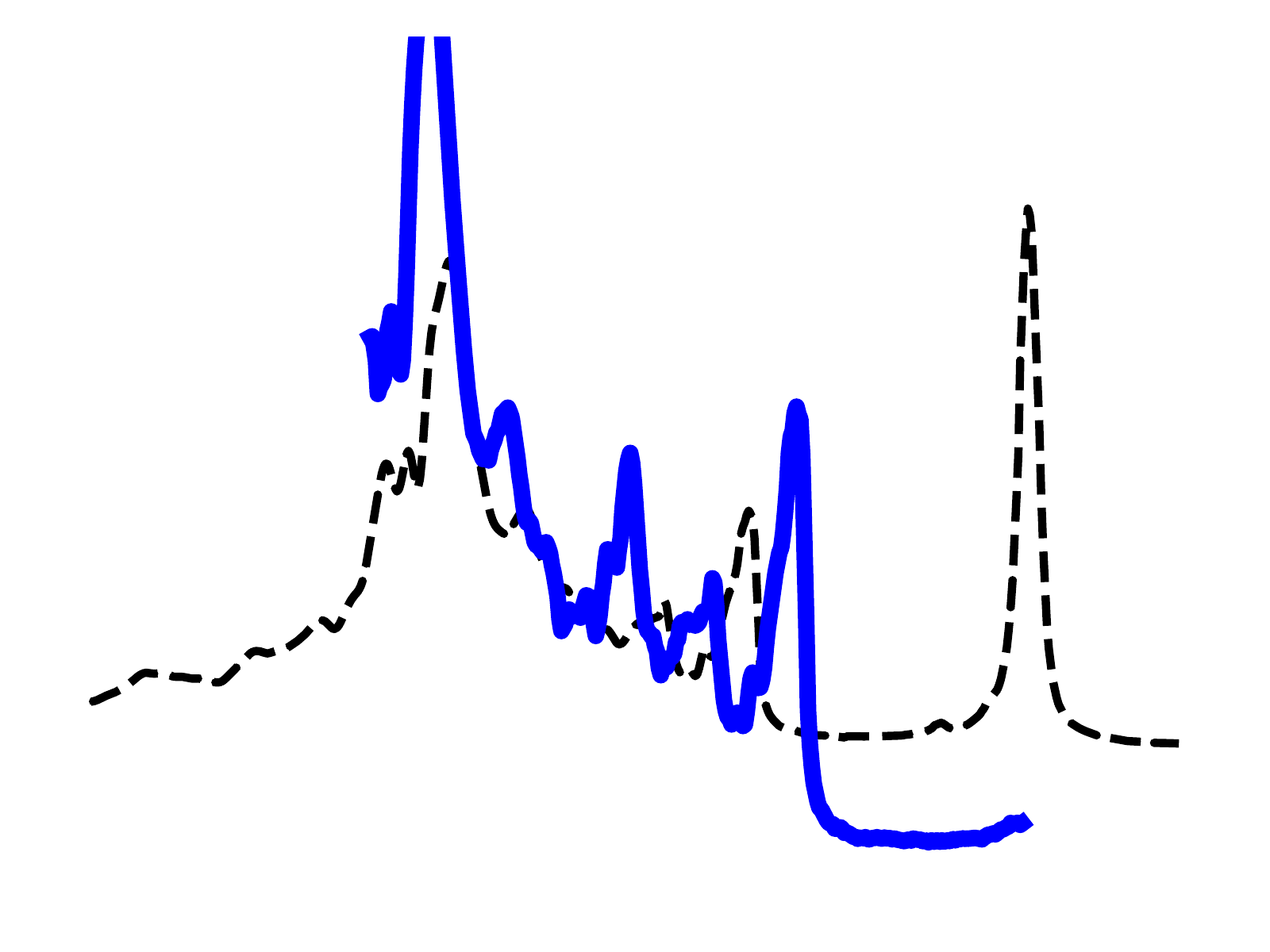}%
		\includegraphics[width=.33\linewidth]{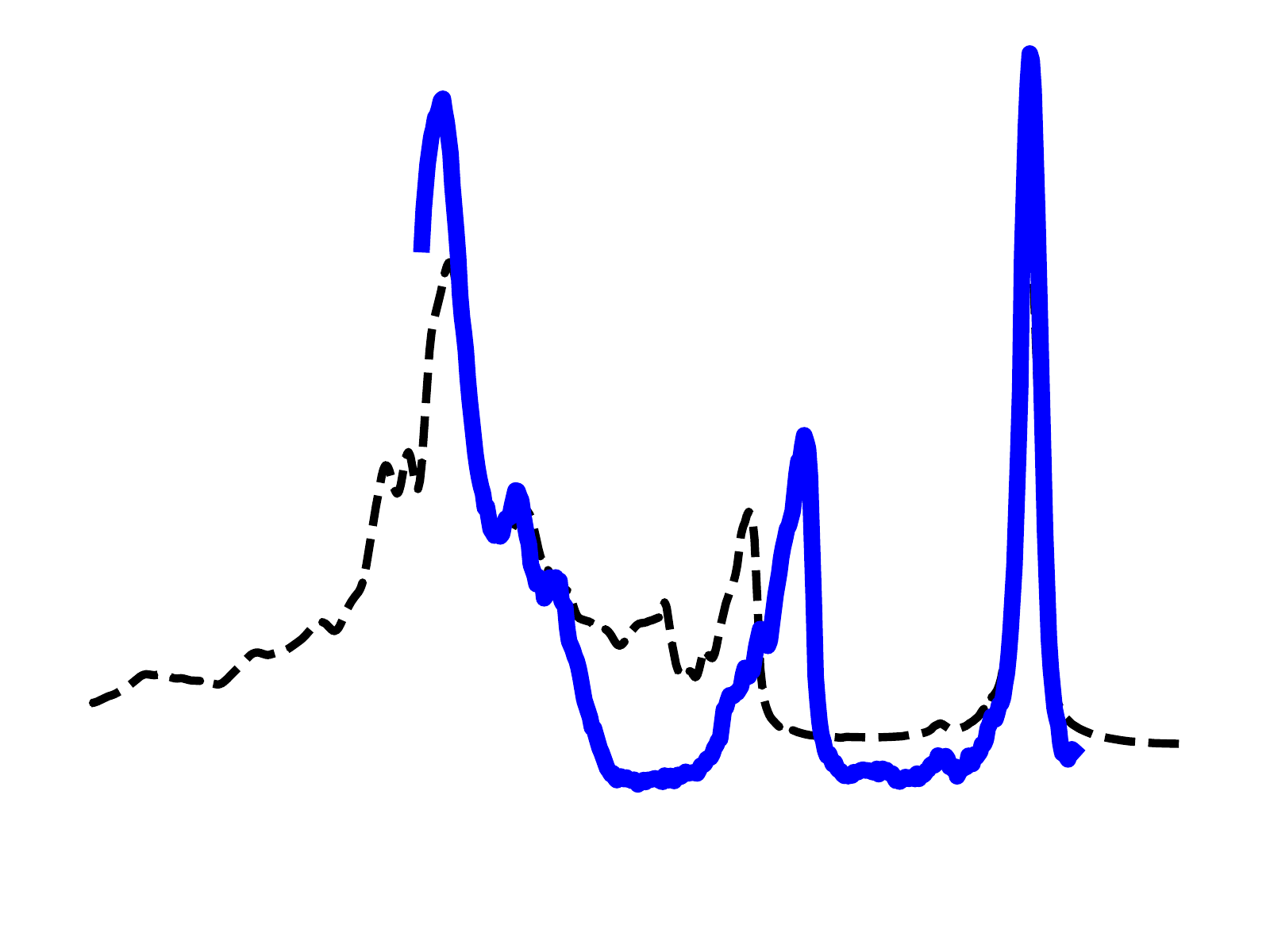}
		\includegraphics[width=.33\linewidth,height=2.1cm]{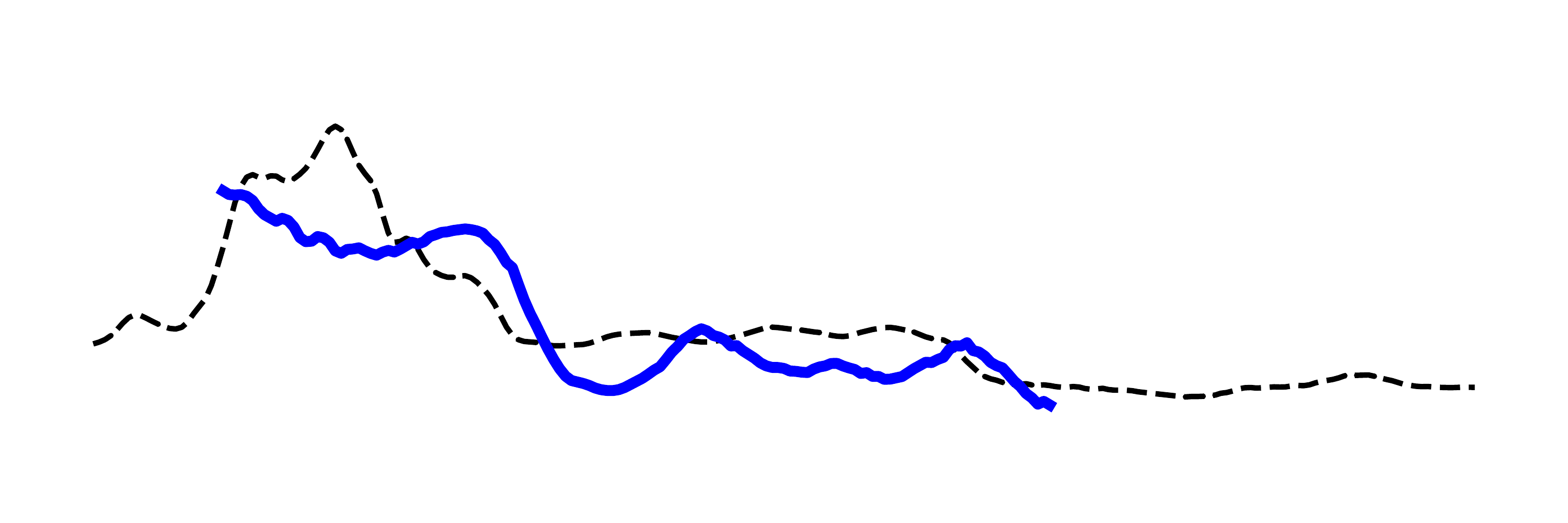}%
		\includegraphics[width=.33\linewidth,height=2.1cm]{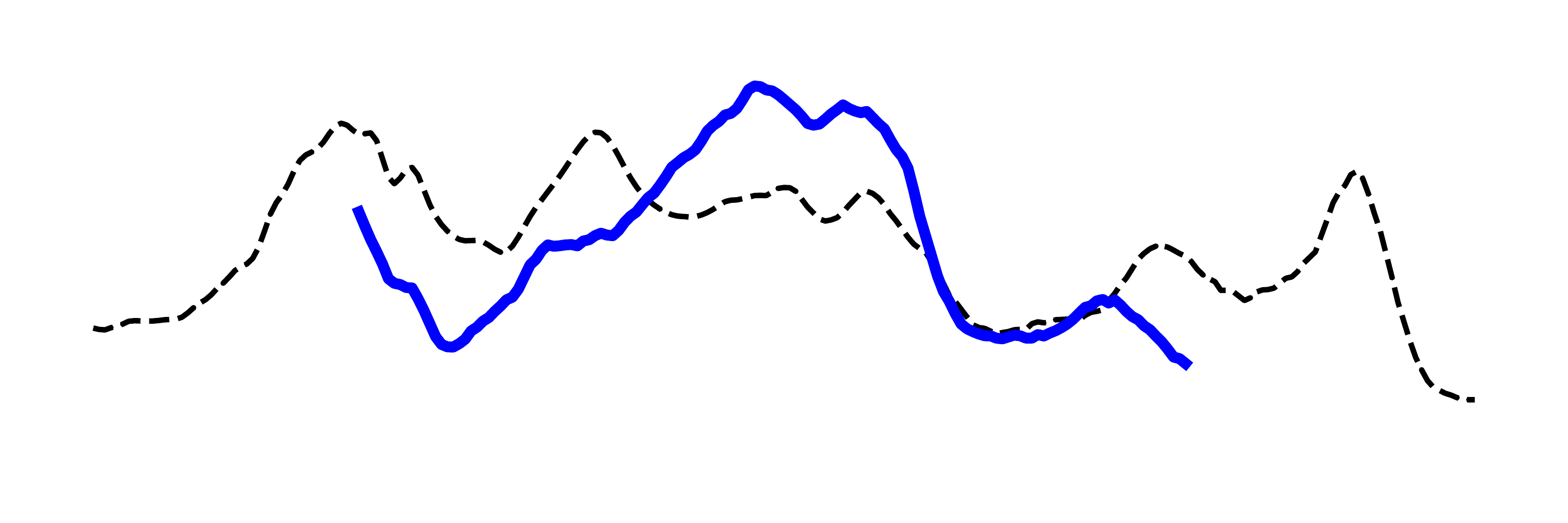}%
		\includegraphics[width=.33\linewidth,height=2.1cm]{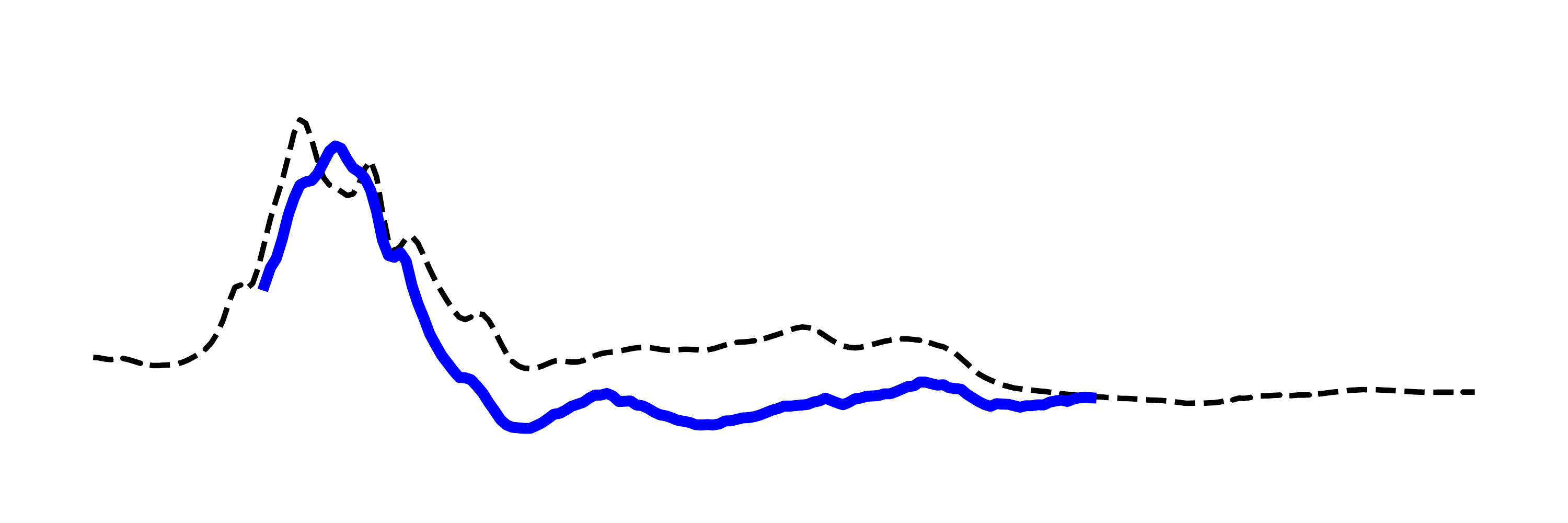}
		
		\caption*{(b) \aiprshort{}-CNN}
	\end{minipage}
	\caption{The three most discriminative shapelets obtained for the datasets Beef, Car, DiatomSizeReduction, ECG200, GunPoint, Herring, MiddlePhalanxOutlineCorrect, OliveOil and Strawberry (rows 1 to 9, respectively) using (a) Learning Shapelets or (b) our \aiprshort{}-CNN architecture. The average discriminative power of the shapelets is evaluated using Eq.~\ref{eq:discriminative_shapelets} and each shapelet is superimposed over its best matching time series in the training set.\label{fig:visu_shapelets}}
\end{figure*}

\begin{figure*}
	\centering
	\begin{minipage}[h]{.45\linewidth}
	\subfloat[][]{
		\begin{tabular}{c}
	     	\includegraphics[width=\linewidth]{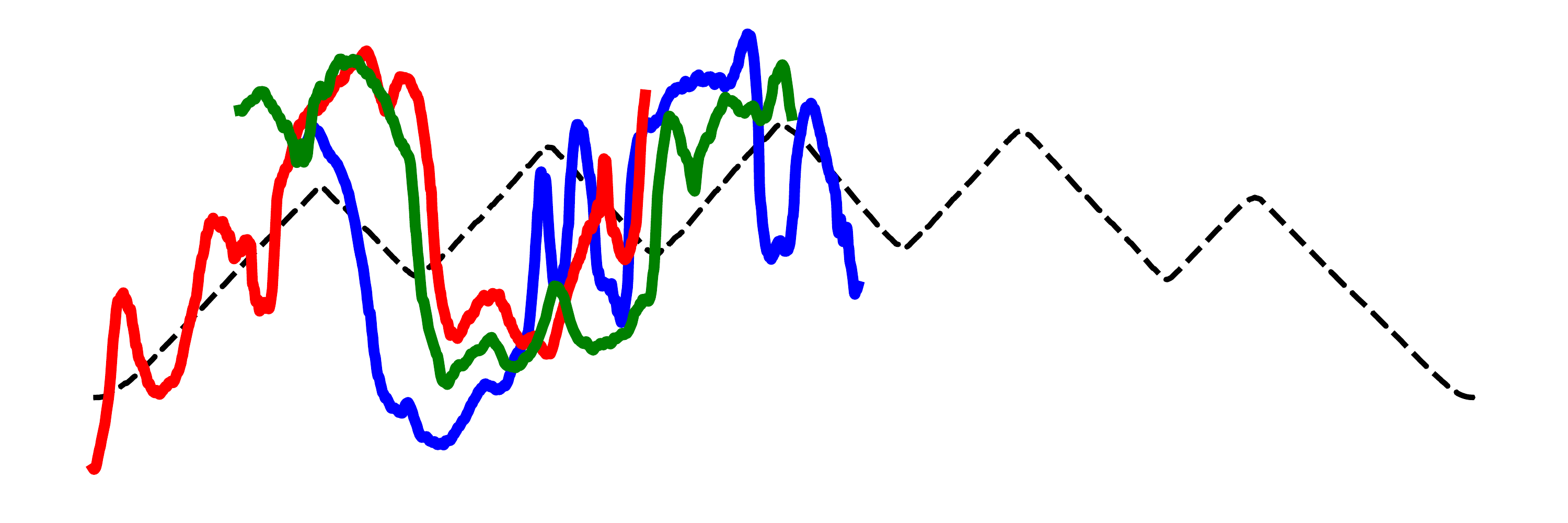}\\
		   	\includegraphics[width=\linewidth]{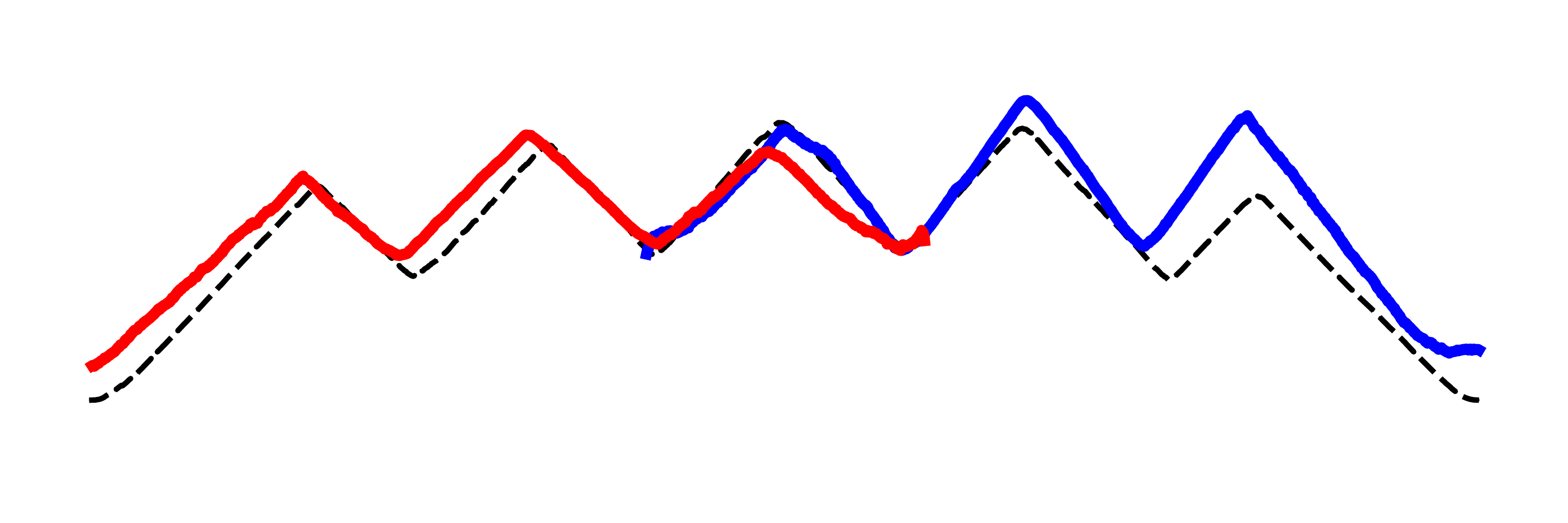}	
		 \end{tabular}
    }
    \end{minipage}
    \hfill
    \begin{minipage}[h]{.45\linewidth}
    \subfloat[]{
		\includegraphics[width=\linewidth]{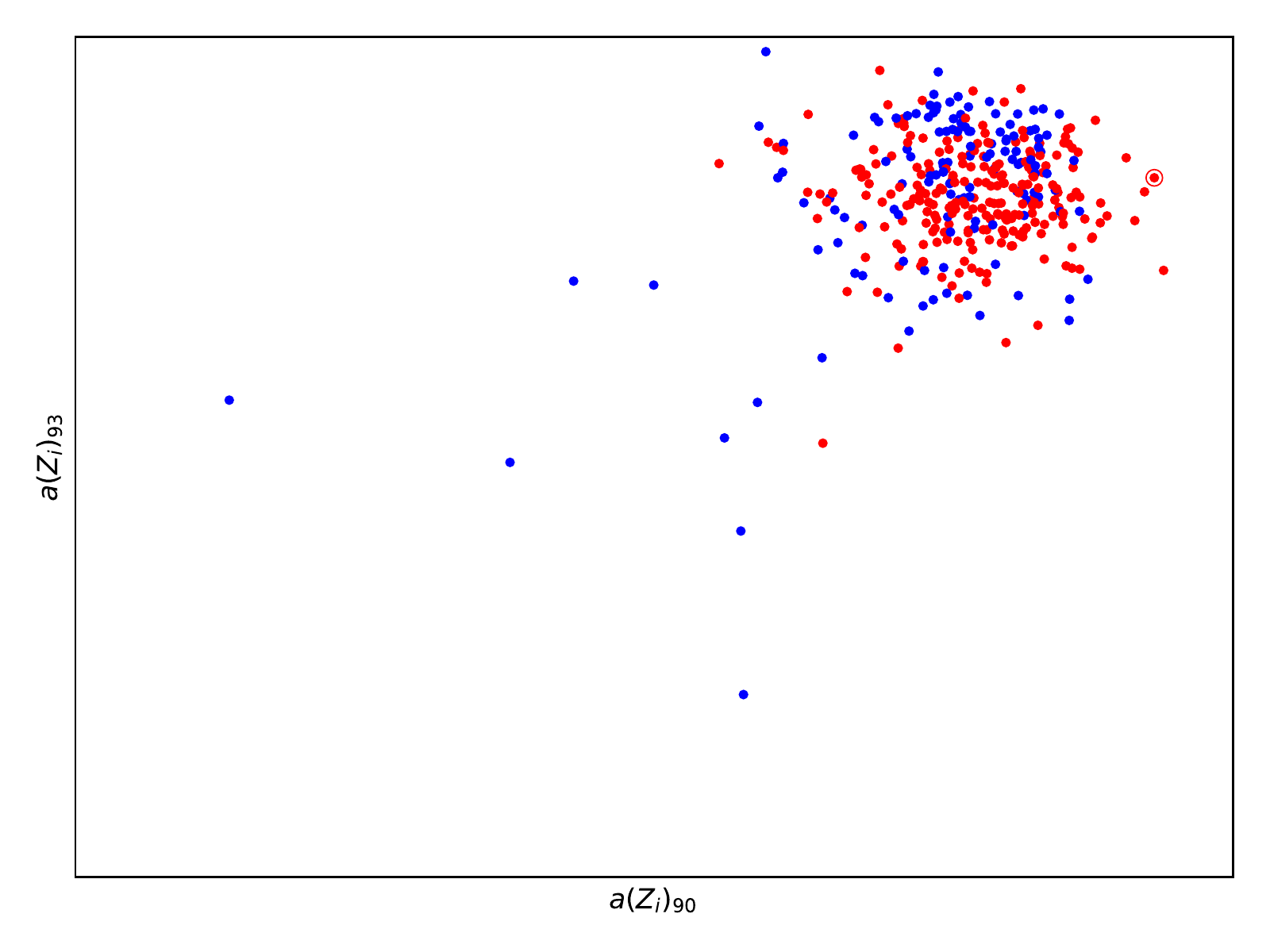}
    }
   	\end{minipage}
    \caption{
      Explaining the decision for a test time series from the HandOutlines dataset.
      (a) The series together with the shapelets that were prominent for the classification decision.
      The results for LS are shown at the top (3 shapelets) and ours \aiprnn{} at the bottom (2 shapelets, see (b)).
      (b) All time series of the dataset shown in a 2D embedding using the activation values for the two shapelets from (a). The series shown in (a) is circled in red (top right corner) and belongs to the "red" class. 
	\label{fig:visu_testseries}}
\end{figure*}

\subsection{Qualitative Results}

Our method aims at producing interpretable results in the sense that shapelets should be similar to sub-parts of some series from the dataset.
We first validate that our \aiprshort{} scheme actually ensures that shapelets are similar to the training data.
Then we show how shapelets that look like subseries are helpful to make the decision process interpretable.

We first illustrate our training process and its impact on a single shapelet in Figure~\ref{fig:shapelet_evolve}.
In this figure, we show the evolution of a given shapelet for the Wine dataset at epochs 20, 200, 800 and 8,000.
One can see from the loss values reported in Figures~\ref{fig:shapelet_evolve:wass} and~\ref{fig:shapelet_evolve:xent} that these correspond to different stages in our learning process. 
At epoch 20, the Wasserstein loss is far from the 0 value ($L_d=0$ corresponds to a case where the discriminator cannot distinguish between shapelets and real subseries), and this indeed corresponds to a shapelet that looks very different from an actual subseries.
As epochs go, both the Wasserstein loss $L_d$ and the cross-entropy one $L_c$ get closer to 0, leading to both realistic and discriminative shapelets.

To further check the effect of our regularization, we focus on the most discriminative shapelets for a bunch of datasets, as it would be misleading to look at a random shapelet: a shapelet might well be similar to a series but useless for the classification.
The discriminative power, for class $j$, of the shapelet at index $s$ with respect to the $i$-th time series in the training set is evaluated as:
\begin{equation}
	\mathcal{P}_{i,j}(s) = a(Z_i)_s \cdot w_{s,j}
\end{equation}
where $a(Z_i)_s$ is the $s$-th component (i.e. the one that corresponds to shapelet $s$) of the activation map for the time series $Z_i$ and $w_{s,j}$ is the weight connecting that $s$-th component to the $j$-th output in the logistic layer of our classifier.
As we aim at evaluating the overall discriminative power of a shapelet in a multi-class setting, and given that we use of a softmax activation at the input of our logistic layer, we can define the cross-class discriminative power of a shapelet as:
\begin{equation}
	\mathcal{P}_{i}(s) = \sum_j \left(a(Z_i)_s \cdot (w_{s,j} - \max_{j'} w_{s,j'}) \right)^2.
	\label{eq:discriminative_shapelets}
\end{equation}

\begin{figure}
	\centering
    	\includegraphics[width=.33\linewidth]{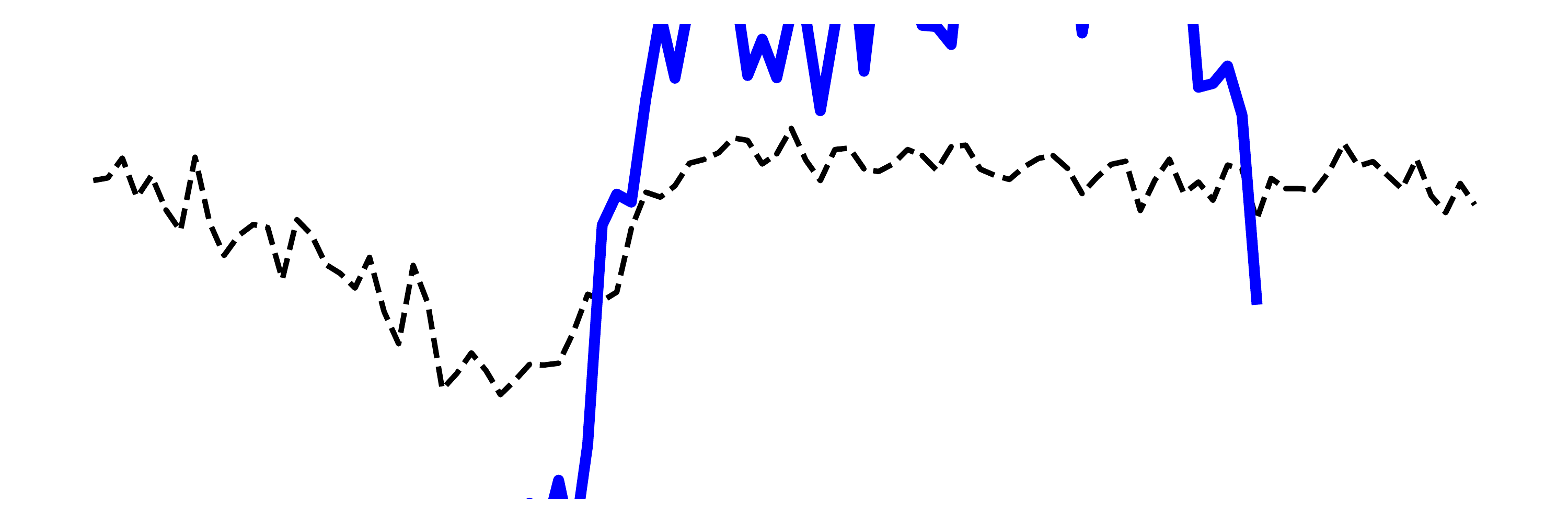}%
	    \includegraphics[width=.33\linewidth]{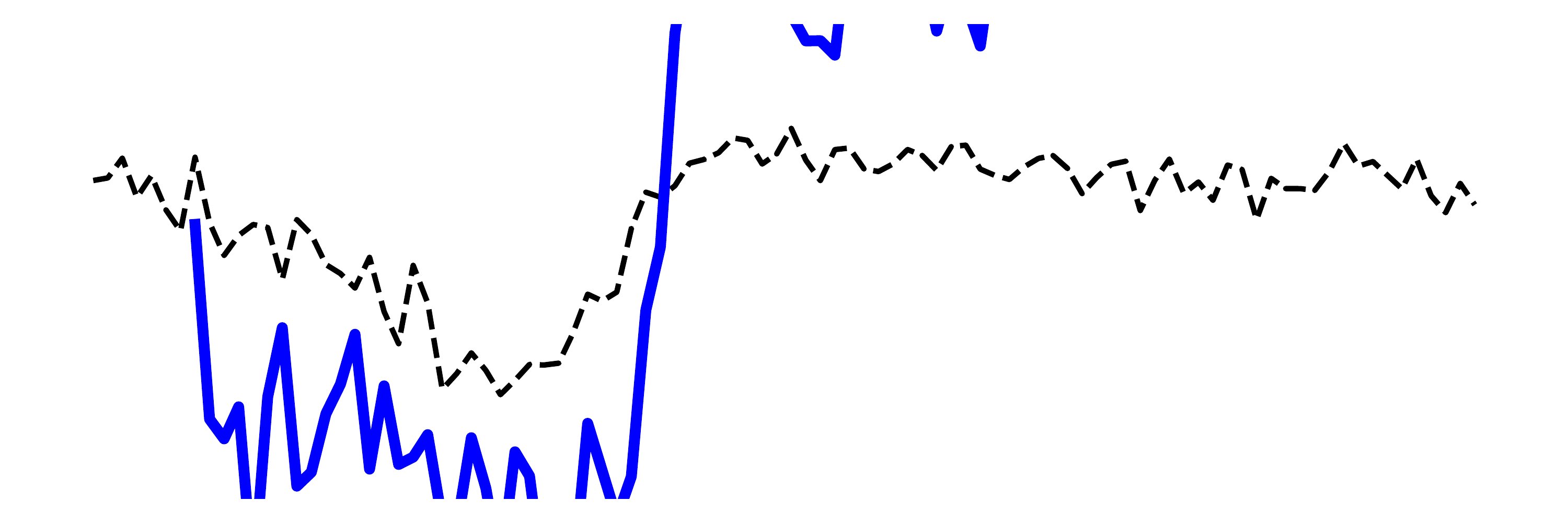}%
    	\includegraphics[width=.33\linewidth]{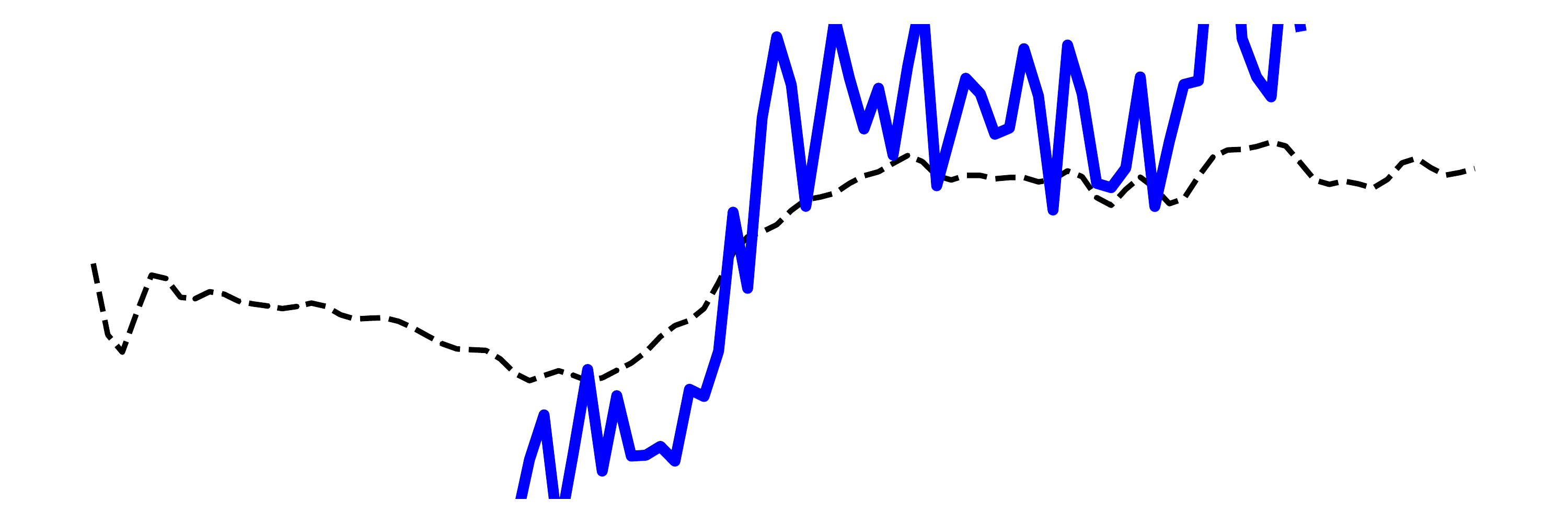}
    	\includegraphics[width=.33\linewidth]{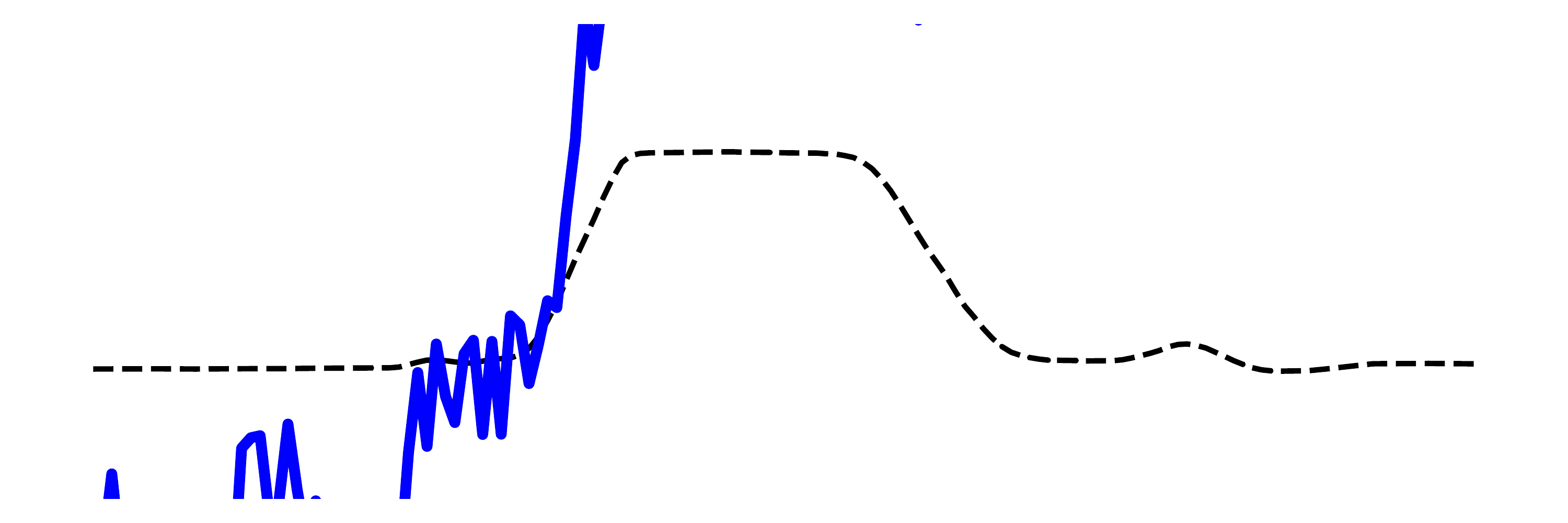}%
	    \includegraphics[width=.33\linewidth]{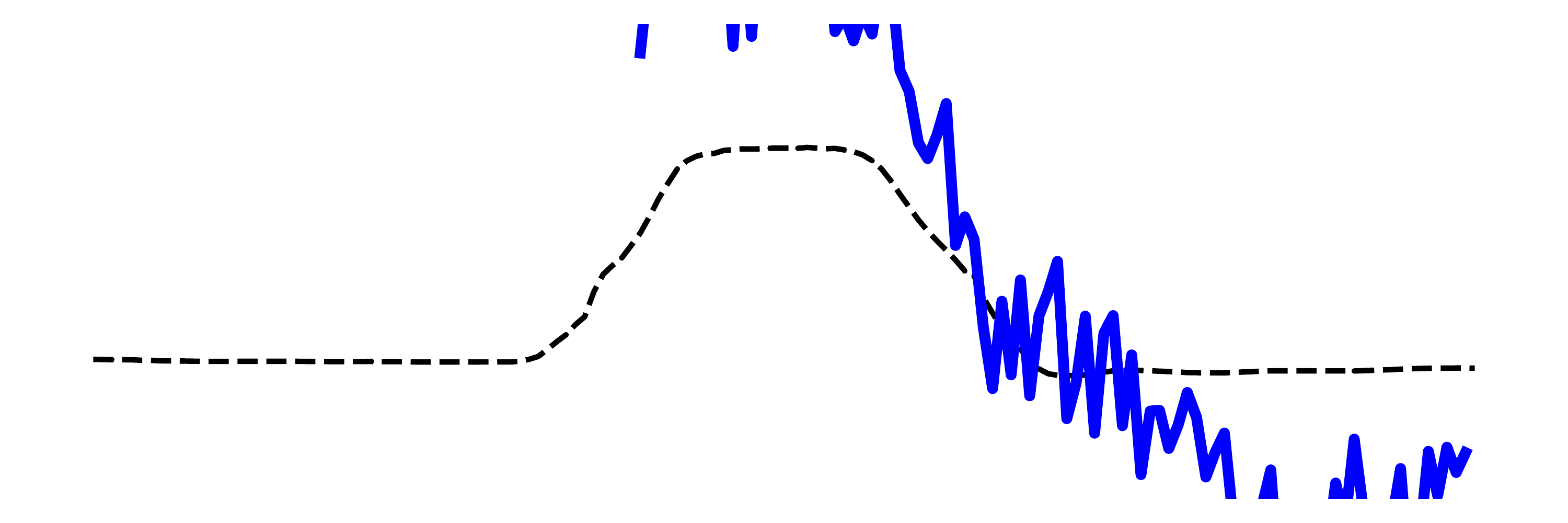}%
    	\includegraphics[width=.33\linewidth]{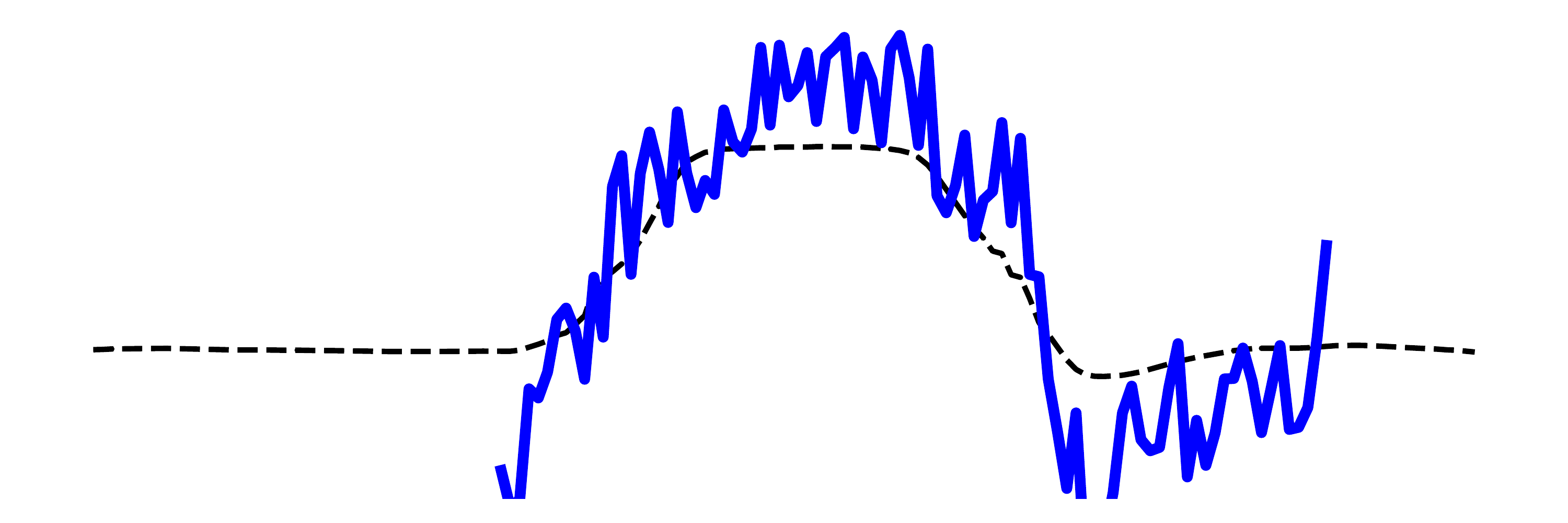}
    	\includegraphics[width=.33\linewidth]{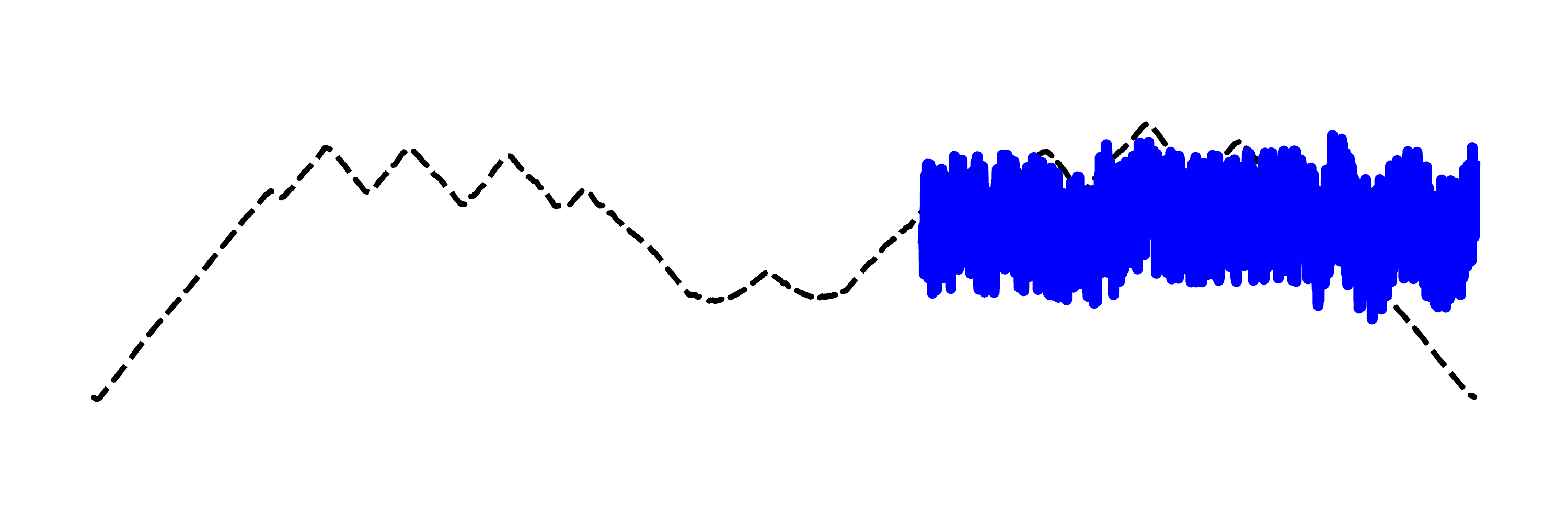}%
	    \includegraphics[width=.33\linewidth]{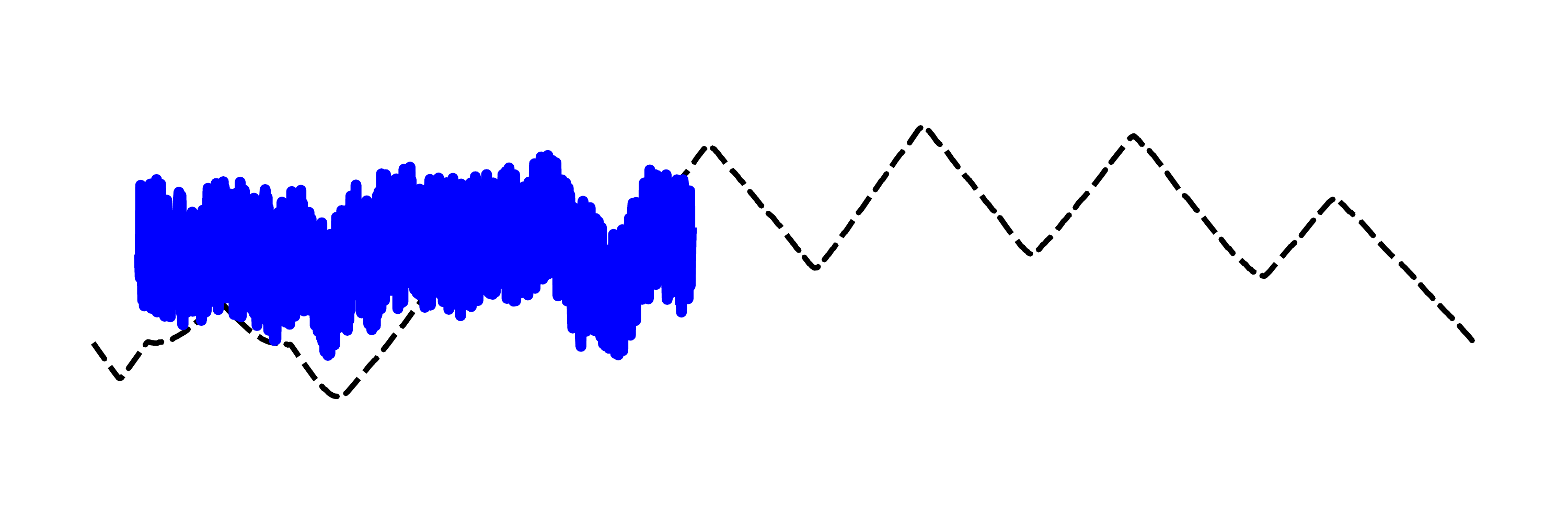}%
    	\includegraphics[width=.33\linewidth]{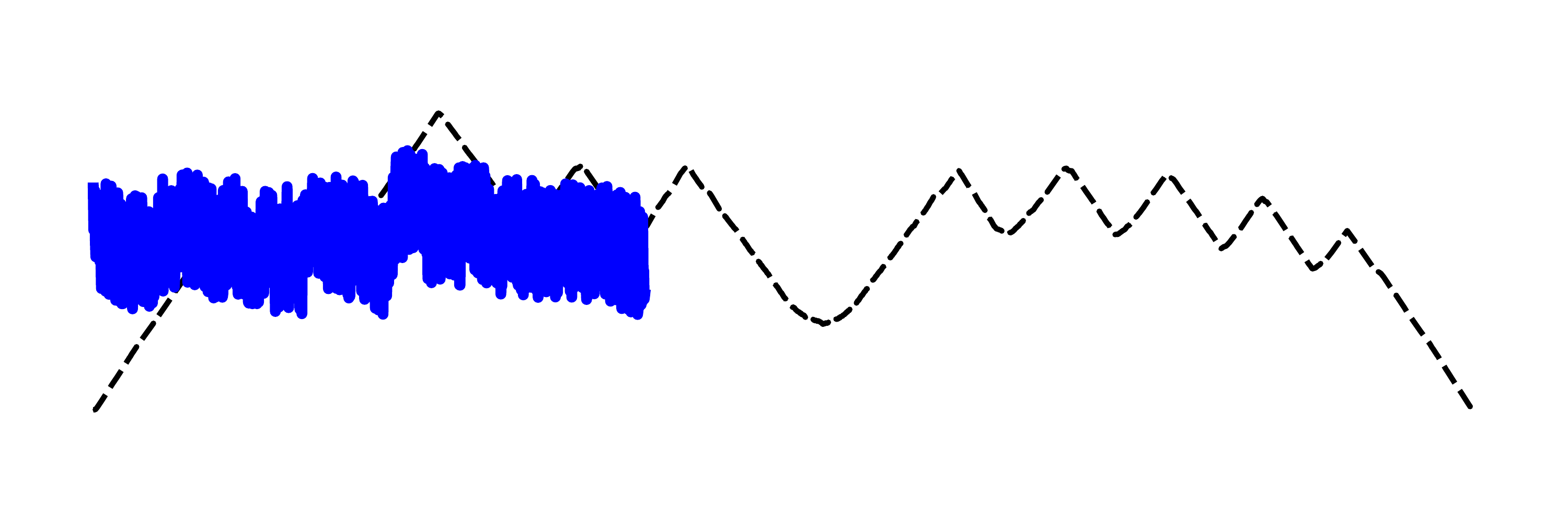}
    	\includegraphics[width=.33\linewidth]{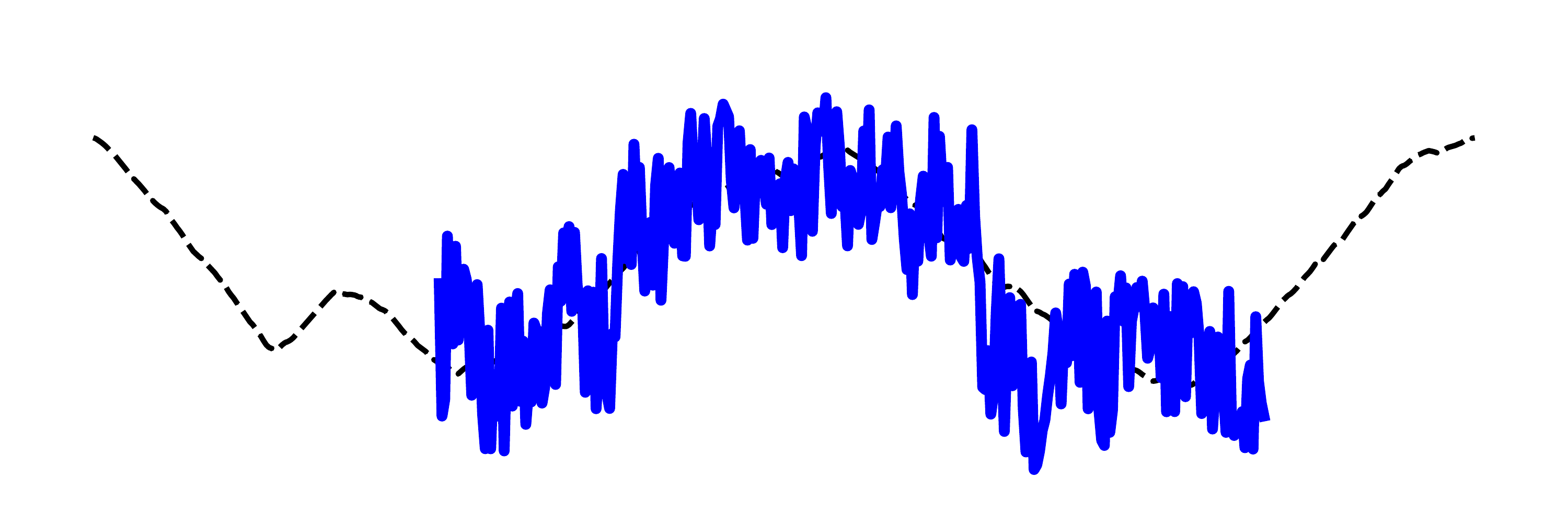}%
	    \includegraphics[width=.33\linewidth]{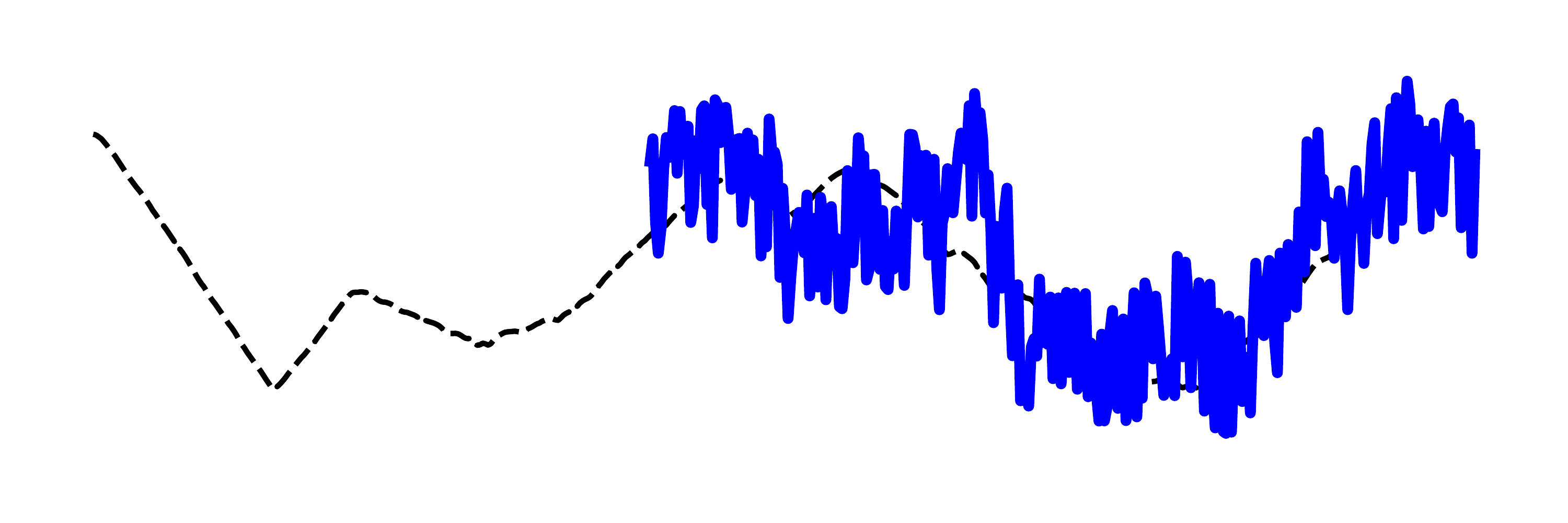}%
    	\includegraphics[width=.33\linewidth]{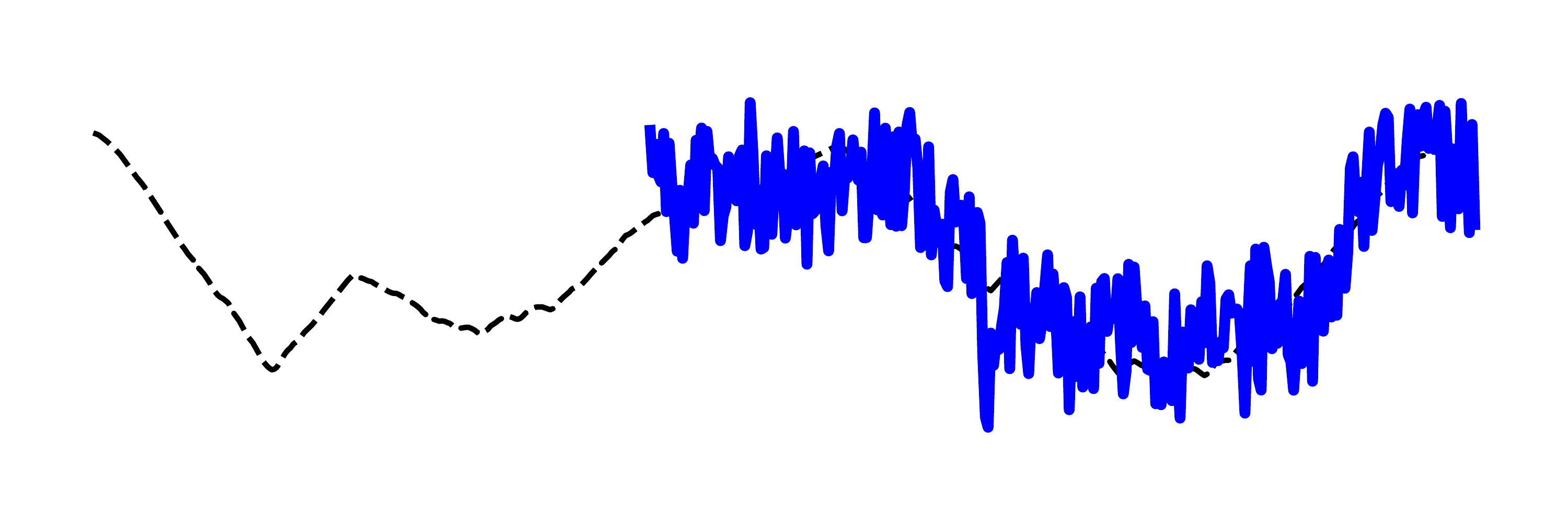}
    	\includegraphics[width=.33\linewidth]{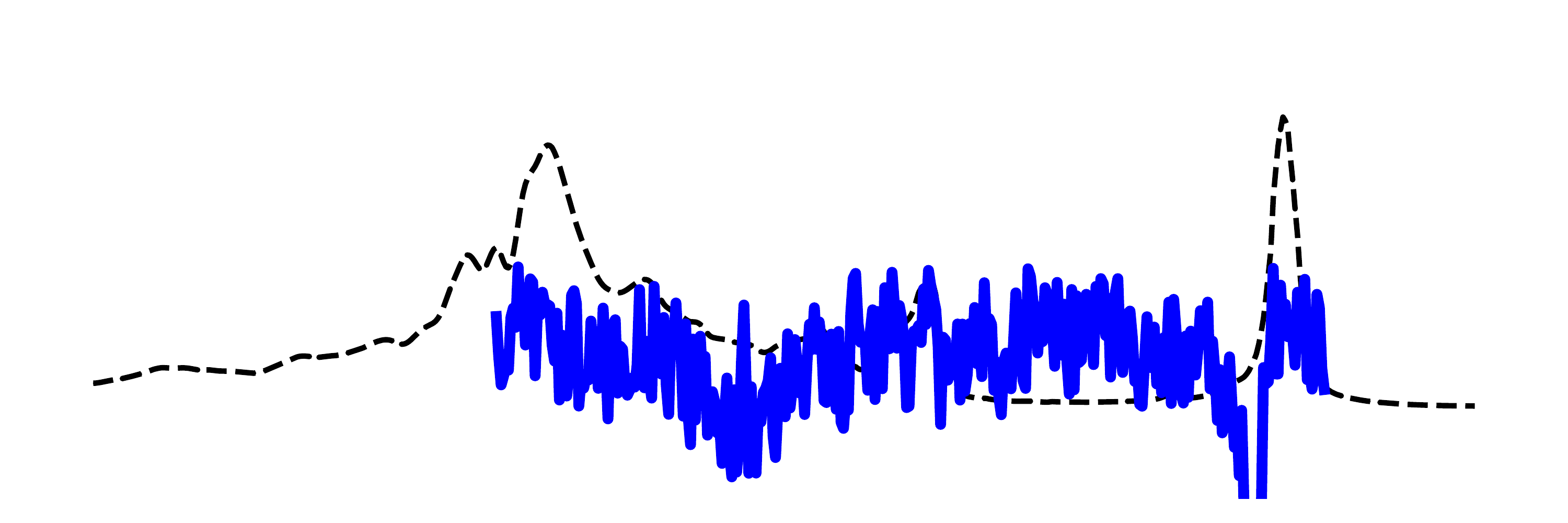}%
	    \includegraphics[width=.33\linewidth]{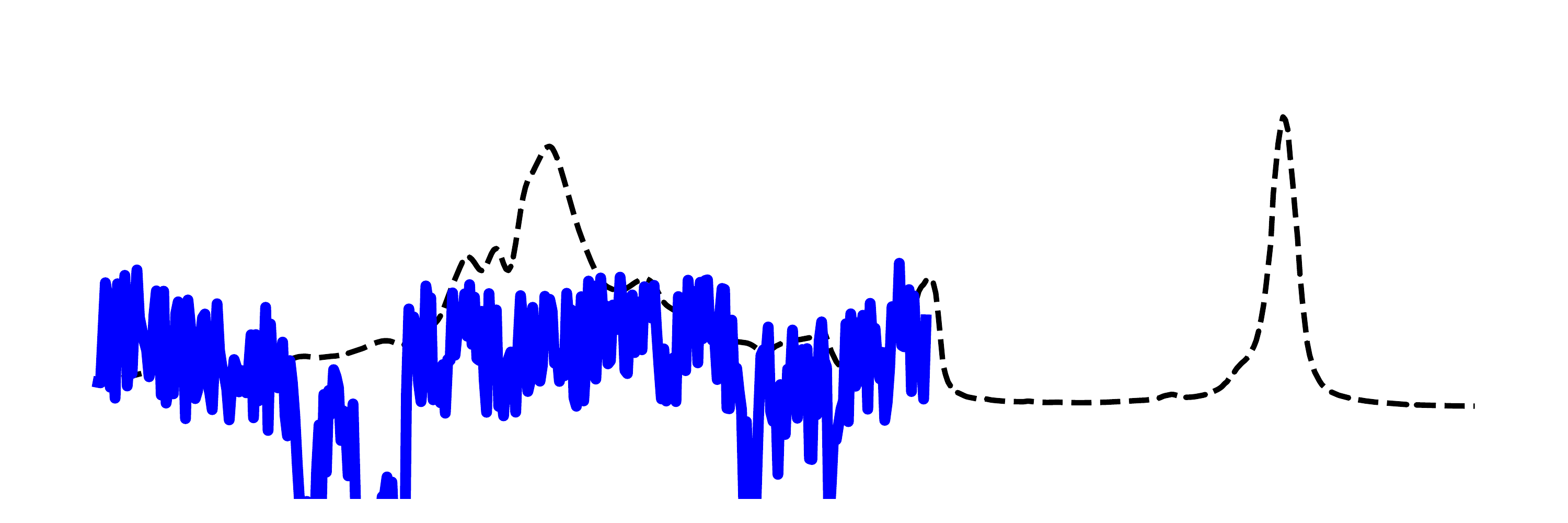}%
    	\includegraphics[width=.33\linewidth]{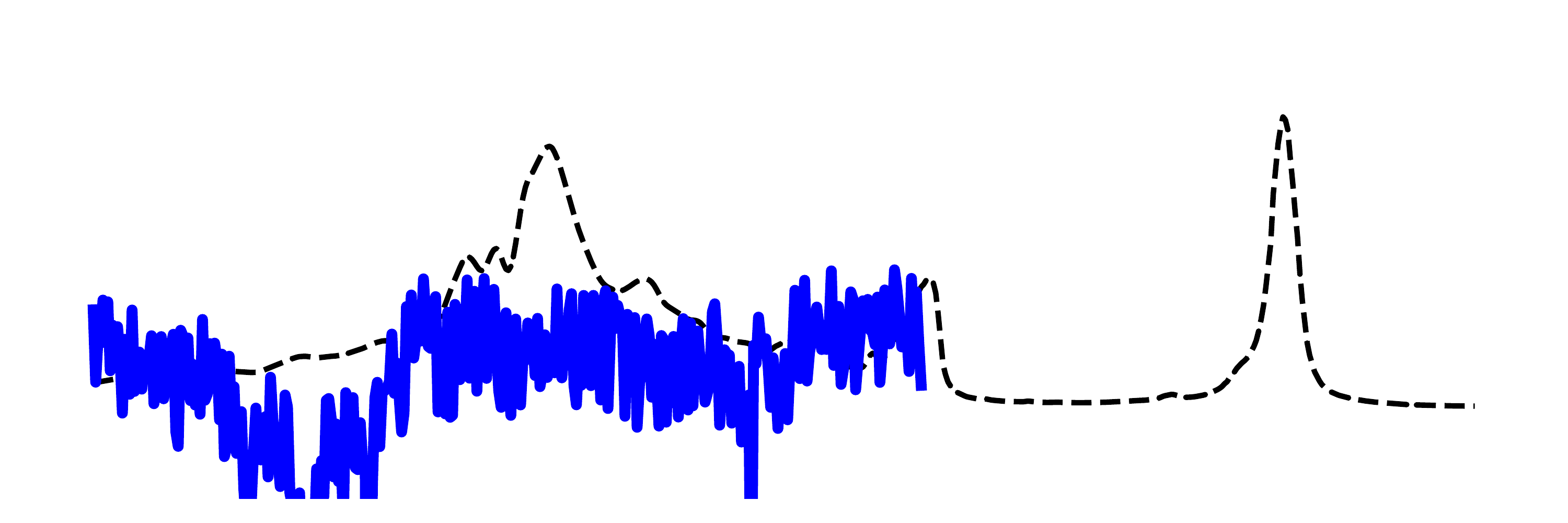}
	
	\caption{Shapelets obtained on the datasets (from top to bottom) ECG200, GunPoint, HandOutlines, Herring and OliveOil learned using only the bottom part of the architecture presented in Figure~\ref{fig:arn} (a simple CNN) superimposed over the best matching time series in the training set.\label{fig:visu_shapelets_cnn}}
\end{figure}

This is the criterion that we use to rank our shapelets in terms of discriminative power and to select the three most discriminative shapelets in Figure~\ref{fig:visu_shapelets}. This figure shows a significant improvement in terms of adequation of the shapelets to the training time series when using our \aiprnn{} model in place of a standard LS one.
Examples of the shapelets learned using only the classifier part of our neural network architecture (a simple CNN) are shown in Figure~\ref{fig:visu_shapelets_cnn}. This figure reveals that an unregularized network fails at generating interpretable shapelets just as LS does. This shows that the actual benefit in interpretability indeed comes from our \aiprshort{} scheme. Our regularization strategy allows to generate shapelets that are both discriminative and representative of the training data.

Another important aspect, in terms of interpretability, is the explanation that can be provided to an end-user to explain a classification decision.
For a given test time series, we produce two representations that help the user understand and trust the decision of a classifier.
First, in Figure~\ref{fig:illus_intro} and Figure~\ref{fig:visu_testseries}a, we present the shapelets that were the most important to make a classification decision (according to Equation~\ref{eq:discriminative_shapelets}).
One can notice that in both cases, shapelets extracted by \aiprnn{} better fit the time series at stake and hence help the end-user focus on the particular pattern in the time series that leads to the decision (e.g. the series of three peaks for HandOutlines or the overall shape of the central hump for Herring).
Next, in Figure~\ref{fig:visu_testseries}b, we present a 2D embedding of all the time series of the dataset, using the two most important shapelets for the considered time series.
One can see that the considered time series (circled in red) lies in a part of the space where there are only ``red class'' time series.
With these two representations for a test time series, the end-user:
knows what are the most important shapelets (and their location) used by the model for its decision,
and, can be convinced that these shapelets are good or sufficient to isolate the time series into a given class.
When the considered time series correspond to actual subseries as with our method, this allows the end-user to better understand the decision process.

\subsection{Quantitative Results}
Our \aiprshort{} is able to recover shapelets that are discriminative and similar to the input, as expected.
We want to quantify if this is achieved at the expense of classification accuracy and/or computation time. Our goal is to be much faster than exhaustive shapelet search methods (our baseline is Shapelets~\cite{ye2009time}), much more accurate than very fast random shapelet selection-based methods (our baseline is FS~\cite{Rak13}) and as accurate and as fast as single model shapelet learning methods (our baseline is LS~\cite{Grabocka2014}).

\subsubsection{Accuracy}
\begin{figure}[t!]
\centering
	\includegraphics[width=0.8\textwidth]{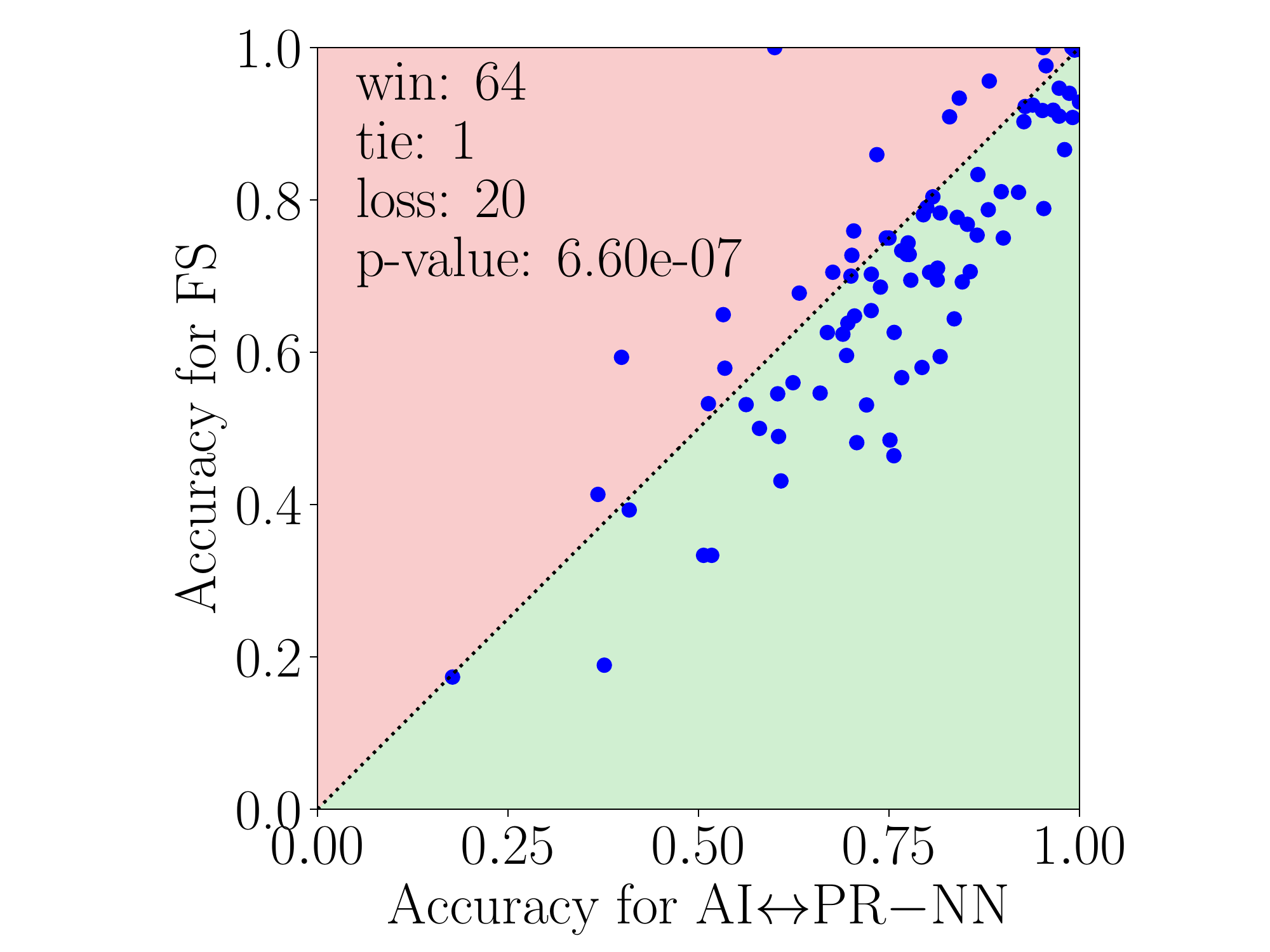}
	\caption{Accuracy comparison between Fast Shapelets and our \aiprshort{} method on 85 datasets (each point is a dataset) of the UCR/UEA repository~\cite{UCRArchive}.}\label{fig:FSvsA}
\end{figure}

\begin{figure}[h!]
\centering
	\includegraphics[width=0.8\textwidth]{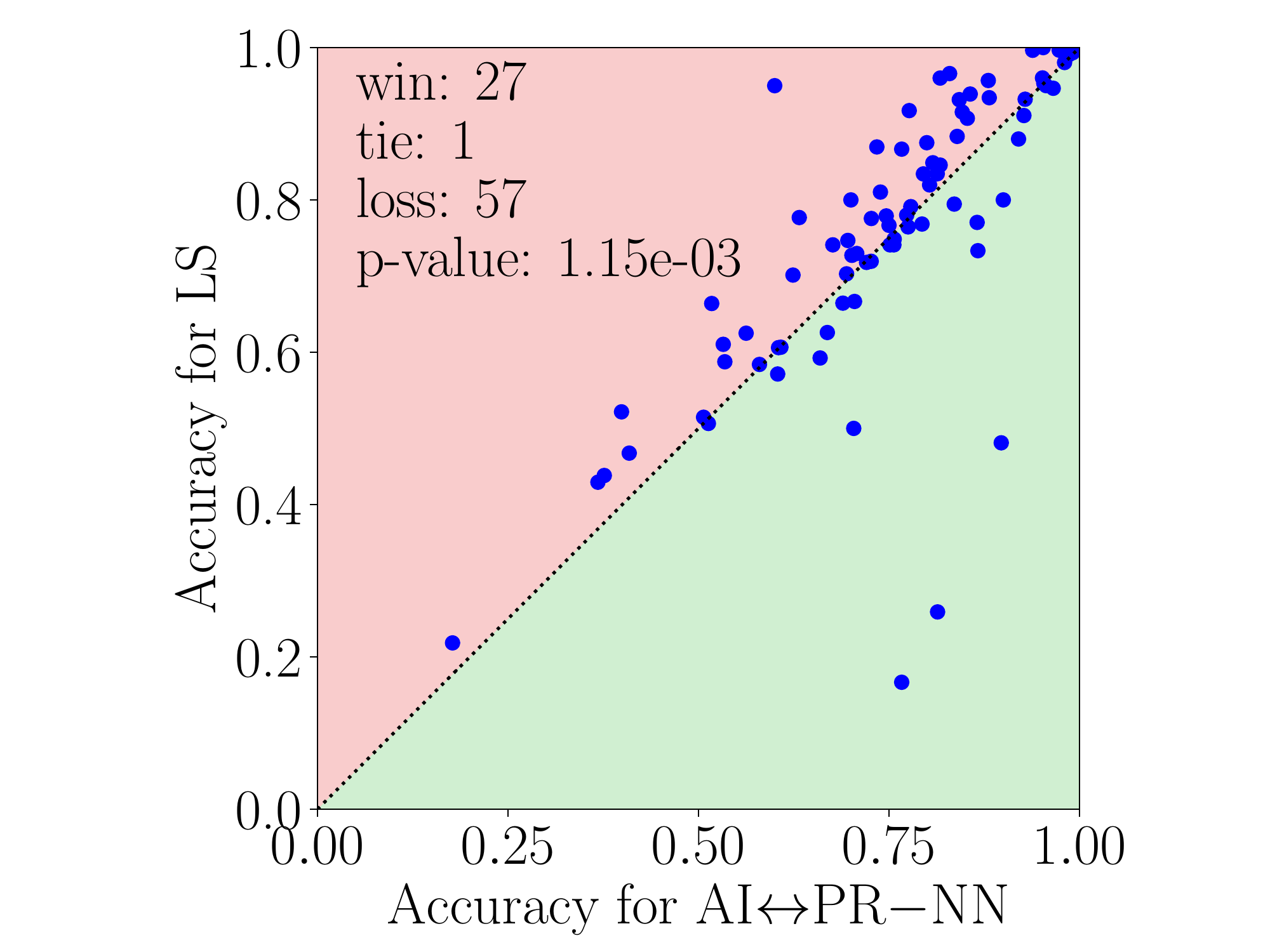}
	\caption{Accuracy comparison between Learning Shapelets and our \aiprshort{} method on 85 datasets (each point is a dataset) of the UCR/UEA repository~\cite{UCRArchive}.}\label{fig:LSvsA}
\end{figure}

\begin{figure}[h!]
\centering
	\includegraphics[width=0.8\textwidth]{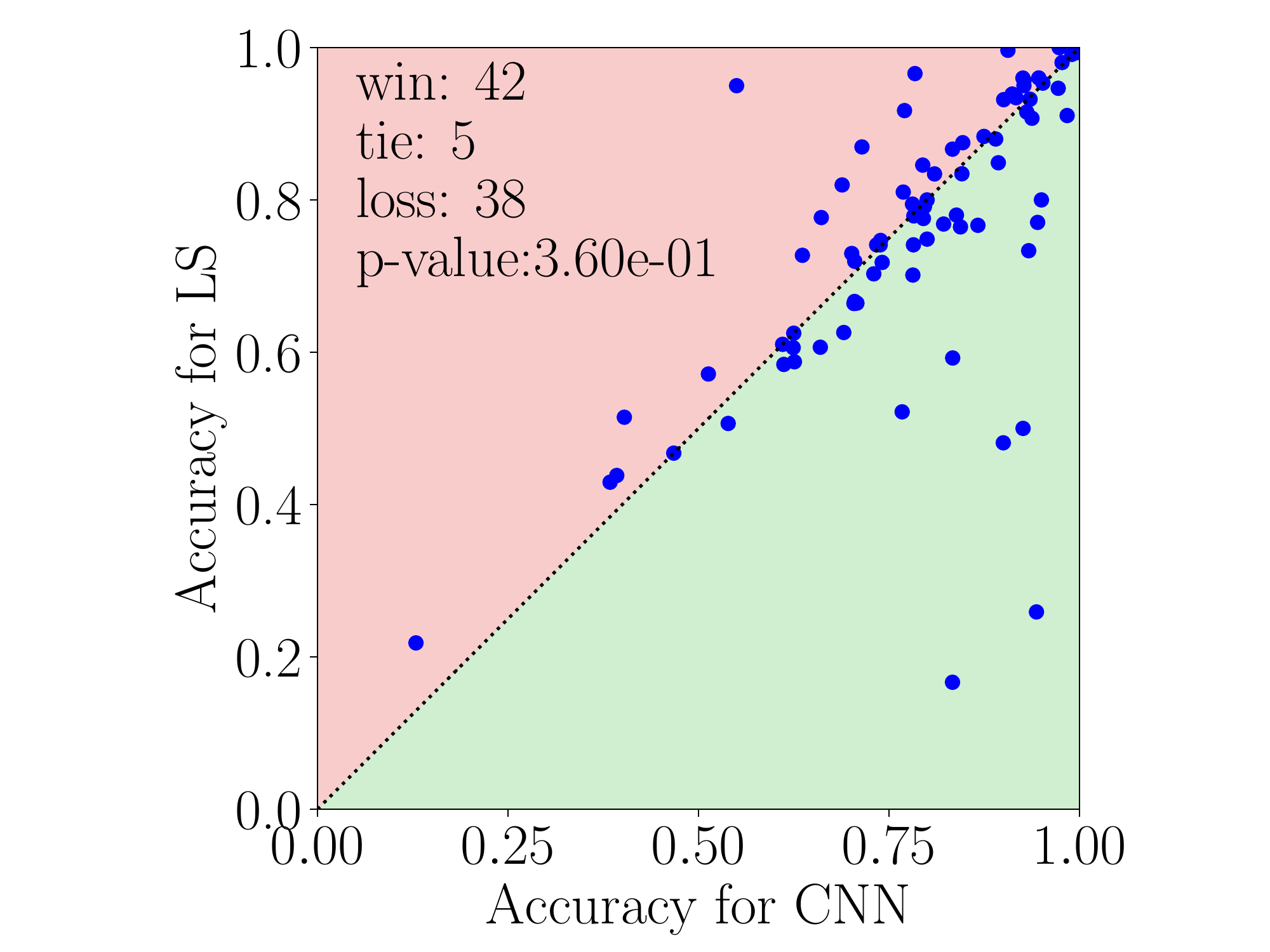}
	\caption{Accuracy comparison between Learning Shapelets and a simple CNN on 85 datasets of the UCR/UEA repository~\cite{UCRArchive}.}\label{fig:LSvsCNN}
\end{figure}

We analyze the accuracies obtained by FS, LS and our \aiprnn{} method on the 85 datasets using scatter plots.\footnote{See Appendix \ref{appendix} for detailed dataset information and accuracy.}
The results of the shapelet-based baselines used in this section come from~\cite{Bagnall2017} (the results for Shapelets~\cite{ye2009time} are not available because the method already does not scale on small size datasets).
We compare FS versus \aiprnn{} in Figure~\ref{fig:FSvsA} and LS versus \aiprnn{} in Figure \ref{fig:LSvsA}. We also show how a simple CNN (without the adversarial regularization) compares against LS in Figure~\ref{fig:LSvsCNN}. 
We indicate the number of \emph{win/tie/loss} for our method and we provide a Wilcoxon significance test \cite{Demsar2006} with the resulting $p$-value ($> 0.01$: none of the two methods is significantly better than the other). The points on the diagonal are datasets for which the accuracy is identical for both competitors.
Figure~\ref{fig:FSvsA} shows that, as expected, our method yields significantly better performance than FS. Compared to LS, for most datasets, the difference in accuracy is low, with a small edge (significant) for LS: on average for the 85 datasets, LS obtains an accuracy of 0.77 whereas \aiprnn{} obtains an accuracy 0.76. On three datasets (namely HandOutlines, NonInvasiveFetalECGThorax1 and OliveOil), our \aiprnn{} method and its regularization seems to be strongly positive (and detrimental on one dataset), in terms of generalization. The simple CNN seems to give slightly better (non significant) results than LS (and thus than our \aiprnn{}): on average for the 85 datasets, the simple CNN obtains an accuracy of 0.8. This means that our backbone neural network architecture is a good candidate to jointly learn interpretable shapelets and classify time series with little loss on accuracy.

\begin{table}
\centering
\caption{Complexity of four different shapelet-based TSC algorithms (Shapelet~\cite{ye2009time}, FS~\cite{Rak13}, LS~\cite{Grabocka2014} and \aiprnn{}). $n$ is the number of examples in the training set, $T$ is the average length of the time series, and $c$ is the number of classes.}\label{tab:complexity}
	\begin{tabular}{ccccc}
		\hline
		Shapelet     & FS          & LS             & \aiprnn{}        \\
		$O(n^2 \cdot T^4)$ & $O(n \cdot T^2)$  & $O(n \cdot T^2 \cdot c)$ & $O(n \cdot T^2 \cdot c)$ \\
		\hline
	\end{tabular}	
\end{table}

\subsubsection{Training Time}
We provide both a theoretical complexity study (see Table~\ref{tab:complexity}) of all the baselines and of our \aiprnn{} method. Some complexities were already given in Section \ref{sec:related}. Our method is based on a classifier and a discriminator, and both of them are simple CNNs. So the complexity of our algorithm ($O(n \cdot T^2 \cdot c)$) is related to training a CNN and should depend mainly on the number of examples ($n$), the average length of the time series ($T$), and the number of classes ($c$, since the latter is used to decide the number of shapelets to be learned). Note that for both LS and \aiprnn, the parameter $n$ could be considered as a (quite big) constant since the number of epochs (i.e. the number of times the algorithm "sees" the entire dataset) is fixed in the experiments. However, in LS, this number still depends on $n$ whereas it is fixed once and for all (to $8000$) in \aiprnn{}. This difference is in favor of LS for small datasets and in favor of \aiprnn{} for larger ones. 

To have a better grasp on the actual training time of all methods, we ran the methods on a single dataset (ElectricDevices) and recorded the CPU time. The experiments were conducted on a Debian Cluster using Intel(R) Xeon(R) CPU E5-2650 v4 Processor (12 core 2.20 GHz CPU) with 32GB memory. The results are averaged over five runs. The implementation code of our baselines is taken from~\cite{Bagnall2017} (as for the accuracy results).
As expected, the original Shapelet~\cite{ye2009time} method does not finish in 48 hours for this medium size dataset. FS finishes in 12.1 minutes, LS finishes in 2323 minutes, and our method takes 142 minutes. The theoretical complexity of LS and \aiprnn{} is identical so these results were surprising. We suspected that the JAVA implementation of LS was not well optimized and we re-implemented the LS method with Keras\footnote{\url{https://keras.io/}}.
With this new implementation, the training phase took only $71$ minutes for LS on this dataset (compared to 142 for \aiprnn{}) which shows that the time difference between the two algorithms is mainly related to the implementation (and the hyper-parameters related to the number of epochs).

%% file: sections/supp.tex
\renewcommand\arraystretch{1.18}
\begin{landscape}
\begin{appendix}
\section{Dataset information and accuracy comparison}\label{appendix}
\begin{longtable}{@{\extracolsep{\fill}}|l|cccc|cccc|@{}}
\hline
DatasetName                    & nb\_train & nb\_test & length  & class & FS     & LS     & CNN    & AI$\leftrightarrow$PR   \\ \hline
Adiac                          & 390       & 391      & 176  & 37    & 0.5934 & 0.5217 & 0.7673 & 0.3990 \\ \hline
ArrowHead                      & 36        & 175      & 251  & 3     & 0.5943 & 0.8457 & 0.7943 & 0.8171 \\ \hline
Beef                           & 30        & 30       & 470  & 5     & 0.5667 & 0.8667 & 0.8333 & 0.7667 \\ \hline
BeetleFly                      & 20        & 20       & 512  & 2     & 0.7000 & 0.8000 & 0.8000 & 0.7000 \\ \hline
BirdChicken                    & 20        & 20       & 512  & 2     & 0.7500 & 0.8000 & 0.9500 & 0.9000 \\ \hline
Car                            & 60        & 60       & 577  & 4     & 0.7500 & 0.7667 & 0.8667 & 0.7500 \\ \hline
CBF                            & 30        & 900      & 128  & 3     & 0.9400 & 0.9911 & 0.9900 & 0.9867 \\ \hline
ChlorineConcentration          & 467       & 3840     & 166  & 3     & 0.5464 & 0.5924 & 0.8336 & 0.6596 \\ \hline
CinCECGTorso                   & 40        & 1380     & 1639 & 4     & 0.8594 & 0.8696 & 0.7145 & 0.7341 \\ \hline
Coffee                         & 28        & 28       & 286  & 2     & 0.9286 & 1.0000 & 1.0000 & 1.0000 \\ \hline
Computers                      & 250       & 250      & 720  & 2     & 0.5000 & 0.5840 & 0.6120 & 0.5800 \\ \hline
CricketX                       & 390       & 390      & 300  & 12    & 0.4846 & 0.7410 & 0.7385 & 0.7513 \\ \hline
CricketY                       & 390       & 390      & 300  & 12    & 0.5308 & 0.7179 & 0.7410 & 0.7205 \\ \hline
CricketZ                       & 390       & 390      & 300  & 12    & 0.4641 & 0.7410 & 0.7821 & 0.7564 \\ \hline
DiatomSizeReduction            & 16        & 306      & 345  & 4     & 0.8660 & 0.9804 & 0.9771 & 0.9804 \\ \hline
DistalPhalanxOutlineAgeGroup   & 400       & 139      & 80   & 3     & 0.6547 & 0.7194 & 0.7050 & 0.7266 \\ \hline
DistalPhalanxOutlineCorrect    & 600       & 276      & 80   & 2     & 0.7500 & 0.7790 & 0.7826 & 0.7464 \\ \hline
DistalPhalanxTW                & 400       & 139      & 80   & 6     & 0.6259 & 0.6259 & 0.6906 & 0.6691 \\ \hline
Earthquakes                    & 322       & 139      & 512  & 2     & 0.7050 & 0.7410 & 0.7338 & 0.6763 \\ \hline
ECG200                         & 100       & 100      & 96   & 2     & 0.8100 & 0.8800 & 0.8900 & 0.9200 \\ \hline
ECG5000                        & 500       & 4500     & 140  & 5     & 0.9227 & 0.9322 & 0.9351 & 0.9287 \\ \hline
ECGFiveDays                    & 23        & 861      & 136  & 2     & 0.9977 & 1.0000 & 1.0000 & 0.9977 \\ \hline
ElectricDevices                & 8926      & 7711     & 96   & 7     & 0.5790 & 0.5875 & 0.6259 & 0.5346 \\ \hline
FaceAll                        & 560       & 1690     & 131  & 14    & 0.6260 & 0.7485 & 0.8000 & 0.7568 \\ \hline
FaceFour                       & 24        & 88       & 350  & 4     & 0.9091 & 0.9659 & 0.7841 & 0.8295 \\ \hline
FacesUCR                       & 200       & 2050     & 131  & 14    & 0.7059 & 0.9390 & 0.9117 & 0.8566 \\ \hline
FiftyWords                     & 450       & 455      & 270  & 50    & 0.4813 & 0.7297 & 0.7011 & 0.7077 \\ \hline
Fish                           & 175       & 175      & 463  & 7     & 0.7829 & 0.9600 & 0.9257 & 0.8171 \\ \hline
FordA                          & 3601      & 1320     & 500  & 2     & 0.7871 & 0.9568 & 0.9273 & 0.8803 \\ \hline
FordB                          & 3636      & 810      & 500  & 2     & 0.7284 & 0.9173 & 0.7704 & 0.7765 \\ \hline
GunPoint                       & 50        & 150      & 150  & 2     & 0.9467 & 1.0000 & 0.9733 & 0.9733 \\ \hline
Ham                            & 109       & 105      & 431  & 2     & 0.6476 & 0.6667 & 0.7048 & 0.7048 \\ \hline
HandOutlines                   & 1000      & 370      & 2709 & 2     & 0.8108 & 0.4811 & 0.9000 & 0.8973 \\ \hline
Haptics                        & 155       & 308      & 1092 & 5     & 0.3929 & 0.4675 & 0.4675 & 0.4091 \\ \hline
Herring                        & 64        & 64       & 512  & 2     & 0.5313 & 0.6250 & 0.6250 & 0.5625 \\ \hline
InlineSkate                    & 100       & 550      & 1882 & 7     & 0.1891 & 0.4382 & 0.3927 & 0.3764 \\ \hline
InsectWingbeatSound            & 220       & 1980     & 256  & 11    & 0.4894 & 0.6061 & 0.6242 & 0.6051 \\ \hline
ItalyPowerDemand               & 67        & 1029     & 24   & 2     & 0.9174 & 0.9602 & 0.9466 & 0.9514 \\ \hline
LargeKitchenAppliances         & 375       & 375      & 720  & 3     & 0.5600 & 0.7013 & 0.7813 & 0.6240 \\ \hline
Lightning2                     & 60        & 61       & 637  & 2     & 0.7049 & 0.8197 & 0.6885 & 0.8033 \\ \hline
Lightning7                     & 70        & 73       & 319  & 7     & 0.6438 & 0.7945 & 0.7808 & 0.8356 \\ \hline
Mallat                         & 55        & 2345     & 1024 & 8     & 0.9761 & 0.9501 & 0.9271 & 0.9561 \\ \hline
Meat                           & 60        & 60       & 448  & 3     & 0.8333 & 0.7333 & 0.9333 & 0.8667 \\ \hline
MedicalImages                  & 381       & 760      & 99   & 10    & 0.6237 & 0.6645 & 0.7079 & 0.6895 \\ \hline
MiddlePhalanxOutlineAgeGroup   & 400       & 154      & 80   & 3     & 0.5455 & 0.5714 & 0.5130 & 0.6039 \\ \hline
MiddlePhalanxOutlineCorrect    & 600       & 291      & 80   & 2     & 0.7285 & 0.7801 & 0.8385 & 0.7732 \\ \hline
MiddlePhalanxTW                & 399       & 154      & 80   & 6     & 0.5325 & 0.5065 & 0.5390 & 0.5130 \\ \hline
MoteStrain                     & 20        & 1252     & 84   & 2     & 0.7772 & 0.8834 & 0.8746 & 0.8395 \\ \hline
NonInvasiveFatalECGThorax1     & 1800      & 1965     & 750  & 42    & 0.7104 & 0.2590 & 0.9435 & 0.8137 \\ \hline
NonInvasiveFatalECGThorax2     & 1800      & 1965     & 750  & 42    & 0.7537 & 0.7705 & 0.9450 & 0.8656 \\ \hline
OliveOil                       & 30        & 30       & 570  & 4     & 0.7333 & 0.1667 & 0.8333 & 0.7667 \\ \hline
OSULeaf                        & 200       & 242      & 427  & 6     & 0.6777 & 0.7769 & 0.6612 & 0.6322 \\ \hline
PhalangesOutlinesCorrect       & 1800      & 858      & 80   & 2     & 0.7436 & 0.7646 & 0.8438 & 0.7751 \\ \hline
Phoneme                        & 214       & 1896     & 1024 & 39    & 0.1735 & 0.2184 & 0.1292 & 0.1772 \\ \hline
Plane                          & 105       & 105      & 144  & 7     & 1.0000 & 1.0000 & 1.0000 & 0.9524 \\ \hline
ProximalPhalanxOutlineAgeGroup & 400       & 205      & 80   & 3     & 0.7805 & 0.8341 & 0.8098 & 0.7951 \\ \hline
ProximalPhalanxOutlineCorrect  & 600       & 291      & 80   & 2     & 0.8041 & 0.8488 & 0.8935 & 0.8076 \\ \hline
ProximalPhalanxTW              & 400       & 205      & 80   & 6     & 0.7024 & 0.7756 & 0.7951 & 0.7268 \\ \hline
RefrigerationDevices           & 375       & 375      & 720  & 3     & 0.3333 & 0.5147 & 0.4027 & 0.5067 \\ \hline
ScreenType                     & 375       & 375      & 720  & 3     & 0.4133 & 0.4293 & 0.3840 & 0.3680 \\ \hline
ShapeletSim                    & 20        & 180      & 500  & 2     & 1.0000 & 0.9500 & 0.5500 & 0.6000 \\ \hline
ShapesAll                      & 600       & 600      & 512  & 60    & 0.5800 & 0.7683 & 0.8217 & 0.7933 \\ \hline
SmallKitchenAppliances         & 375       & 375      & 720  & 3     & 0.3333 & 0.6640 & 0.7040 & 0.5173 \\ \hline
SonyAIBORobotSurface1          & 20        & 601      & 70   & 2     & 0.6855 & 0.8103 & 0.7687 & 0.7388 \\ \hline
SonyAIBORobotSurface2          & 27        & 953      & 65   & 2     & 0.7901 & 0.8751 & 0.8468 & 0.7996 \\ \hline
StarLightCurves                & 1000      & 8236     & 1024 & 3     & 0.9178 & 0.9466 & 0.9721 & 0.9655 \\ \hline
Strawberry                     & 613       & 370      & 235  & 2     & 0.9027 & 0.9108 & 0.9838 & 0.9270 \\ \hline
SwedishLeaf                    & 500       & 625      & 128  & 15    & 0.7680 & 0.9072 & 0.9376 & 0.8528 \\ \hline
Symbols                        & 25        & 995      & 398  & 6     & 0.9337 & 0.9317 & 0.9005 & 0.8422 \\ \hline
SyntheticControl               & 300       & 300      & 60   & 6     & 0.9100 & 0.9967 & 0.9967 & 0.9733 \\ \hline
ToeSegmentation1               & 40        & 228      & 277  & 2     & 0.9561 & 0.9342 & 0.9167 & 0.8816 \\ \hline
ToeSegmentation2               & 36        & 130      & 343  & 2     & 0.6923 & 0.9154 & 0.9308 & 0.8462 \\ \hline
Trace                          & 100       & 100      & 275  & 4     & 1.0000 & 1.0000 & 1.0000 & 0.9900 \\ \hline
TwoLeadECG                     & 23        & 1139     & 82   & 2     & 0.9245 & 0.9965 & 0.9061 & 0.9385 \\ \hline
TwoPatterns                    & 1000      & 4000     & 128  & 4     & 0.9083 & 0.9933 & 0.9958 & 0.9910 \\ \hline
UWaveGestureLibraryAll         & 896       & 3582     & 945  & 8     & 0.7887 & 0.9534 & 0.9520 & 0.9531 \\ \hline
UWaveGestureLibraryX           & 896       & 3582     & 315  & 8     & 0.6946 & 0.7912 & 0.7965 & 0.7786 \\ \hline
UWaveGestureLibraryY           & 896       & 3582     & 315  & 8     & 0.5958 & 0.7030 & 0.7300 & 0.6943 \\ \hline
UWaveGestureLibraryZ           & 896       & 3582     & 315  & 8     & 0.6382 & 0.7468 & 0.7390 & 0.6960 \\ \hline
Wafer                          & 1000      & 6164     & 152  & 2     & 0.9968 & 0.9961 & 0.9972 & 0.9935 \\ \hline
Wine                           & 57        & 54       & 234  & 2     & 0.7593 & 0.5000 & 0.9259 & 0.7037 \\ \hline
WordSynonyms                   & 267       & 638      & 270  & 25    & 0.4310 & 0.6066 & 0.6599 & 0.6082 \\ \hline
Worms                          & 181       & 77       & 900  & 5     & 0.6494 & 0.6104 & 0.6104 & 0.5325 \\ \hline
WormsTwoClass                  & 181       & 77       & 900  & 2     & 0.7273 & 0.7273 & 0.6364 & 0.7013 \\ \hline
Yoga                           & 300       & 3000     & 426  & 2     & 0.6950 & 0.8343 & 0.8457 & 0.8133 \\ \hline
\end{longtable}
\end{appendix}
\end{landscape}